\documentclass{naturep}
\usepackage[T1]{fontenc}
\usepackage[utf8]{inputenc}
\usepackage[letterpaper,margin=0.65in,includefoot]{geometry}
\usepackage{graphicx}
\let\figinclude\includegraphics
\usepackage[table]{xcolor}
\usepackage{amsmath}
\usepackage{mathptmx}
\usepackage{amssymb}
\usepackage{placeins}
\usepackage{longtable}
\usepackage{caption}
\usepackage{enumitem}
\usepackage{array,booktabs,tabularx,ragged2e,multirow}
\usepackage[protrusion=false]{microtype}
\usepackage{refcount}
\usepackage{etoolbox}
\usepackage{float}
\usepackage{listings}
\usepackage{tikz}
\usetikzlibrary{positioning,arrows.meta,calc,fit,decorations.pathreplacing,patterns,backgrounds}
\usepackage{xurl}
\usepackage{lineno}
\newcommand*\patchAmsMathEnvironmentForLineno[1]{%
  \expandafter\let\csname old#1\expandafter\endcsname\csname #1\endcsname
  \expandafter\let\csname oldend#1\expandafter\endcsname\csname end#1\endcsname
  \renewenvironment{#1}{\linenomath\csname old#1\endcsname}{\csname oldend#1\endcsname\endlinenomath}}
\newcommand*\patchBothAmsMathEnvironmentsForLineno[1]{\patchAmsMathEnvironmentForLineno{#1}\patchAmsMathEnvironmentForLineno{#1*}}
\AtBeginDocument{\patchBothAmsMathEnvironmentsForLineno{equation}\patchBothAmsMathEnvironmentsForLineno{align}}
\usepackage[colorlinks=true]{hyperref}
\hypersetup{pdftitle={A general-purpose System One decision model for clinical records},
  citecolor=citeblue,linkcolor=citeblue,urlcolor=citeblue}
\usepackage[capitalise,noabbrev]{cleveref}
\crefname{figure}{Fig.}{Figs.}\Crefname{figure}{Fig.}{Figs.}

\definecolor{citeblue}{HTML}{1A5DAB}
\definecolor{todo}{HTML}{B0361F}
\makeatletter
\def\@cite#1#2{\textsuperscript{\textcolor{citeblue}{#1\if@tempswa , #2\fi}}}
\makeatother

\makeatletter\patchcmd{\@maketitle}{\noindent\sloppy}{\noindent\raggedright}{}{}\makeatother%
\patchcmd{\thebibliography}{\section*{\refname}}%
  {\par\addvspace{14pt}\setlength{\parskip}{2pt}\global\let\refsbookmark\dorefsbookmark}{}%
  {\PackageWarningNoLine{template}{Could not remove the reference-list heading}}
\makeatletter\def\@openbib@code{\setlength{\itemsep}{0pt}\setlength{\parsep}{1.0pt}}\makeatother
\let\refsbookmark\relax
\newcommand*{\dorefsbookmark}{\pdfbookmark[0]{References}{mainrefs}\global\let\refsbookmark\relax}
\makeatletter
\AtBeginDocument{\let\refs@bibitem\@bibitem\def\@bibitem#1{\refs@bibitem{#1}\refsbookmark\ignorespaces}}
\makeatother

\setlist[itemize]{leftmargin=1.2em,itemsep=3pt,topsep=4pt,parsep=0pt,label=\textbullet}
\renewcommand{\figurename}{Fig.}
\DeclareCaptionLabelSeparator{bar}{\space\textbar\space}
\newcolumntype{L}[1]{>{\RaggedRight\arraybackslash}p{#1}}
\newcolumntype{Y}{>{\RaggedRight\arraybackslash}X}

\newcommand{\EDbegin}{%
  \clearpage
  \setcounter{figure}{0}\setcounter{table}{0}%
  \renewcommand{\figurename}{Extended Data Fig.}\renewcommand{\tablename}{Extended Data Table}%
  \def\theHfigure{ED\arabic{figure}}\def\theHtable{ED\arabic{table}}%
  \pdfbookmark[0]{Extended Data}{extendeddata}}

\definecolor{todocolor}{HTML}{B0361F}
\newcommand{\model}[1]{\textsf{\system-#1}}
\newcommand{\nd}{\texttt{not\_documented}}
\newcommand{\institution}{MGB}
\newcommand{\dataset}{Klaros-12M}
\newcommand{\testset}{MGBQA-1M}
\newcommand{\system}{CS1}
\newcommand{\mainmodel}{CS1-4B}
\newcommand{\CSoneEICUTimelines}{92.8}
\newcommand{\CSoneEHRSHOT}{94.3}
\newcommand{\CSoneEHRSHOTPrognosis}{0.766}
\newcommand{\CSoneNsets}{21}
\newcommand{\CSoneAboveOpenN}{5}
\newcommand{\CSonePooledECE}{0.029}
\newcommand{\CSoneECEMin}{0.007}
\newcommand{\CSoneECEMax}{0.100}
\newcommand{\CSoneMsOne}{683}
\newcommand{\CSoneMsTen}{955}
\newcommand{\CSoneMsForty}{1,801}
\newcommand{\CSoneSecForty}{1.80}
\newcommand{\CSoneAnswersPerSec}{22.2}
\newcommand{\CSoneGBHours}{500}
\newcommand{\CSoneSpeedupTwentySeven}{6.4}
\newcommand{\CSoneSpeedupVega}{6.4}
\newcommand{\CSoneTimingIQRMax}{4.2}
\newcommand{\CSoneNotesPerHour}{2,000}
\newcommand{\EXTscoredPatients}{19,027}

\makeatletter
\newcommand{\EDtenpt}{%
  \renewcommand\normalsize{\@setfontsize\normalsize\@xpt\@xiipt
    \abovedisplayskip 10\p@ \@plus2\p@ \@minus5\p@
    \abovedisplayshortskip \z@ \@plus3\p@
    \belowdisplayshortskip 6\p@ \@plus3\p@ \@minus3\p@
    \belowdisplayskip \abovedisplayskip
    \let\@listi\@listI}%
  \renewcommand\small{\@setfontsize\small\@ixpt{11}%
    \abovedisplayskip 8.5\p@ \@plus3\p@ \@minus4\p@
    \abovedisplayshortskip \z@ \@plus2\p@
    \belowdisplayshortskip 4\p@ \@plus2\p@ \@minus2\p@
    \def\@listi{\leftmargin\leftmargini \topsep 4\p@ \@plus2\p@ \@minus2\p@
                \parsep 2\p@ \@plus\p@ \@minus\p@ \itemsep \parsep}%
    \belowdisplayskip \abovedisplayskip}%
  \renewcommand\footnotesize{\@setfontsize\footnotesize\@viiipt{9.5}%
    \abovedisplayskip 6\p@ \@plus2\p@ \@minus4\p@
    \abovedisplayshortskip \z@ \@plus\p@
    \belowdisplayshortskip 3\p@ \@plus\p@ \@minus2\p@
    \def\@listi{\leftmargin\leftmargini \topsep 3\p@ \@plus\p@ \@minus\p@
                \parsep 2\p@ \@plus\p@ \@minus\p@ \itemsep \parsep}%
    \belowdisplayskip \abovedisplayskip}%
  \renewcommand\scriptsize{\@setfontsize\scriptsize\@viipt\@viiipt}%
  \renewcommand\tiny{\@setfontsize\tiny\@vpt\@vipt}%
  \renewcommand\large{\@setfontsize\large\@xiipt{14}}%
  \renewcommand\Large{\@setfontsize\Large\@xivpt{18}}%
  \renewcommand\LARGE{\@setfontsize\LARGE\@xviipt{22}}%
  \renewcommand\huge{\@setfontsize\huge\@xxpt{25}}%
  \renewcommand\Huge{\@setfontsize\Huge\@xxvpt{30}}%
  \def\@listI{\leftmargin\leftmargini \parsep 4\p@ \@plus2\p@ \@minus\p@
              \topsep 8\p@ \@plus2\p@ \@minus4\p@ \itemsep4\p@ \@plus2\p@ \@minus\p@}%
  \let\@listi\@listI
  \setlength\parindent{15\p@}\setlength\topskip{10\p@}\setlength\maxdepth{.5\topskip}%
  \setlength\footnotesep{6.65\p@}%
  \setlength\smallskipamount{3\p@ \@plus 1\p@ \@minus 1\p@}%
  \setlength\medskipamount{6\p@ \@plus 2\p@ \@minus 2\p@}%
  \setlength\bigskipamount{12\p@ \@plus 4\p@ \@minus 4\p@}%
  \normalsize}
\makeatother
\newcommand{\EDsetup}{%
  \EDbegin
  \EDtenpt
  \newcommand{\hd}[2]{\begin{tabular}[b]{@{}c@{}}##1\\##2\end{tabular}}%
  \ifdefined\nolinenumbers\nolinenumbers\fi
  \sloppy}

\begin{document}
\title{A general-purpose System One decision model for clinical records}
\author{\raggedright Andrew~Zhang\textsuperscript{1,2,3}, Trinh~T.L.~Vuong\textsuperscript{1,2}, Sophia~J.~Wagner\textsuperscript{1,2}, Dongmin~Bang\textsuperscript{1,2}, Tong~Ding\textsuperscript{1,2}, Yuchang~Su\textsuperscript{1,2}, Mary~Cipperman\textsuperscript{1,2}, Christina~S.~Chen\textsuperscript{1,2}, Judith~Lefkes\textsuperscript{1,2}, Ramin~Kahidi\textsuperscript{1,2,3}, Ming~Y.~Lu\textsuperscript{1,2}, Caiwei~Tian\textsuperscript{1,2}, Tianhao~Luo\textsuperscript{1,2}, Jingnan~Ma\textsuperscript{1,2}, Rowland~Pettit\textsuperscript{1,2}, Long~Phi~Le\textsuperscript{1,2,3,\textdagger} \& Faisal~Mahmood\textsuperscript{1,2,\textdagger,*}\endgraf}
\maketitle
\begin{affiliations}
\item Mass General Brigham AI Institute, Boston, MA.
\item Harvard Medical School, Boston, MA.
\item Massachusetts Institute of Technology, Cambridge, MA.
\end{affiliations}
\vspace{-6pt}\noindent{\slshape\textsuperscript{\textdagger}Co-senior authors.\\
\textsuperscript{*}Corresponding author: Faisal Mahmood (\href{mailto:faisalmahmood@bwh.harvard.edu}{faisalmahmood@bwh.harvard.edu}).}\par\vspace{4pt}

\begin{abstract}
Clinical care, health-system operations and biomedical discovery depend on extracting reliable answers from patient records, yet applying diverse clinical questions across entire populations remains difficult and computationally expensive. Task-specific models offer limited flexibility, whereas generative language models require costly, repeated processing of lengthy records. Here we present Clinical System One (\system{}), a 4-billion-parameter decision model for fast, typed judgments from clinical records, along with two agentic harnesses in which \system{} works together with a generalist language model for high-volume screening. \mainmodel{} is trained on 12.4 million questions and answers from 3.6 million patients across a multi-hospital health system. Training combines structured-data keys, observed outcomes and language-model supervision to support questions across clinical notes, diagnostic reports and longitudinal patient timelines. \mainmodel{} answers multiple questions with predefined response options in a single forward pass and accommodates diverse questions and criteria without task-specific retraining. Each answer carries a probability that can route uncertain cases to human review, and on-demand attribution to the individual sentences it rests on. On 963,408 held-out patients, \mainmodel{} outperformed open decision models up to 27 billion parameters on 13 of 14 tasks scored independently of its training teacher. Across 21 external data sets, its aggregate performance approached that of the strongest tested 27-billion-parameter model, with 6.4-fold faster inference in a matched workload. A single desktop graphics processor reads about 2,000 notes per hour at 40 questions each, allowing deployment on modest local hardware. Integrating \mainmodel{} with a language model agent extended these capabilities from individual documents to longitudinal chart review and population screening: the combined system produced cited answers over complete multi-year patient histories and screened a patient population against the eligibility criteria of 17 registered clinical trials. In the retrospective trial-screening analysis, this approach placed four times as many judged-eligible patients among its top 20 candidates as keyword search and reduced language-model token use 690-fold relative to a language model reading every record. These findings support a division of labour in clinical artificial intelligence, with small decision models performing repeated record reading and generative models formulating questions and synthesizing evidence, and provide a foundation for continuously queryable clinical records supporting registry abstraction, quality measurement, cohort discovery, trial screening and clinical review.
\end{abstract}
\vspace{4pt}
\newpage
\section*{}
\vspace{-2.5em}
Clinical records support a large volume of fast, narrow judgements. Trial teams and sponsors screen thousands of candidates against dozens of eligibility criteria\cite{jin2024trialgpt,penberthy2012effort}. Registries and cohort studies ask one phenotype question of every patient in a health system\cite{banda2018advances}. Payers and coding teams review whether the record supports what was billed or requested, quality programmes audit every discharge for the same measures\cite{casalino2016quality}, and clinicians reviewing a chart ask how a problem evolved over years of notes\cite{overhage2020physician}. Each question is simple for a clinician who is familiar with the patient context, but across whole populations or long records, the cost quickly balloons beyond a feasible effort threshold. In Kahneman's terms these are System~1 judgements, fast and typed, as opposed to the deliberate System~2 reasoning of diagnosis and treatment planning\cite{kahneman2011thinking}.

Established automated workflows do not fit this workload. Clinical natural-language processing pipelines\cite{savova2010ctakes,chapman2001negex} and supervised phenotyping or risk models\cite{gehrmann2018comparing,rajkomar2018scalable} are fast but fixed to the concepts they were built for. Large language models (LLMs) accept questions in natural language\cite{agrawal2022large,vanveen2024adapted,singhal2023large,moor2023foundation}, but every question is a separate generation, often with explicit reasoning\cite{wei2022chain}, so cost grows with patients $\times$ questions $\times$ record length. Caching a shared record prefix\cite{kwon2023efficient,zheng2024sglang} or asking several questions in one prompt\cite{cheng2023batch} reduces the repeated reading, but every answer is still generated token by token. Their answers come without calibrated probabilities that could route uncertain cases to a human for review\cite{jiang2021can,griot2025metacognition}, and rarely with evidence a reviewer can check without rereading the chart\cite{wu2025sourcecheckup}. Data-use agreements for clinical text also often rule out hosted models\cite{physionet2023llm,physionet2025llm}, limiting local usage to small LLMs that have lower accuracy and higher hallucination rates.

A new class of open models targets exactly this interface. ``System One'' decision models take a state and a set of typed questions, and return an independent probability distribution for every question in one forward pass\cite{typesafe2026jev}. JEV-27B\cite{autotrust2026jev27b}, Clef\cite{cloudflare2026clef} and the Decision 2.0 family of 0.6--27-billion-parameter models\cite{vllmsr2026decision20} are publicly available. They are trained on public and synthetic data, however, and none reports an evaluation on clinical records. An independent medical evaluation, of the hosted Jev service on examination and case-report benchmarks, found it fast and accurate on literature questions but overconfident and behind a reasoning LLM on complex diagnosis, and recommended verifying and recalibrating its probabilities for every task\cite{madridgarcia2026jevmedicine}. A second study tested Jev as a judge of the factual accuracy of generated radiology reports\cite{huang2026jevradiology}, and small open decision models for medicine have begun to appear\cite{lighteternal2026biodecision,lionai2026meddecider,ma2026medjev}. A team that wants to screen a population or review charts with such a model needs to know whether it reads real records reliably, whether its probabilities can decide which cases need review, and whether its answers can be checked.

The distribution of a patient population in a real hospital system may look very different from the curated clinical data sets that are publicly available. Institutional clinical records can supply what public corpora cannot: decades of real patient trajectories and clinical text. Structured data give the exact answer to many questions about a record, and longitudinal follow-up gives the observed outcome of every prediction. General-purpose decision models have yet to be thoroughly evaluated and adapted for clinical question answering on real patient records. Even if models trained on a health system's own records predict outcomes well\cite{rajkomar2018scalable,jiang2023health,steinberg2021clmbr}, they may fail to transfer to other sites or remain calibrated there\cite{wong2021external,futoma2020myth,zech2018variable,wornow2023shaky}. In order to become trusted for real-world use, clinical models must be evaluated not only on held-out patient data from within a healthcare system (to assess its performance if deployed locally), but also show generalization to other hospitals with different patient populations.

Here we present Clinical System One (\mainmodel{}), a System One decision model for clinical records trained on \dataset{}: more than 12 million decisions about 3.6 million patients at \institution{}, a multi-hospital US academic health system (\cref{fig:overview}). We evaluate \mainmodel{} from single decisions to whole populations. We first establish that \mainmodel{} makes individual decisions reliably and efficiently, by assessing its accuracy and calibration on held-out \institution{} patients (\testset{}) and on \CSoneNsets{} public data sets from other institutions, against 13 publicly available decision models ranging from 0.6B to 27B parameters. We then show that each of \system{}'s answers about a note or patient timeline can be traced to the passages it rests on, forming the basis of our evidence attribution pipeline. Finally, we show that a generalist LLM using \system{} as a tool can answer questions about a whole patient record with citations, and show that a similar agentic system can screen a large patient population for eligibility using real clinical trial criteria at a small fraction of the cost it would take for an LLM to review each record.

\section*{Results}

\subsection{Clinical System One (\system{}) design and training.}

\begin{figure}[p]
\centering
\figinclude[width=0.98\linewidth]{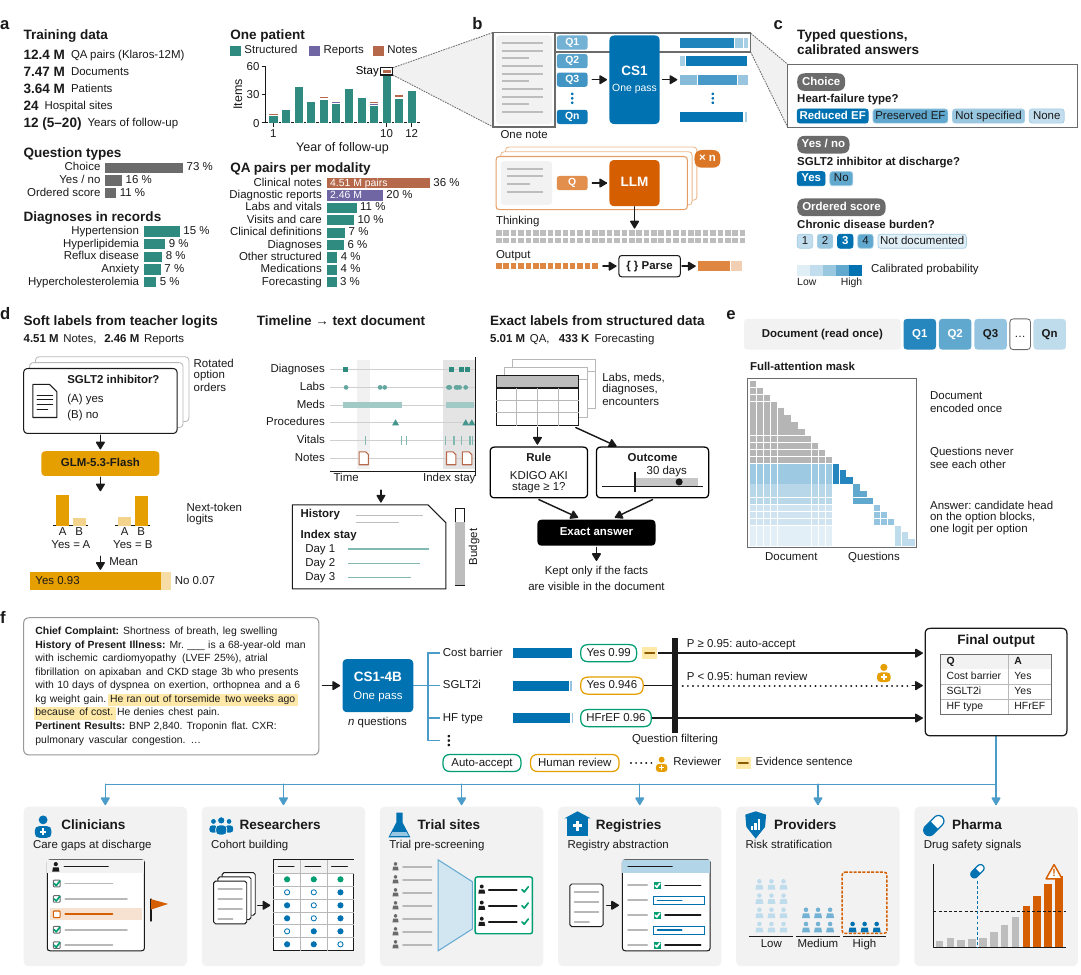}
\caption{\textbf{Clinical System One reads a clinical record once and answers many typed questions with calibrated probabilities.} \textbf{a}, The \dataset{} training data: 12.4 million question--answer (QA) pairs from 3.64 million patients at 24 hospital sites (median follow-up 12 years, interquartile range 5--20), asked of 7.47 million documents. One patient, items per year for a synthetic patient whose totals equal the cohort medians. Below, the share of pairs by question type and by modality (colour, input type, as in \cref{fig:performance}a), and the five most frequent conditions in the 3.01 million structured training records. \textbf{b}, \system{} answers any number of typed questions about a document in one forward pass, whereas a generative LLM can write hundreds of tokens per question. \textbf{c}, The three question types, each option shaded by \mainmodel{}'s probability (darkest, the answer). \textbf{d}, Ground truth. Left, soft labels for 4.51 million note and 2.46 million report pairs: the GLM-5.3-Flash teacher's next-token probabilities, averaged over rotated option orders. Middle, a record timeline serialized into a budgeted text document. Right, exact labels for 5.01 million timeline and 433,000 forecasting pairs from rules or observed outcomes, kept only when the facts are visible in the document. \textbf{e}, Shared-document packing: the document is read once and each question attends only to the document and itself; the candidate head returns one logit per option. \mainmodel{} has 4.2 billion parameters and was initialized from Decision 2.0 Nox-4B. \textbf{f}, Top, one synthetic note read once by \mainmodel{}: answers at $P\ge0.95$ are accepted automatically and the rest go to a reviewer; highlighted, the evidence for the cost question. Bottom, the same output serves six user groups. Patient examples are synthetic.}
\label{fig:overview}
\end{figure}

\system{} reads one document with any number of typed questions and returns a probability distribution over each question's options (\cref{fig:overview}b,c). Questions (choices among named options, ordered scores or yes/no claims, each option with a criterion for boundary cases) are written at request time, so a new criterion, phenotype or outcome definition needs no retraining. Documents are clinical notes, diagnostic reports, multi-note records and structured patient timelines rendered as text up to a decision point (Methods).

\mainmodel{} was initialized from Decision 2.0 Nox-4B\cite{vllmsr2026decision20} and trained on 12,416,448 document--question pairs from 3,643,798 \institution{} patient records (\cref{fig:overview}a and Extended Data Table~1). Its labels are of three kinds (\cref{fig:overview}d): option probabilities read out of an LLM teacher, GLM-5.3-Flash\cite{zai2026glm53flash}, on institutional hardware (Extended Data Fig.~1); exact answers computed by rules from structured data; and outcomes observed after the decision point, such as readmission. Its fine-tuning used no records from other institutions and no public data set.

The document is encoded once, and each question is a short segment that attends to the document and to itself but never to another question, so each answer equals that of a single-question prompt and an extra question costs only its own tokens (\cref{fig:overview}e and Extended Data Fig.~2). On one NVIDIA GB10 desktop graphics processor (commonly referred to as a DGX Spark), \mainmodel{} answered 40 questions about a 2,900-token note in \CSoneSecForty\,s, against 11.54\,s for Decision 2.0 Vega-27B (\CSoneSpeedupVega{} times faster); the Decision 2.0 models were run through their released inference code and not further optimized (Extended Data Table~2 and Methods). The speed-ups held for documents of 512 to 16,384 tokens, and show the efficiency of asking multiple questions in a single pass: reading the document once per question would have taken \mainmodel{} 15 times longer (Extended Data Fig.~3).

\subsection{Performance on held-out patients.}

\begin{figure}[p]
\centering
\figinclude[width=0.98\linewidth]{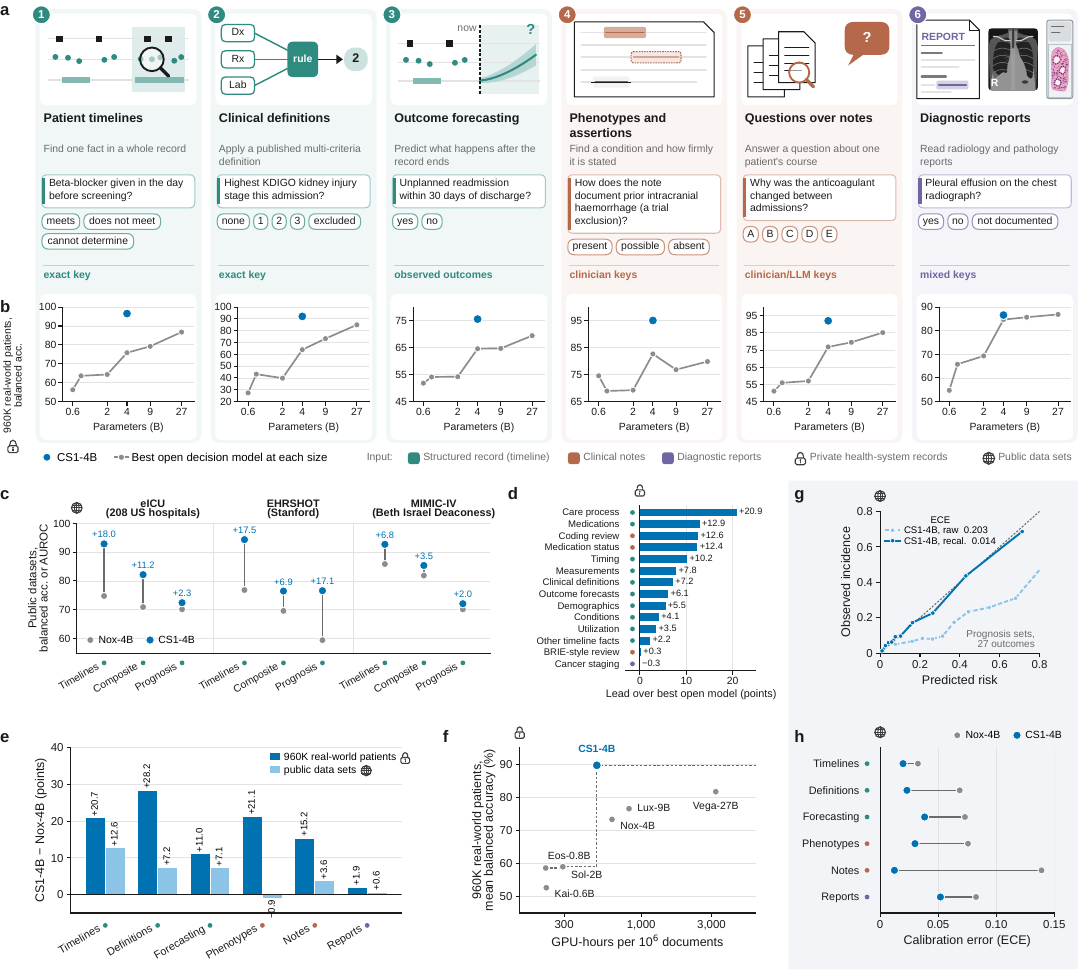}
\caption{\textbf{Performance on held-out patient records and transfer to public data sets.} \textbf{a}, Six task types, with example questions and the source of their answer keys (Extended Data Table~4); colour, input type. Padlock, private health-system records, never public; globe, public data sets, whose records may be in open models' training data. \textbf{b}, \testset{} (963,408 held-out patients): mean balanced accuracy over each task type's tasks keyed by rules, observed outcomes or an open LLM (14 of 28 tasks; Extended Data Table~3) against model size; blue, \mainmodel{}; grey, the best of the ten open decision models at each size. \textbf{c}, The structured-record public sets by source institution: Nox-4B (grey) and \mainmodel{} (blue), with their difference (points); balanced accuracy (\%), prognosis mean AUROC $\times$100. Other public sets and every open model, Extended Data Table~4. \textbf{d}, \mainmodel{} minus the best open model on each of the 14 tasks of b (balanced-accuracy points). \textbf{e}, \mainmodel{} minus Nox-4B per task type, on \testset{} and on the \CSoneNsets{} public data sets. \textbf{f}, Mean of b over the six task types against GPU-hours per million documents at 40 questions on one GB10 (Extended Data Table~2); dashed, Pareto frontier. \textbf{g}, The three public prognosis sets (27 outcomes): observed incidence against predicted risk, raw and after one log-odds shift per outcome to its incidence in the other half of the patients. \textbf{h}, Expected calibration error (ECE) of the top answer per task type, pooled over the public data sets. Dots in c--e and h, task type. \mainmodel{}: fine-tuned on \dataset{} only, raw probabilities unless stated.}
\label{fig:performance}
\end{figure}

On real patient records, we compared \mainmodel{} with ten open decision models: the six Decision 2.0 models, JEV-27B, Clef, pplx-decider-v1-27b\cite{perplexity2026pplxdecider} and BioDecision-4B\cite{lighteternal2026biodecision}; MedDecider-9B and -27B\cite{lionai2026meddecider} were evaluated on the public data sets only (Methods). Each baseline model was run through its official implementation with a question input format that was optimized for that model. Tasks fall into six task types, three read from structured records and three from text (\cref{fig:performance}a). Because always giving the most common answer scores well in raw accuracy on several tasks, the primary metric is balanced accuracy, the mean recall over answer classes.

On \testset{} (28 tasks; 963,408 held-out \institution{} patients, one question each; Extended Data Table~3), \mainmodel{} was the most accurate decision model in five of the six task types (\cref{fig:performance}b). It led on 13 of the 14 tasks whose keys come from rules, observed outcomes or an open LLM: it reached 96.6\% averaged over the eight timeline tasks, against 86.8\% for JEV-27B, the best open model there; 92.1\% on the clinical-definition task (Vega-27B 85.0\%); 75.6\% on outcome forecasting (JEV-27B 69.4\%); and 95.0\% on medication status (BioDecision-4B 82.6\%). The exception was cancer staging from diagnostic reports, where pplx-decider-v1-27b reached 86.9\% against 86.6\%. Averaged over the six task types, \mainmodel{} scored 89.7, well above its initialization, Nox-4B (73.3), and above Vega-27B (81.7), the best open model (\cref{fig:performance}b). \mainmodel{} thus read the real-world care trajectories of held-out patients accurately, whereas the open models generalized poorly to institutional data. \testset{} is also the only evaluation in this study whose records and keys no open model could have seen, because they are restricted to institutional systems. Among the 14 tasks keyed by rules, outcomes or an open LLM, the gap between \mainmodel{} and the best open model was largest on tasks keyed by \institution{}'s own structured data and conventions (care process, 20.9 points; medications, 12.9; coding review, 12.6; medication status, 12.4) and smallest on the tasks that most resemble public benchmarks (BRIE-style chart review, 0.3; cancer staging, $-$0.3; \cref{fig:performance}d), which points to the value of training on real-world patient trajectories and use cases.

\subsection{Generalization to external data sets.}

We evaluated the ability of \mainmodel{} to transfer its institutionally trained knowledge to public data sets. The \CSoneNsets{} public data sets come from other institutions: timelines rendered from the structured records of MIMIC-IV\cite{johnson2023mimiciv,johnson2023mimicivnote}, eICU\cite{pollard2018eicu} and EHRSHOT\cite{wornow2023ehrshot}, and public notes and reports (\cref{fig:performance}a,c, Extended Data Table~4 and Methods). Averaged over the six task types, \mainmodel{} scored 86.7, against 81.7 for its initialization, Nox-4B, and 88.6 for Vega-27B, the strongest open model, while answering \CSoneSpeedupVega{} times faster; it was above Vega-27B on patient timelines, outcome forecasting and questions over notes. A 27-billion-parameter model was the best open model on 20 of the \CSoneNsets{} sets.

\mainmodel{} exceeded every open model, even the 27-billion-parameter models, on \CSoneAboveOpenN{} sets, most clearly on structured records: \CSoneEICUTimelines\% on eICU-Timelines and \CSoneEHRSHOT\% on EHRSHOT-Timelines, against 91.1\% and 90.5\% for JEV-27B, the best open model on both, and a mean AUROC of \CSoneEHRSHOTPrognosis{} on EHRSHOT-Prognosis, the best outcome ranking of any system (Clef, 0.681). It trailed the best open model by 7.4--12.8 points on the three clinical-definition sets, ranked outcomes below every 27-billion-parameter model on MIMIC- and eICU-Prognosis, and, of the eleven sets read from notes and reports, led narrowly on MIPA and BRIE but trailed on the other nine, by 1.1--5.9 points. These leads are not an artefact of its training layout: on MIMIC-IV timelines rendered both ways, \mainmodel{}, like every other model, scored higher in the encounter layout it was not trained on (Extended Data Table~5). Training on \institution{} records also lifted \mainmodel{} over its initialization on these sets, by 12.6 points on patient timelines, 7.1--7.2 points on clinical definitions and outcome forecasting, and $-$0.9 to 3.6 points on the three task types read from notes and reports (\cref{fig:performance}c,e).

Because the training data of most open models are undisclosed, we cannot exclude that the published benchmarks, or the records on which any of the sets are based, were seen in their training\cite{sainz2023nlp}, so their scores may partly reflect exposure to the underlying records. Thus the public datasets primarily serve to show how \mainmodel{}'s finetuning on internal \institution{} data transfers to improved performance over Nox-4B at other institutions, and are less useful as a direct comparator between \mainmodel{} and other public models.

\subsection{Calibrated probabilities at population scale.}

At population scale, probabilities decide which answers are accepted automatically and which a person reviews (\cref{fig:overview}f). \mainmodel{} results are first reported without an explicit calibration stage on validation data. On the public data sets, the expected calibration error (ECE) of its top answer was 0.012--0.052 across the six task types, below that of its initialization, Nox-4B, in every task type (0.033--0.139; \cref{fig:performance}h). Pooled over five human-keyed public benchmarks (14,756 items), its ECE was \CSonePooledECE{}, about half that of Nox-4B (0.058), but above that of JEV-27B (0.010) and Clef (0.024), and per set it ranged from \CSoneECEMin{} to \CSoneECEMax{} (Extended Data Table~2). Its outcome forecasts rank patients effectively, but absolute risks did not transfer: over the 27 outcomes of the three public prognosis sets, its mean predicted risk was 0.39 against an observed incidence of 0.19, and the calibration error of its predicted risks was 0.203 (\cref{fig:performance}g), as expected where intensive-care or emergency cohorts have baseline incidences unlike those of a general hospital population. Within each outcome, however, its predictions had about the right spread (median calibration slope 1.03), so one correction per outcome sufficed: shifting each outcome's predicted risks on the log-odds scale to match its incidence\cite{saerens2002adjusting}, estimated on the other half of the patients, reduced the calibration error to 0.014 and the Brier score from 0.182 to 0.110, below that of Nox-4B after the same correction (0.118). A site that knows the incidence of its outcomes can thus read \mainmodel{}'s forecasts as risks without retraining.

Because the document is encoded once, cost grows with the number of documents rather than with documents $\times$ questions. At 40 questions per document, \mainmodel{} costs \CSoneGBHours{} GB10-hours per million documents, against 630 for Nox-4B, 830 for Lux-9B and 3,210 for Vega-27B (Extended Data Table~2). On \testset{}, \mainmodel{} was more accurate than every Decision 2.0 model and cheaper than those of 4 billion parameters or more (\cref{fig:performance}f); the three smaller ones cost 230--290 GB10-hours but were less accurate. One desktop graphics processor thus reads about \CSoneNotesPerHour{} notes, 40 questions each, per hour.

\subsection{Sentence-level evidence attribution.}%

\begin{figure}[p]
\centering
\figinclude[width=0.98\linewidth]{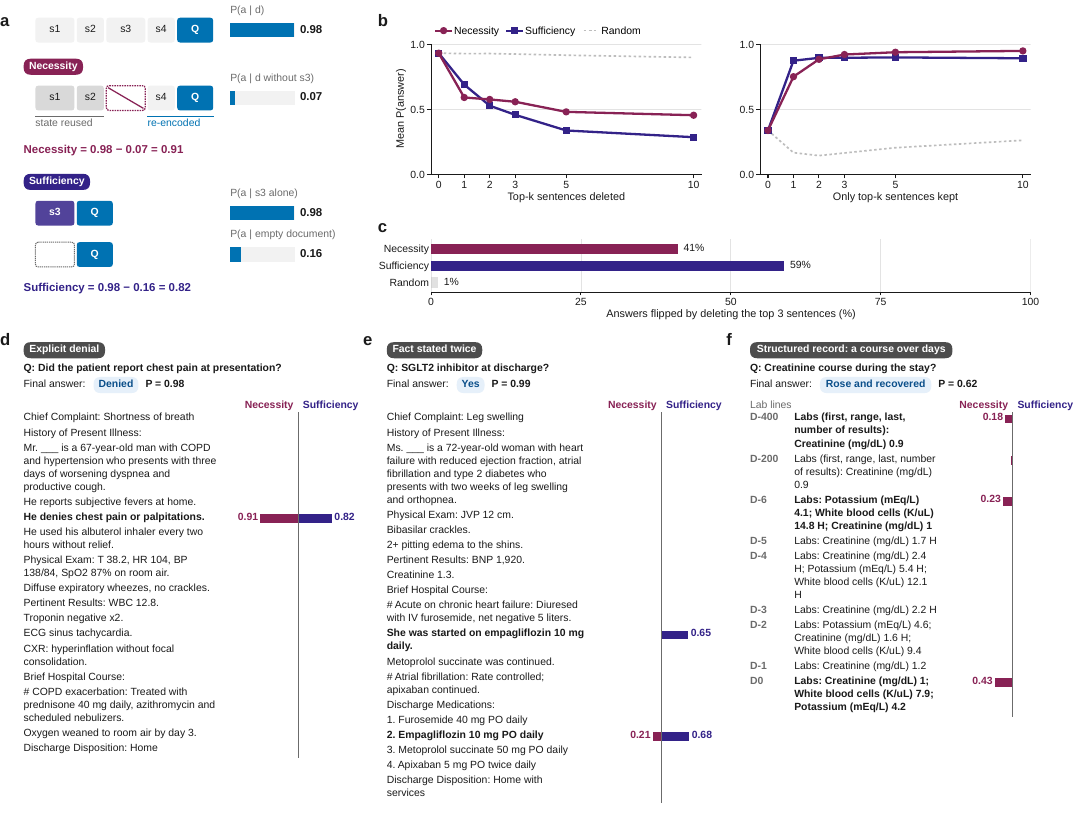}
\caption{\textbf{\system{} provides evidence attribution over notes and timelines.} \textbf{a}, Deletion attribution: a sentence's necessity is $P(a\mid d)-P(a\mid d\setminus s)$, the drop in the probability of the answer $a$ when the sentence is deleted and the document is re-encoded; its sufficiency is $P(a\mid s\text{ alone})-P(a\mid\text{empty document})$; the bars and numbers are those of the denial sentence in d. Re-encoding reuses the model's state for the text before the deleted sentence; long documents are screened in 10-sentence chunks before sentences are measured individually in the top chunks. \textbf{b}, Faithfulness on the 90 evidence-dependent note--question pairs (answer probability below 0.5 for an empty document) of 200 pairs from 100 MIMIC-IV discharge notes not used to train \mainmodel{} (aggregate only): mean probability of the original answer after deleting the top-$k$ sentences by necessity or sufficiency (lower is more faithful) or after keeping only them (higher is more faithful); dotted, random order. Median time per answer on one GB10: 24\,s (necessity) and 10\,s (sufficiency). \textbf{c}, Answers flipped by deleting the top three sentences of the same pairs: 41\% (necessity) and 59\% (sufficiency), against 1\% for random order. \textbf{d}--\textbf{f}, Evidence in two synthetic notes and a synthetic structured record: for each sentence or line, its necessity (left) and sufficiency (right), as defined in a; values below zero (a sentence that lowers the answer's probability) are drawn as zero. \textbf{d}, An explicit denial: deleting one sentence drops the answer from 0.98 to 0.07. \textbf{e}, A fact stated twice: deleting either mention lowers it by at most 0.21, while each alone gives 0.97 or 0.94. \textbf{f}, The laboratory lines of a day-by-day timeline, rendered as \system{} reads structured data: for the course of creatinine over the stay (rose and recovered, $P=0.62$), the evidence spans days. Deleting the last value (D0), the admission value (D$-$6) or the earlier outpatient value (D$-$400) drops $P$ to 0.19, 0.39 and 0.44, while no single line alone gives the answer. Faithfulness was measured on notes only. All text in d--f is synthetic.}
\label{fig:attribution}
\end{figure}

To identify which parts of a clinical record drive \system{}'s answers, we measured how the predicted probability changes when individual sentences are removed or considered in isolation\cite{deyoung2020eraser,cohenwang2024contextcite} (\cref{fig:attribution}a).
We used two complementary measures, necessity and sufficiency. Necessity is defined as the decrease in probability when a sentence is removed from the record; sufficiency, in contrast, is defined as the increase in probability when that sentence is presented alone, relative to an empty document.
Together, these measures distinguish evidence that is essential to an answer from evidence that independently supports it, even when the same clinical fact appears in multiple places.

We quantified the faithfulness of the attributions on the 90 note--question pairs whose answer depends on the note (probability below 0.5 for an empty document), among 200 pairs from 100 MIMIC-IV discharge notes not used in training (\cref{fig:attribution}b,c).
Removing the three highest-ranked sentences changed the model's answer in 41\% of cases when ranked by necessity and 59\% when ranked by sufficiency, compared with 1\% for randomly selected sentences. Conversely, retaining only the highest-ranked sentences recovered a mean probability of 0.89 for the original answer, compared with 0.19 for random selections.
These results indicate that the identified sentences have a substantially greater influence on the model's decisions than other, arbitrary parts of the record.
Attribution incurs additional computation, with median processing times of 24\,s per answer for necessity and 10\,s for sufficiency on a single NVIDIA GB10 machine. We therefore compute attribution on demand, allowing reviewers to inspect the sentences underlying selected decisions without incurring this cost for every answer in a population-scale screen.

We next examined attribution in synthetic notes with known supporting evidence (\cref{fig:attribution}d--f).
For a patient presenting with dyspnoea, \system{} identified an explicit denial of chest pain with probability 0.98. Removing the sentence documenting the denial reduced this probability to 0.07, identifying it as the principal evidence for the answer (\cref{fig:attribution}d). In a second example, empagliflozin was documented both in the hospital course and in the discharge medication list. Either mention independently supported the answer that the patient was prescribed an SGLT2 inhibitor at discharge, while removing either had little effect because the other remained (\cref{fig:attribution}e). More complex questions required evidence distributed across the record. For a structured patient timeline, creatinine rose from 0.9 to 2.4\,mg/dL before returning to 1.0\,mg/dL. The model identified this course as rising and recovering, with the largest contributions coming from laboratory measurements on different days rather than a single observation (\cref{fig:attribution}f).
Extended Data Fig.~4 shows further cases, including an answer resting on silence, and the limit of single-sentence attribution: evidence repeated across days makes no line necessary, and a pattern across days is carried by no single line.

\subsection{Question answering over longitudinal patient records.}

\begin{figure}[p]
\centering
\figinclude[width=\linewidth]{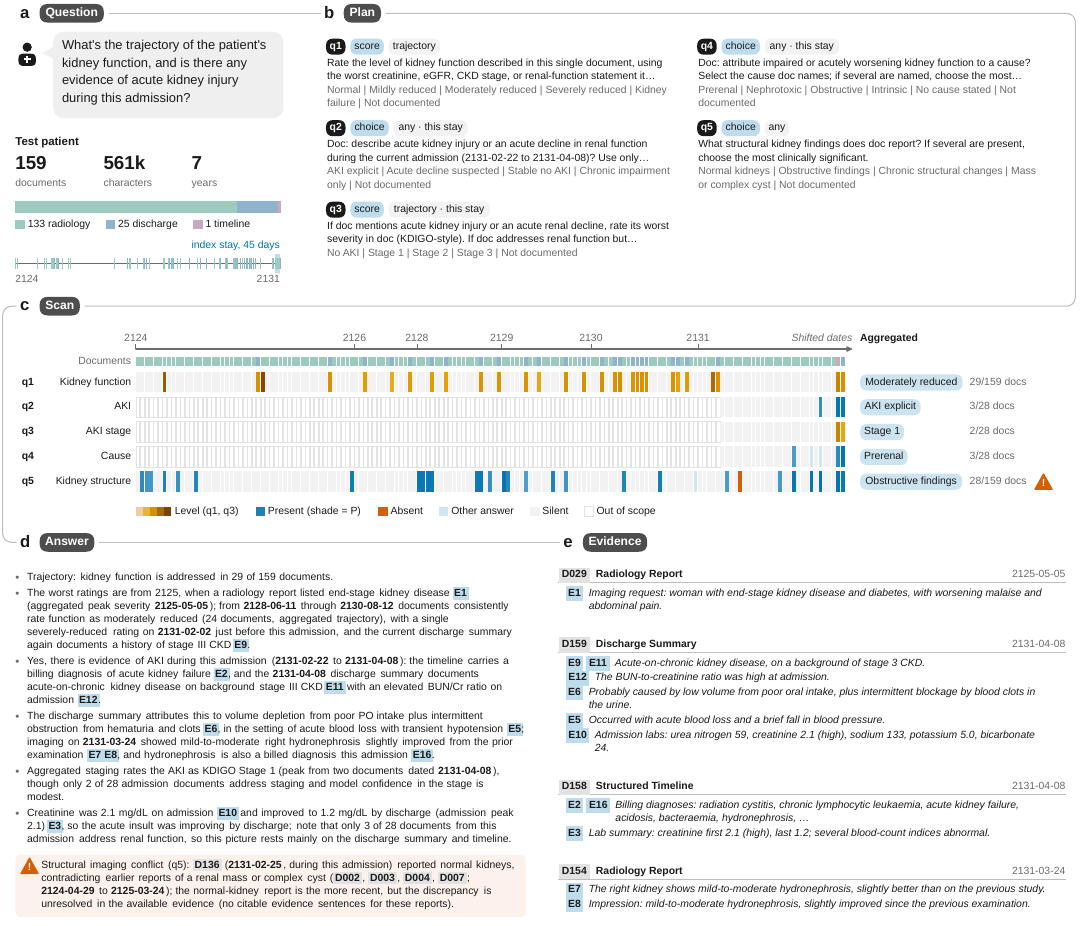}
\caption{\textbf{Evidence-backed answers over a whole patient record.} \textbf{a}, A clinical question and the summarized record of a MIMIC-IV test patient. \textbf{b}, An LLM (GLM-5.3) plans 5 typed sub-questions, each with an aggregation rule and a scope. \textbf{c}, \mainmodel{} answers every sub-question for every document (402 typed decisions; long documents chunked, nothing truncated); rows are labelled by topic. Level rows (trajectory and latest rules) show each document's level, light to dark in option order; the other rows show whether a document reports the finding (blue, shaded by probability), reports it absent (orange), gives another answer or is silent. Right, code aggregates each sub-question across documents and time: the aggregated answer and the documents that support it; a warning triangle marks conflicting documents, explained in d. \textbf{d}, The LLM composes a cited answer from the aggregated answers and the evidence sentences (one bullet per sentence). \textbf{e}, \system{} attribution selects 16 evidence sentences; the 12 cited in d are shown, grouped by document (a sentence cited twice shares a line). The LLM plans from a manifest of the record (document types, counts and dates) and never reads the record; at the last step it sees only the aggregated answers and the evidence sentences. Aggregation rules: any, the highest probability over documents; latest, the most recent document that addresses the sub-question; trajectory, the per-document level over time; a conflict is confident support and confident contradiction in different documents. Sub-questions scoped to this admission are asked only of its documents (white cells, out of scope). Evidence sentences are selected by \system{} sufficiency attribution; in compliance with the MIMIC-IV data-use policy, the evidence sentences and any record text quoted in the answer are shown as paraphrases, not verbatim. Dates are shifted by MIMIC-IV de-identification and documents are relabelled D001\ldots in date order. Results shown are from a single run, with GLM-5.3 at its default reasoning effort and \mainmodel{} (raw outputs).}
\label{fig:patientqa}
\end{figure}

Whole-record question answering is a routine but labour-intensive part of chart review\cite{overhage2020physician}. In order to answer a question about a patient's history, a reviewer must find every relevant note, report and structured result across years of care, and LLMs answering questions over long or whole records\cite{fleming2024medalign,adams2025longhealth} often omit information spread across documents\cite{cahoon2026brie}. General LLMs plan and write well but are too costly to read every record, whereas \system{} answers typed questions about one document at a time; combining typed questions by explicit rules has been proposed to bridge the two\cite{madridgarcia2026jevmedicine}. In order to address this bottleneck, we introduce a whole-record question-answering pipeline that pairs a general LLM, GLM-5.3\cite{zai2026glm53}, with \system{} as a tool\cite{yao2023react,schick2023toolformer}: rather than reading full records itself, the LLM plans and cites, while \mainmodel{} supplies every typed decision (\cref{fig:patientqa}). First, the LLM sees only a manifest of the record (document types, counts and dates) and plans typed sub-questions, each with an aggregation rule and a scope (\cref{fig:patientqa}a,b). Next, the harness answers every sub-question for every document, and code aggregates the answers across documents and time and flags confident disagreements between documents (\cref{fig:patientqa}c). Finally, the LLM composes an answer whose statements are cited to the sentences that \mainmodel{} answered from (\cref{fig:patientqa}d,e).

We first illustrate the pipeline on a MIMIC-IV test patient with 159 documents from 26 admissions over seven years. The question concerned the trajectory of kidney function and any acute kidney injury in the current admission. To answer it, the harness made 402 typed decisions, traced chronic kidney disease across the record, identified the acute kidney injury with its causes and recovery, and cited 12 sentences. Extended Data Figs.~5 and 6 show two further examples, one asked about anticoagulation and bleeding and one about the course of cancer. Because the LLM plans from the manifest and never reads the record itself, the LLM's cost does not grow with the length of the record, and every statement in the answer points to a bounded segment of the raw record.

To assess how well the pipeline answered, each answer was checked against its full record by GLM-5.3 at high reasoning effort\cite{zheng2023judging} and by reading short excerpts (Methods). Our three answers were correct in substance with minor errors (one of them due to a mistake in the raw record). In the kidney example, the worst level of kidney function came from the indication line of a radiology request that listed end-stage renal disease, which the rest of the record contradicts, and two statements repeated errors of the aggregated answers (a count of documents and a conflict built on cysts misread as masses). The anticoagulation answer omitted a left ventricular assist device, a further indication for anticoagulation. It also missed the documented site of the index admission's bleed and flagged a conflict with head CT reports that addressed only intracranial bleeding. The cancer answer stated progression more firmly than mixed recent findings support, listed a radiosensitizing course of cisplatin as systemic therapy and a bone marrow transplant that rests on a single billing code, and omitted several regimens. Most of these errors are easily flagged in the answer. The end-stage renal disease rests on a request's indication line and the bleeding conflict on head CT reports, so a reviewer can catch and correct them by opening a few sentences rather than rereading the whole record.

\begin{figure}[p]
\centering
\figinclude[width=0.96\linewidth]{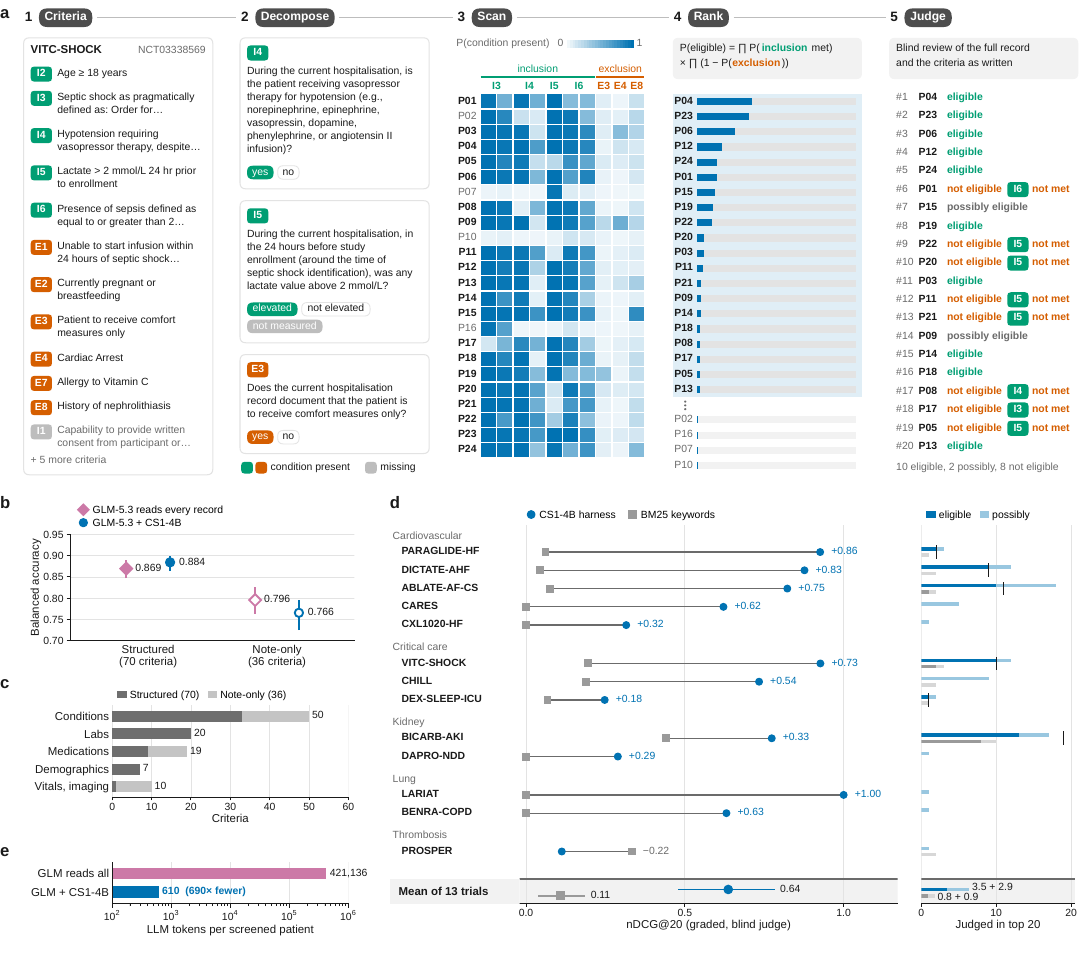}
\caption{\textbf{Population-level cohort building with real clinical trial criteria.} \textbf{a}, The pipeline on one of the 17 evaluated trials, VITC-SHOCK (NCT03338569\cite{wacker2022vitaminc}), and 1,000 MIMIC-IV test patients. (1) Eligibility criteria as written (green, inclusion; orange, exclusion; grey, not checkable from records; 12 of 17 shown, shortened). (2) An LLM (GLM-5.3) turns them into atomic typed questions, each with a time scope. (3) \mainmodel{} reads every document of every patient (21,866 documents); a question's probability is its maximum over the parts of the record within its scope. Shown are \mainmodel{}'s top 20 patients and four drawn at random from the rest (P01--P24). (4) Code combines the answers into P(eligible) and ranks the patients (shaded, top 20). (5) A blind LLM judge reading the full record graded the top 20: 10 eligible, 2 possibly eligible and 8 not eligible. \textbf{b}, Criterion-level balanced accuracy for seven trials on 200 patients, for the LLM reading every record and for the agentic harness (95\% intervals by patient bootstrap). Filled, 70 structured criteria with rule-computed answers: 0.869 against 0.884, a difference of 0.015 ($-$0.005 to 0.039). Open, 36 note-only criteria, against labels on which two full-record LLM reads agree: 0.796 against 0.766, a difference of $-$0.030 ($-$0.060 to $-$0.009). \textbf{c}, The criteria of b by topic (dark, structured; light, note-only). \textbf{d}, Ranking all 1,000 patients for each trial by \mainmodel{}'s P(eligible) and by BM25 keyword search over the full record. The union of both top 20s was graded by a blind GLM-5.3 judge in two passes (663 pairs; weighted $\kappa$\cite{cohen1968weighted} between passes 0.84). Left, nDCG@20\cite{jarvelin2002cumulated} for the 13 trials with any judged-eligible or possibly eligible patient, grouped by therapeutic area (numbers, \mainmodel{} minus BM25), and the mean with its 95\% interval: 0.64 against 0.11 (\mainmodel{} higher in 12 of 13 trials). Right, judged-eligible (dark) and possibly eligible (light) patients in each top 20; ticks, judged-eligible patients in the pool. \textbf{e}, LLM tokens per screened patient to rank patients for the 13 trials of d: 421,136 for the LLM reading every record against 610 with \mainmodel{}, 690 times fewer (verification, \mainmodel{}'s own compute and judge calls not counted).}
\label{fig:screening}
\end{figure}

\subsection{Clinical trial eligibility screening across a patient population.}

Screening is a primary bottleneck in trial development\cite{penberthy2012effort}. In order to ensure participant safety and reliable scientific results, researchers must check and rank every candidate against predefined eligibility criteria written in free-form text. This process can be tedious\cite{unlu2025manual}, especially given dozens of criteria and a database of thousands or millions of patients. LLMs that read clinical notes have screened patients for individual trials\cite{unlu2024rectifier,wornow2024zeroshot,syed2026translating} and, in a randomized trial, sped up prescreening\cite{unlu2025manual}, but they read each candidate's record for every trial. To solve this, we introduce an automated screening pipeline that accelerates and streamlines patient selection. We frame screening as a ranking problem: Given a list of trial criteria and a population of patients, a trial screening system must produce a ranking of the patients such that higher-ranked patients are more likely to be eligible. Our method uses an LLM (GLM-5.3) to translate a trial's free-text eligibility criteria into a series of atomic typed questions, each bound to a specific timeframe, much as earlier systems translated criteria into cohort queries\cite{yuan2019criteria2query}. \mainmodel{} then does high-volume question answering by scanning every medical document for every candidate. We aggregate these answers into a standardized eligibility score (\cref{fig:screening}a).

We tested this pipeline on 1,000 test patients from the MIMIC-IV database (21,866 medical documents), each screened for 17 trials from ClinicalTrials.gov\cite{zarin2011clinicaltrials}, and evaluated its accuracy at two levels: individual criteria matching (tested on 7 trials and 200 patients) and overall patient ranking (tested across the 13 trials whose judged pool contained an eligible or possibly eligible patient). For criteria that could be answered using rule-based calculations from structured data, our pipeline matched the raw LLM, which read every single record: our pipeline achieved a balanced accuracy of 0.884 compared to the LLM's 0.869 (difference 0.015, 95\% CI $-$0.005 to 0.039). For complex criteria that could only be answered using unstructured clinical notes, our pipeline performed slightly lower than the benchmark LLM (0.766 accuracy vs 0.796; difference $-$0.030, $-$0.060 to $-$0.009), using a reference standard adjudicated by the LLM itself (\cref{fig:screening}b,c).

To grade how well the pipeline ranked whole patients by multiple criteria, a blinded LLM judge\cite{zheng2023judging} reviewed the full medical records of the top 20 patients selected by each method across the 13 trials whose judged pool contained an eligible or possibly eligible patient. Performance was measured using nDCG@20 (Normalized Discounted Cumulative Gain)\cite{jarvelin2002cumulated}, a metric from 0 to 1 that rewards placing judged-eligible patients at the very top of the list; 1 is the best ordering of the judged patients, 0 is assigned to a top 20 without any. Our pipeline achieved a mean nDCG@20 of 0.64, significantly outperforming standard BM25 keyword search, which scored just 0.11. Our method beat keyword search in 12 out of the 13 trials (\cref{fig:screening}d), and on average placed 3.5 judged-eligible patients in its top 20 list per trial, compared to just 0.8 patients for the keyword search. These grades are judgements by the same LLM (GLM-5.3) that wrote the questions and keyed the note-only criteria, and LLM judges can favour their own outputs\cite{panickssery2024llm}; the weighted $\kappa$\cite{cohen1968weighted} of 0.84 between its two grading passes measures consistency, not agreement with clinicians.

Our pipeline is exceptionally cost-effective, using 690 times fewer LLM tokens per patient to rank candidates compared to the pure LLM approach, which required reading every record (\cref{fig:screening}e). Note that this token count excludes the internal processing of \mainmodel{}, which generated roughly 4.0 million answers by evaluating all 519 trial questions (17 trials) across 7,695 individual record units for the 1,000 patients. Because the benchmark LLM (which reads every record) is so expensive, we ran it as a baseline for 200 patients at the individual criterion level; ranking all 1,000 patients would be cost-prohibitive. Consequently, the final 1,000-patient ranking comparison was conducted against the much cheaper keyword search.

\section*{Discussion}

\system{} makes it practical to ask many narrow questions of every record in a large patient population. For users who read records at scale, this changes what is affordable. A trial team can have every candidate in a large population ranked within days, with an LLM reading only the criteria rather than every record. A clinician can ask a question about a patient's whole history and receive an answer whose every statement points to a sentence in the chart. A registry can extract structured definitions from clinical free-text without extensive manual curation; confident answers can be auto-accepted and uncertain ones can be routed to a reviewer.

These use cases are enabled by \system{}'s training on real-world patient records. On held-out patients of the multi-hospital \institution{} system, it was the most accurate decision model tested in five of six task types. On public data sets from other institutions, it led every open model on the structured timelines of eICU and EHRSHOT and on outcome forecasting in EHRSHOT, gained most over its initialization where records are structured, and had about half the calibration error of its initialization. Averaged over the six task types, it came within 2 points of the strongest 27-billion-parameter model, Vega-27B, which, at 27 billion parameters, has \CSoneSpeedupTwentySeven{} times the latency of \mainmodel{}.

For a team deciding whether to use \system{}, these results suggest the following. First, it is most reliable where the records are structured and the keys exact, and on the institution's own tasks; on most public note and report sets the larger open models were better, so criteria read from free text deserve more review. Second, its outcome forecasts rank patients well and can be easily recalibrated to the local incidence of each outcome. Third, the cost is low enough to ask many questions of every patient: a small desktop GPU reads about \CSoneNotesPerHour{} notes an hour at 40 questions each, and a more powerful machine would make this many times faster. Fourth, every answer can be traced to the sentences it rests on. Finally, \system{} composes effectively with general LLMs, which keep the flexibility of language while delegating population-scale reading to a small, fast decision model: given \system{} as a tool, an LLM answered questions about whole patient records with cited sentences and ranked patients for real trials while reading none of the records itself.

Rapid decision models such as \system{} are complementary to foundation models of the electronic health record\cite{kraljevic2024foresight,shmatko2025delphi}. Apollo\cite{zhang2026apollo}, trained on 25 billion events from 7.2 million patients of the same health system, compresses a patient's multimodal history into one representation that supports forecasting of disease onset, progression and treatment response, and similarity search across a population. Such a representation answers what is likely to happen to a patient and which patients resemble one another; each endpoint is read out by a model fitted for it, and attributions point to learned associations across events. \system{} answers a different kind of question: what the record states, posed in words at request time without task-specific training, with a calibrated probability and the sentences that support it. The two can work together. \system{} can turn criteria written in words into labels, cohorts and endpoints on which representation models are fitted and evaluated, and can check, sentence by sentence, the patients that a representation model forecasts to be at risk or retrieves as similar; a representation model can, in turn, shortlist the patients that \system{} reads.

The comparison also shows what open models do well. The 27-billion-parameter models were the best open model on 20 of the 21 public data sets and read notes, reports, long multi-document records and assertion status well. Their failures cluster on content that institutional supervision covers: time windows, laboratory thresholds, definitions and outcomes. \mainmodel{} closed this gap on the eICU and EHRSHOT timelines but not on the MIMIC-IV timelines or on clinical definitions. Part of the open models' advantage may also come from training data we cannot audit: of the open models, only JEV-27B's corpus is public, and we found no MIMIC text in it (Methods), whereas the Decision 2.0 and Clef corpora are undisclosed. The gap ran the other way on institutional data: the open models, which may have seen the public data sets in training, generalized poorly to the real-world care trajectories of \institution{} patients, where \mainmodel{} led in five of six task types. This is the setting that matters for deployment. Most hospitals' own records, with their diverse structured data and free text from routine care, are likely to resemble \institution{}'s more closely than curated research databases such as MIMIC-IV, eICU-CRD and EHRSHOT.

This study has important limitations. First, the data curation depends on design choices regarding timeline, definition and forecasting sets, which were built from each source's records and rendered as text by a rule-based serializer, and whose rule keys define each concept exactly as posed. Although we aimed at universal questions, these can differ in clinical intent, billing codes may lack timing and severity\cite{omalley2005measuring}, and outcomes may stay unobserved when care happened elsewhere\cite{lin2018outofsystem}, as, for example, in MIMIC-IV, which mostly includes emergency department and ICU visits. Second, three public keys are written by models, HISTAI's by our teacher, BRIE's by an open LLM with its authors' code and ReXVQA's by its authors' LLM pipeline, as are the 14 secondary teacher-keyed \testset{} tasks. An LLM with reasoning agreed with human keys on 94--98\% of items on three benchmarks but on only 90.6\% of phenotype judgements, below the most common answer, because its notion of a phenotype was broader than the annotators' (Methods), so model-written keys can diverge systematically from clinical annotation\cite{yamga2026llmphenotyping}. Third, \system{} is built for reading judgements over records, not for differential diagnosis from case descriptions, where System One models trail reasoning LLMs\cite{madridgarcia2026jevmedicine}. Finally, \system{} reads what records state and forecasts what records predict. If information is missing from the context provided to \system{}, its answers may be materially affected. \system{} is a tool for research, operations and review with a person in the loop, has not been validated prospectively and must not make autonomous decisions about individual patients.

\mainmodel{} is the first Clinical System One model. Its weights will follow as a credentialed release upon institutional approval. Future versions of \system{} will also read medical images, such as radiographs and pathology slides, alongside clinical text and structured records. We expect calibrated, evidence-linked answers to typed questions to become a standard layer between clinical records and the many programs, people and agents that need to read them at scale.

\begin{methods}

\subsection{Public evaluation data.}
The \CSoneNsets{} public data sets cover the six task types of \cref{fig:performance}a and are used for evaluation only (documents and scored pairs, Extended Data Table~4). We built the nine timeline, definition and forecasting sets from MIMIC-IV, eICU-CRD and EHRSHOT, the MedAgentBench records, BRIE and the two pathology sets; the others are published benchmarks converted to our question format without changing their keys. \mainmodel{} was fine-tuned only on \dataset{} data, so none of the \EXTscoredPatients{} patients behind the scored pairs contributed a fine-tuning record.
\paragraph{Structured records.} Patients come from MIMIC-IV v3.1\cite{johnson2023mimiciv,johnson2024mimiciv31}, eICU-CRD v2.0\cite{pollard2018eicu,pollard2019eicu20} (intensive-care admissions at 208 US hospitals) and EHRSHOT\cite{wornow2023ehrshot}. Each document is one patient's structured record up to a decision point, rendered as text by a deterministic rule-based serializer with only relative time printed, in the encounter layout (MIMIC-IV sets; the index admission day by day, earlier admissions summarized) or the timeline layout (eICU-CRD and EHRSHOT sets; periods counted back from the index day, recent days in full and older periods summarized); the two layouts are compared on the same MIMIC-IV documents in Extended Data Table~5. Documents fit 7,400 Qwen3 tokens (8,000 for MIMIC- and EHRSHOT-Timelines), the oldest history being shortened first. Keys are computed by rules on the structured record or are outcomes observed after the decision point. A rule is used only if the layout prints every field it reads, and a rule key is kept only when the rule applied to what the rendered text shows gives the same answer as on the full record. eICU-CRD tables are mapped to MIMIC-IV-shaped tables, so that the same rules, definitions, outcomes and question wording apply, and one stay is drawn per adult patient with hospital quotas proportional to the square root of eligible patients.
\paragraph{Patient timelines and clinical definitions.} MIMIC-Timelines and EHRSHOT-Timelines pose template-generated questions (existence in a window, thresholds, trends, counts, orderings and others) with fixed option sets; a ``no'' is keyed only when the relevant part of the record is shown in full, answers are balanced per template, and a seeded subset of test documents is scored with 40 questions each. MIMIC-Composite poses 22 published composite definitions as Choices whose option criteria state the definition's logic in plain language: eight eMERGE/PheKB phenotypes\cite{kirby2016phekb}, six Charlson and Elixhauser comorbidity groups\cite{quan2005coding}, two NHSN infection-surveillance categories\cite{horan2008nhsn}, five laboratory definitions and one stroke quality measure. eICU-Composite poses the 17 of them that port to eICU-CRD, and EHRSHOT-Composite curated published definitions and scores, with answers balanced per kind. eICU-Timelines and the Composite sets score one question per patient.
\paragraph{Outcome forecasting.} Each question asks about an event in an explicit window after a cut in the record, which shows only facts recorded by the cut; one outcome is asked per patient, at natural prevalence. MIMIC-Prognosis cuts at discharge or 24\,h after admission and asks twelve yes/no questions (outcomes of all three sets, Extended Data Table~6); a death before the event counts as ``no'', and windows that end after the patient's data end are dropped. eICU-Prognosis cuts 24\,h after hospital admission or at the first intensive-care discharge and asks about six events in the rest of the stay, because eICU-CRD records nothing after discharge. EHRSHOT-Prognosis cuts at an index day and asks each patient one question in 22 kinds; the two kinds without a single event option are reported separately. The headline is the mean AUROC\cite{hanley1982meaning} over the outcomes with at least 10 positives and 10 negatives among the scored pairs (Extended Data Table~6). The three sets differ in cuts and outcomes and are not comparable with one another.
\paragraph{MedAgentBench records.} From the de-identified FHIR records of 100 patients shipped with Med\-Agent\-Bench\cite{jiang2025medagentbench}, we rendered timelines at 1--6 cut points per patient and posed Choices in eight families, keyed by exact rules with the visibility check; a ``no'' also requires every list in the window to be complete. Answer-balanced selection kept at most 22 pairs per question.
\paragraph{BRIE.} We regenerated BRIE chart-review questions\cite{cahoon2026brie} on MIMIC-IV-Note v2.2\cite{johnson2023mimicivnote} with the authors' code and an open LLM (Gemma-4-26B-A4B\cite{gemmateam2026gemma4,google2026gemma4card}), which wrote items and keys from each patient's earlier discharge summaries and radiology reports; each item became a Choice with a not-documented option.
\paragraph{Published benchmarks.} MIMIC-IV-Ext-Apixaban-Trial-Criteria\cite{woo2025apixaban} (numeric criteria banded at trial thresholds), EHRNoteQA\cite{kweon2024ehrnoteqa,kweon2024ehrnoteqadata}, LongHealth\cite{adams2025longhealth}, MIMIC-III phenotypes\cite{gehrmann2018comparing,moseley2020phenotype,johnson2016mimiciii,johnson2016mimiciiiv14}, MIMIC-CXR-VQA\cite{bae2023ehrxqa,wu2021chestimagenome,bae2024mimiccxrvqa} (presence questions on MIMIC-CXR\cite{johnson2019mimiccxr} reports) and ReXVQA\cite{pal2025rexvqa} (a seeded stratified subset). Both radiograph sets are asked of the written report only (for MIMIC-CXR-VQA, findings the report does not mention count as absent), so the results are not comparable with image-based VQA.
\paragraph{Pathology reports.} Both sets come from the test split of a labelled collection of public pathology reports. TCGA\cite{weinstein2013tcga}: optical-character-recognition text of one diagnostic report per patient\cite{kefeli2024tcga} and 31 questions (histology, stage, grade, receptor status and others), keyed by the patient's Genomic Data Commons\cite{grossman2016gdc} records where the keyed value is stated in the report; a review of 196 pairs found no key errors. HISTAI\cite{nechaev2025histai}: 1,000 seeded cases (English translations) without an independent key, so the key is the teacher's rotation-averaged answer, the teacher is not scored, and \system{}, distilled from that teacher, has a built-in advantage; a confident subset (teacher probability $\ge0.9$) is reported separately.
\paragraph{Phenotypes and assertions.} Each note--phenotype pair of MIPA\cite{yamga2026mipa} (16 phenotypes in MIMIC-IV discharge summaries) is a two-option Choice whose criteria paraphrase MIPA's definitions. Each note--entity pair of van Aken et al.\cite{vanaken2021assertion} (mentions in MIMIC-III notes labelled present, possible or absent) is a Choice with a further not-documented option; entities with conflicting labels are dropped, and balanced accuracy averages the three class recalls.

\subsection{Institutional data (\dataset{} and \testset{}).}
\dataset{} and \testset{} were drawn from \institution{}, a multi-hospital US academic health system whose research data warehouse holds clinical notes, diagnostic reports and structured records. The \institution{} institutional review board approved the retrospective analysis of the electronic health records. The participants were not directly involved or recruited for the study. Documents were split by patient into training, validation and test sets; \dataset{} holds the training split and \testset{} the test split. Structured records were rendered in the timeline layout of the eICU-CRD and EHRSHOT sets (Public evaluation data). \mainmodel{} was trained on four groups of \dataset{} data (Extended Data Table~1): clinical notes and diagnostic reports with teacher readout labels, and structured timelines with exact labels for outcome forecasting (observed outcomes) and for structured criteria (the MIMIC-IV rule engine ported to \institution{} records, with censoring at each patient's end of follow-up). Composite definitions and outcomes marked as held out in the question bank are labelled only on test documents. The test split, \testset{}, used only for the internal evaluation (Fig.~2b and Extended Data Table~3), has 28 tasks over the six task types of Fig.~2a, with one test question per held-out patient. Keys come from rules, observed outcomes, quality-filtered structured tables, a cancer registry or an open LLM (Gemma-4-26B-A4B\cite{gemmateam2026gemma4,google2026gemma4card}); nine note and five report tasks are keyed by the GLM-5.3-Flash teacher, measure agreement with it and are secondary. The open decision models were run on the same test set; the reference is the most common training answer per question.

\subsection{Labels.}
\paragraph{Teacher readout.} GLM-5.3-Flash\cite{zai2026glm53flash,glm5team2026}, a mixture-of-experts LLM with 320 billion parameters (18 billion active), served with vLLM\cite{kwon2023efficient}, receives one raw completion prompt with a fixed instruction, the document, the question, the options as \texttt{(A) \{key\}: \{criterion\}}, the line ``Answer with a single letter.'' and the prefill \texttt{Answer: (}. For options $k_0,\dots,k_{n-1}$, rotation $s$ shows option $k_{(s+i)\bmod n}$ under letter $\ell_i$; with $r_s(i)=P(\ell_i\mid\text{prompt}_s)$ and valid-letter mass $m_s=\sum_i r_s(i)$, the label is
\begin{equation*}
\bar p(k_j)=\frac1n\sum_{s=0}^{n-1}\frac{r_s\big((j-s)\bmod n\big)}{m_s},\qquad p(k)\propto\bar p(k)^{1/T_{\mathrm{t}}}.
\end{equation*}
The $n$ cyclic rotations place every option at every position once, which cancels purely positional bias\cite{zheng2024robust,pezeshkpour2024sensitivity}. \mainmodel{}'s soft labels use $T_{\mathrm{t}}=1$.
\paragraph{Exact and outcome labels.} Rules compute timeline and composite-definition answers from the tables the document was rendered from (including derived quantities such as eGFR by CKD-EPI 2021\cite{inker2021ckdepi} and the FIB-4 liver-fibrosis index\cite{sterling2006fib4}), with the visibility check of Public evaluation data applied to \dataset{} and public labels alike. Outcomes are computed from encounters, transfers, deaths and laboratory values after the decision point, with the windows and exclusions above.
\paragraph{Reasoning annotator.} To measure how far a model-written key can be trusted, GLM-5.3\cite{zai2026glm53}, a mixture-of-experts LLM with 744 billion parameters (40 billion active), with reasoning and greedy decoding (one question per request, accepting only an explicit option letter) answered the human-keyed benchmarks. It agreed with the human keys on 94.0\% of Apixaban judgements, 97.9\% of EHRNoteQA items and 94.4\% of LongHealth items, but on only 90.6\% of MIMIC-III phenotype judgements, below the most common answer (92.5\%): it called phenotypes more liberally than the annotators.

\subsection{Model.}
\mainmodel{} is a full fine-tune of Decision 2.0 Nox-4B\cite{vllmsr2026decision20}, a Qwen3.5\cite{qwen2026qwen35} hybrid of 4.2 billion parameters (24 Gated DeltaNet linear-attention layers\cite{yang2025gated} and 8 gated full-attention layers\cite{qiu2025gated}), and keeps that model's runtime prompt and candidate head. Every question is typed: a Choice over 2--11 options, a yes/no question, or a Score with ordered levels 1 to $N$; \nd{} is offered wherever the document can be silent. The prompt holds the document, the question and one block per option, with its key and criterion; yes/no questions take the runtime's yes/no type, and Choices and Scores its choice type, with options keyed A, B, C, \dots. A question has at most 11 options. The candidate head layer-normalizes the hidden states at the last token of every option block and at the final token of the question, projects them with option- and query-side MLPs into a 256-dimensional key and query space and returns one logit per option; a softmax over the question's options gives the answer distribution. One packed sequence holds the document once, followed by one segment per question whose position indices restart at the end of the document, so that the document and any one segment are token for token the runtime's input for that question (Extended Data Fig.~2). In the full-attention layers a block mask (FlexAttention\cite{dong2024flex}) lets each segment attend to the whole document and causally to itself, never to another segment. The Gated DeltaNet layers run the document once and start every segment from its final recurrent state, recomputing the four-token causal convolution at each segment's start with the document's last tokens as left context. This extends packing without cross-contamination\cite{krell2021efficient} to a shared prefix in both layer types; training packs several documents per sequence. The document keeps at most 15,356 tokens, a segment at most 1,024 (option descriptions are shortened first), and a document with its segments at most 20,480; documents with more questions are split over units that repeat the document. One temperature per question type and number of options can be fitted on \dataset{} validation documents\cite{guo2017calibration}; \mainmodel{}'s results are raw probabilities.

\subsection{Training.}
Each example is one document with all of its labelled questions, as at inference. The loss is the soft cross-entropy between the student's distribution and the target, a strictly proper scoring rule\cite{gneiting2007strictly}, averaged over questions\cite{hinton2015distilling}; the target is the teacher's readout distribution or, where an exact or outcome label exists for the same document--question pair, the one-hot answer. Option order is shuffled per example for Choices. Only \dataset{} training-split documents are used. Backbone and head were fully fine-tuned with fused AdamW\cite{loshchilov2019adamw} ($\beta_1=0.9$, $\beta_2=0.98$, $\epsilon=10^{-6}$, weight decay 0.01), with gradient norms clipped at 1.0, in three consecutive stages. The first ran 13,591 optimizer steps on the main \dataset{} training corpus, weighted towards structured timelines, at a peak learning rate of $7\times10^{-6}$. The second continued for 11,709 steps on the same corpus, weighted towards notes and reports, re-annealing the learning rate from $3.6\times10^{-6}$ to $7\times10^{-7}$. The third ran 1,141 steps with a fresh optimizer on a set of multi-note documents, with the learning rate decaying from $3\times10^{-6}$ to $3\times10^{-7}$. In total, the 26,441 steps covered about 13.8 billion training tokens and took 26.4\,h on eight NVIDIA B300 GPUs of an institutional cluster (211 GPU-hours).

\subsection{Baselines.}
\paragraph{Open decision models.} All were run through each model's reference implementation: JEV-27B\cite{autotrust2026jev27b} (a frozen Qwen3.8-27B\cite{qwen2026qwen38} with a LoRA and a decision head distilled from a proprietary decision model); Clef\cite{cloudflare2026clef} (a post-trained Qwen3.8-27B that scores all questions of a document in one pass); the six Decision 2.0 models\cite{vllmsr2026decision20} (Kai-0.6B, Eos-0.8B, Sol-2B, Nox-4B, Lux-9B and Vega-27B; each package's own runtime, encoding a document once for all its questions); pplx-decider-v1-27b\cite{perplexity2026pplxdecider} (a fully fine-tuned Qwen3.8-27B hybrid; option-letter readout at the release's temperature; 8,192-token limit); BioDecision-4B\cite{lighteternal2026biodecision}; and MedDecider-9B and -27B\cite{lionai2026meddecider} (LoRA adapters on Qwen3.5-9B\cite{qwen2026qwen35} and Qwen3.8-27B, with the release's per-type temperatures). For each model, the mapping of our question types onto its interface was chosen among two or three variants by agreement with the teacher on 20 validation notes, without benchmark keys. Inputs longer than a model accepts were cut to their head (60\%) and tail (40\%) (Clef: at the end, per its card). JEV-27B's public training corpus contains no MIMIC text; the Decision 2.0, Clef and pplx-decider training rows are not public, so exposure to public benchmarks cannot be excluded for them or for \mainmodel{}, which starts from Nox-4B. The hosted TypeSafe Jev service\cite{typesafe2026jev} was run only on openly licensed sets (LongHealth, TCGA pathology reports) and synthetic records.
\paragraph{Other references.} The teacher readout (rotation-averaged, temperature-scaled) and the most common answer per question.

\subsection{Agent harnesses.}
A general LLM (GLM-5.3\cite{zai2026glm53}) uses \mainmodel{} as a tool\cite{yao2023react,schick2023toolformer} that takes documents and typed questions with criteria and returns option probabilities and, on request, evidence sentences. Structured records are rendered in the timeline layout for \cref{fig:screening} and by an earlier in-house serializer (index admission day by day, earlier admissions summarized) for the worked examples.

\emph{Population trial screening.} Seventeen ClinicalTrials.gov\cite{zarin2011clinicaltrials} trials, with criteria as written, are screened in a seeded sample of 1,000 MIMIC-IV test patients at the discharge of an index admission, using every structured timeline, discharge summary and radiology report up to that discharge (21,866 documents), each timeline piece one unit and a patient's notes packed in date order into units of at most 8,000 tokens (7,695 units). Per trial, one GLM-5.3 call writes atomic typed questions, with an option for missing information where absence is possible, and a second gives each a time scope (index admission, a window of $d$ days before it, or any time). \mainmodel{} answers every question on every unit; a question's probability is its maximum over units with an in-scope document. $P(\text{eligible})=\prod P(\text{inclusion met})\times\prod\big(1-P(\text{exclusion met})\big)$, missing inclusion information counting as not met. \emph{Criterion evaluation} (seven trials, a seeded 200 patients): the comparator is GLM-5.3 reading each whole record per trial; structured criteria that a rule computes exactly as stated are keyed by that rule, and note-only criteria by the agreeing labels of two full-record GLM-5.3 reads (high reasoning effort; different seeds and criterion orders). Balanced accuracy at $P(\text{met})\ge0.5$ is macro-averaged over criteria with both classes. \emph{Ranking}: all 1,000 patients are ordered per trial by $P(\text{eligible})$ and by BM25 over the full record with the inclusion criteria as query; the union of the two top 20s is graded by GLM-5.3\cite{zheng2023judging} at high reasoning effort, blind to ranker and model answers, in two passes (criteria in order and reversed\cite{wang2024fair}) of 2 (eligible), 1 (possibly eligible) or 0, the final grade being the floor of their mean. The judged pool holds 663 patient--trial pairs (linearly weighted $\kappa$\cite{cohen1968weighted} between passes 0.84). nDCG@20\cite{jarvelin2002cumulated} is $\sum_{i=1}^{20}(2^{g_i}-1)/\log_2(i+1)$ for the grades $g_i$ of a ranker's top 20, divided by the same sum for the ideal ordering of the judged pool, over the 13 trials with any patient graded above 0; intervals resample trials and, for differences, patients within trials\cite{efron1979bootstrap}; all bootstraps use 2,000 resamples. \emph{Cost}: server-reported LLM tokens per screened patient over these 13 trials, for the question-writing and time-scope calls, shared over the 1,000 patients (pipeline), or for GLM-5.3 reading each of the seeded 200 patients' records (baseline); verification of top candidates, \mainmodel{} compute and reference and judge calls are not counted.

\emph{Patient-level question answering.} From a manifest of the record (document types, counts and dates), the LLM writes typed sub-questions, each with a \nd{} option, an aggregation rule (any, latest, trajectory, majority or count) and a scope; \mainmodel{} answers every sub-question over every document in scope, chunking long documents; code aggregates across documents and time, separating documented absence from silence and flagging conflicts; sufficiency attribution selects evidence sentences in the decisive documents ($s_{\text{suf}}\ge1$ nat and $p(a\mid\text{sentence alone})\ge0.5$); and the LLM, which never reads the full record, composes the answer from cited sentences and aggregates, with citations validated against the evidence list. The three worked examples (\cref{fig:patientqa} and Extended Data Figs.~5 and 6) are MIMIC-IV test patients chosen from the 100 longest test records by structured signals so that each question applies; each is a single run, checked against the full record by GLM-5.3 at high reasoning effort and by reading short excerpts.

\subsection{Evidence attribution.}
Let $u_1,\dots,u_m$ be a document's units (sentences; for a structured record, the lines of its rendered timeline) and $a$ the model's top answer to question $q$. The necessity score of a unit is $s_{\text{nec}}(u_i)=\log p(a\mid d,q)-\log p(a\mid d\setminus u_i,q)$, where $d\setminus u_i$ is the document with $u_i$ deleted and re-encoded; the sufficiency score, $s_{\text{suf}}(u_i)=\log p(a\mid u_i,q)-\log p(a\mid\varnothing,q)$, compares the unit alone with an empty document and also finds redundantly stated evidence\cite{li2016erasure,deyoung2020eraser}. \Cref{fig:attribution} shows the corresponding probability differences, which order the units identically. Because deleting $u_i$ leaves the text before it unchanged, the model's state at $u_i$ (full-attention keys and values, Gated DeltaNet recurrent states and convolution inputs) is computed once and reused\cite{pan2025marconi}, and only the rest of the document and the question are re-encoded; this is exact up to floating-point rounding. Documents of more than 34 units are screened hierarchically: chunks of 10 units are deleted one at a time, and the units of the three chunks whose deletion lowers $\log p(a)$ most are then scored singly; every reported score is exact. Up to three units with $s_{\text{nec}}\ge0.1$ are listed as evidence. When no unit is necessary and $p(a\mid\varnothing,q)<0.5$, units with $p(a\mid u_i,q)\ge0.5$ and $s_{\text{suf}}\ge1$ are listed as redundant evidence; when $p(a\mid\varnothing,q)\ge0.5$ the answer rests on silence and nothing is highlighted. Faithfulness was evaluated on 200 pairs from 100 MIMIC-IV discharge notes of at most 3,000 tokens (per note, one question whose teacher answer is not its most common answer and one at random; evidence-dependent, the 90 with $p(a\mid\varnothing)<0.5$): the top $k\in\{1,2,3,5,10\}$ units by necessity or sufficiency were deleted (deletion) or kept alone (insertion)\cite{deyoung2020eraser,samek2017evaluating}, the area under each curve is the original answer's mean probability over $k$, and two random orderings serve as baseline; results are for the 90 evidence-dependent pairs.

\subsection{Metrics and statistics.}
Accuracy is computed on the top option. Balanced accuracy\cite{brodersen2010balanced} is the mean recall over the key's answer classes per question, averaged over questions (questions with fewer than 20 items, and item-specific options, are pooled by answer position); on eICU-Timelines and the three Composite sets it is computed per concept and averaged over concepts. Calibration is reported as top-label expected calibration error (15 equal-count bins)\cite{naeini2015obtaining,nixon2019measuring} and Brier score\cite{brier1950verification}, and outcomes also by AUROC\cite{hanley1982meaning}, AUPRC\cite{saito2015precision}, scaled Brier score\cite{steyerberg2010assessing} and calibration slope and intercept\cite{vancalster2019calibration}. For the recalibrated forecasts, each outcome's predicted risks were shifted by one constant on the log-odds scale so that their mean equals the outcome's incidence, with the shift estimated on one half of the patients and applied to the other (two folds by patient); the calibration error of predicted risks uses 10 equal-count bins. Coverage at precision $p$ is the largest share of judgements, taken in order of confidence, whose accuracy is at least $p$\cite{chow1970optimum,geifman2017selective}, with confidence the top option's renormalized probability and tied confidences accepted or rejected together. Intervals are 95\% document- or patient-clustered bootstrap\cite{efron1979bootstrap,field2007bootstrapping} intervals (1,000 resamples); for balanced accuracy on the MIMIC-IV timeline, human-keyed, ReXVQA and pathology sets, a Bayesian bootstrap\cite{rubin1981bayesian} (Dirichlet document weights) keeps every answer class in every resample. Per-concept means use a Bayesian bootstrap over concepts (small-sample $t$ correction) and prognosis means a patient-level Bayesian bootstrap. Paired comparisons use the paired bootstrap, and every system is scored on the same items.

\mainmodel{} and the six Decision 2.0 models were timed back to back on one idle NVIDIA GB10, one model and one document at a time in bf16, with the documents and question counts of Extended Data Table~2 and Extended Data Fig.~3. Each ran through its own released inference code, Decision 2.0 through its official runtime as documented on its model cards (one \texttt{system\_one} call per document with all its questions). We did not further optimize the Decision 2.0 pipeline, which on this GPU uses the generic Transformers\cite{wolf2020transformers} forward pass. \mainmodel{} and Nox-4B share an architecture and differ only in their weights, so the 1.3-fold difference between them comes from the inference code rather than from the model, and an optimized Decision 2.0 runtime would likely narrow it; the gap that comes from model size remains. Times are end-to-end wall-clock times per document, excluding model loading: the median of 10 timed repetitions after two warm-up runs (5 for runs longer than 5\,s). Compute for $10^6$ documents is extrapolated from these measurements.

\subsection{Ethics and data governance.}
MIMIC-IV, MIMIC-IV-Note, MIMIC-III, MIMIC-CXR, eICU-CRD and the derived benchmarks were used under the PhysioNet\cite{pollard2026physionet} credentialed (or restricted) data-use agreements by credentialed investigators; EHRSHOT under its credentialed agreement by individually registered investigators, with local models only\cite{physionet2025llm}. These data are de-identified, and their use was exempt from further review. \institution{} data were used on institution-controlled infrastructure. The examples in Figs.~1 and 3 are synthetic. \Cref{fig:patientqa}, Extended Data Figs.~5 and 6 and \cref{fig:screening}a show MIMIC-IV test patients with relabelled document and patient labels and de-identified, shifted dates, and show record text only as paraphrases, never verbatim; no record identifier is shown.

\subsection{Software and hardware.}
Python 3.13, PyTorch 2.14\cite{paszke2019pytorch} (CUDA 13.0) with FlexAttention and \texttt{torch.\allowbreak compile}\cite{ansel2024pytorch2}, flash-linear-attention 0.5\cite{yang2024fla}, Hugging Face Transformers 5.12\cite{wolf2020transformers} and PEFT 0.21\cite{mangrulkar2022peft}; NumPy 2.5\cite{harris2020numpy}, SciPy 1.18\cite{virtanen2020scipy}, scikit-learn 1.9\cite{pedregosa2011scikit}, pandas 3.0\cite{mckinney2010pandas} and DuckDB 1.5\cite{raasveldt2019duckdb} for analysis; Matplotlib 3.11\cite{hunter2007matplotlib} for figures; vLLM 0.27.0 or later with FlashInfer 0.6.18 or later\cite{ye2025flashinfer} for serving the LLMs. Inference of \mainmodel{} and the open decision models, latency measurements and attribution ran on NVIDIA GB10 (DGX Spark, 128\,GB unified memory); \mainmodel{} was trained on eight NVIDIA B300 GPUs (211 GPU-hours), and the LLMs were served on NVIDIA B300 GPUs of the same institutional cluster.

\end{methods}

\FloatBarrier
\bibliographystyle{naturemag}
\bibliography{refs}

\begin{thebibliography}{100}
\expandafter\ifx\csname url\endcsname\relax
  \def\url#1{\texttt{#1}}\fi
\expandafter\ifx\csname urlprefix\endcsname\relax\def\urlprefix{URL }\fi
\providecommand{\bibinfo}[2]{#2}
\providecommand{\eprint}[2][]{\url{#2}}

\bibitem{jin2024trialgpt}
\bibinfo{author}{Jin, Q.} \emph{et~al.}
\newblock \bibinfo{title}{Matching patients to clinical trials with large
  language models}.
\newblock \emph{\bibinfo{journal}{Nature Communications}}
  \textbf{\bibinfo{volume}{15}}, \bibinfo{pages}{9074} (\bibinfo{year}{2024}).

\bibitem{penberthy2012effort}
\bibinfo{author}{Penberthy, L.~T.}, \bibinfo{author}{Dahman, B.~A.},
  \bibinfo{author}{Petkov, V.~I.} \& \bibinfo{author}{DeShazo, J.~P.}
\newblock \bibinfo{title}{Effort required in eligibility screening for clinical
  trials}.
\newblock \emph{\bibinfo{journal}{Journal of Oncology Practice}}
  \textbf{\bibinfo{volume}{8}}, \bibinfo{pages}{365--370}
  (\bibinfo{year}{2012}).

\bibitem{banda2018advances}
\bibinfo{author}{Banda, J.~M.}, \bibinfo{author}{Seneviratne, M.},
  \bibinfo{author}{Hernandez-Boussard, T.} \& \bibinfo{author}{Shah, N.~H.}
\newblock \bibinfo{title}{Advances in electronic phenotyping: From rule-based
  definitions to machine learning models}.
\newblock \emph{\bibinfo{journal}{Annual Review of Biomedical Data Science}}
  \textbf{\bibinfo{volume}{1}}, \bibinfo{pages}{53--68} (\bibinfo{year}{2018}).

\bibitem{casalino2016quality}
\bibinfo{author}{Casalino, L.~P.} \emph{et~al.}
\newblock \bibinfo{title}{{US} physician practices spend more than \$15.4
  billion annually to report quality measures}.
\newblock \emph{\bibinfo{journal}{Health Affairs}}
  \textbf{\bibinfo{volume}{35}}, \bibinfo{pages}{401--406}
  (\bibinfo{year}{2016}).

\bibitem{overhage2020physician}
\bibinfo{author}{Overhage, J.~M.} \& \bibinfo{author}{McCallie, D.}
\newblock \bibinfo{title}{Physician time spent using the electronic health
  record during outpatient encounters: A descriptive study}.
\newblock \emph{\bibinfo{journal}{Annals of Internal Medicine}}
  \textbf{\bibinfo{volume}{172}}, \bibinfo{pages}{169--174}
  (\bibinfo{year}{2020}).

\bibitem{kahneman2011thinking}
\bibinfo{author}{Kahneman, D.}
\newblock \emph{\bibinfo{title}{Thinking, Fast and Slow}}
  (\bibinfo{publisher}{Farrar, Straus and Giroux}, \bibinfo{address}{New York},
  \bibinfo{year}{2011}).

\bibitem{savova2010ctakes}
\bibinfo{author}{Savova, G.~K.} \emph{et~al.}
\newblock \bibinfo{title}{Mayo clinical {Text Analysis and Knowledge Extraction
  System} ({cTAKES}): architecture, component evaluation and applications}.
\newblock \emph{\bibinfo{journal}{Journal of the American Medical Informatics
  Association}} \textbf{\bibinfo{volume}{17}}, \bibinfo{pages}{507--513}
  (\bibinfo{year}{2010}).

\bibitem{chapman2001negex}
\bibinfo{author}{Chapman, W.~W.}, \bibinfo{author}{Bridewell, W.},
  \bibinfo{author}{Hanbury, P.}, \bibinfo{author}{Cooper, G.~F.} \&
  \bibinfo{author}{Buchanan, B.~G.}
\newblock \bibinfo{title}{A simple algorithm for identifying negated findings
  and diseases in discharge summaries}.
\newblock \emph{\bibinfo{journal}{Journal of Biomedical Informatics}}
  \textbf{\bibinfo{volume}{34}}, \bibinfo{pages}{301--310}
  (\bibinfo{year}{2001}).

\bibitem{gehrmann2018comparing}
\bibinfo{author}{Gehrmann, S.} \emph{et~al.}
\newblock \bibinfo{title}{Comparing deep learning and concept extraction based
  methods for patient phenotyping from clinical narratives}.
\newblock \emph{\bibinfo{journal}{PLoS ONE}} \textbf{\bibinfo{volume}{13}},
  \bibinfo{pages}{e0192360} (\bibinfo{year}{2018}).

\bibitem{rajkomar2018scalable}
\bibinfo{author}{Rajkomar, A.} \emph{et~al.}
\newblock \bibinfo{title}{Scalable and accurate deep learning with electronic
  health records}.
\newblock \emph{\bibinfo{journal}{npj Digital Medicine}}
  \textbf{\bibinfo{volume}{1}}, \bibinfo{pages}{18} (\bibinfo{year}{2018}).

\bibitem{agrawal2022large}
\bibinfo{author}{Agrawal, M.}, \bibinfo{author}{Hegselmann, S.},
  \bibinfo{author}{Lang, H.}, \bibinfo{author}{Kim, Y.} \&
  \bibinfo{author}{Sontag, D.}
\newblock \bibinfo{title}{Large language models are few-shot clinical
  information extractors}.
\newblock In \emph{\bibinfo{booktitle}{Proceedings of the 2022 Conference on
  Empirical Methods in Natural Language Processing (EMNLP)}},
  \bibinfo{pages}{1998--2022} (\bibinfo{year}{2022}).

\bibitem{vanveen2024adapted}
\bibinfo{author}{Van~Veen, D.} \emph{et~al.}
\newblock \bibinfo{title}{Adapted large language models can outperform medical
  experts in clinical text summarization}.
\newblock \emph{\bibinfo{journal}{Nature Medicine}}
  \textbf{\bibinfo{volume}{30}}, \bibinfo{pages}{1134--1142}
  (\bibinfo{year}{2024}).

\bibitem{singhal2023large}
\bibinfo{author}{Singhal, K.} \emph{et~al.}
\newblock \bibinfo{title}{Large language models encode clinical knowledge}.
\newblock \emph{\bibinfo{journal}{Nature}} \textbf{\bibinfo{volume}{620}},
  \bibinfo{pages}{172--180} (\bibinfo{year}{2023}).

\bibitem{moor2023foundation}
\bibinfo{author}{Moor, M.} \emph{et~al.}
\newblock \bibinfo{title}{Foundation models for generalist medical artificial
  intelligence}.
\newblock \emph{\bibinfo{journal}{Nature}} \textbf{\bibinfo{volume}{616}},
  \bibinfo{pages}{259--265} (\bibinfo{year}{2023}).

\bibitem{wei2022chain}
\bibinfo{author}{Wei, J.} \emph{et~al.}
\newblock \bibinfo{title}{Chain-of-thought prompting elicits reasoning in large
  language models}.
\newblock In \emph{\bibinfo{booktitle}{Advances in Neural Information
  Processing Systems 35}}, \bibinfo{pages}{24824--24837}
  (\bibinfo{year}{2022}).

\bibitem{kwon2023efficient}
\bibinfo{author}{Kwon, W.} \emph{et~al.}
\newblock \bibinfo{title}{Efficient memory management for large language model
  serving with {PagedAttention}}.
\newblock In \emph{\bibinfo{booktitle}{Proceedings of the 29th Symposium on
  Operating Systems Principles (SOSP)}}, \bibinfo{pages}{611--626}
  (\bibinfo{year}{2023}).

\bibitem{zheng2024sglang}
\bibinfo{author}{Zheng, L.} \emph{et~al.}
\newblock \bibinfo{title}{{SGLang}: Efficient execution of structured language
  model programs}.
\newblock In \emph{\bibinfo{booktitle}{Advances in Neural Information
  Processing Systems 37}}, \bibinfo{pages}{62557--62583}
  (\bibinfo{year}{2024}).

\bibitem{cheng2023batch}
\bibinfo{author}{Cheng, Z.}, \bibinfo{author}{Kasai, J.} \&
  \bibinfo{author}{Yu, T.}
\newblock \bibinfo{title}{Batch prompting: Efficient inference with large
  language model {APIs}}.
\newblock In \emph{\bibinfo{booktitle}{Proceedings of the 2023 Conference on
  Empirical Methods in Natural Language Processing: Industry Track}},
  \bibinfo{pages}{792--810} (\bibinfo{year}{2023}).

\bibitem{jiang2021can}
\bibinfo{author}{Jiang, Z.}, \bibinfo{author}{Araki, J.},
  \bibinfo{author}{Ding, H.} \& \bibinfo{author}{Neubig, G.}
\newblock \bibinfo{title}{How can we know when language models know? on the
  calibration of language models for question answering}.
\newblock \emph{\bibinfo{journal}{Transactions of the Association for
  Computational Linguistics}} \textbf{\bibinfo{volume}{9}},
  \bibinfo{pages}{962--977} (\bibinfo{year}{2021}).

\bibitem{griot2025metacognition}
\bibinfo{author}{Griot, M.}, \bibinfo{author}{Hemptinne, C.},
  \bibinfo{author}{Vanderdonckt, J.} \& \bibinfo{author}{Yuksel, D.}
\newblock \bibinfo{title}{Large language models lack essential metacognition
  for reliable medical reasoning}.
\newblock \emph{\bibinfo{journal}{Nature Communications}}
  \textbf{\bibinfo{volume}{16}}, \bibinfo{pages}{642} (\bibinfo{year}{2025}).

\bibitem{wu2025sourcecheckup}
\bibinfo{author}{Wu, K.} \emph{et~al.}
\newblock \bibinfo{title}{An automated framework for assessing how well {LLMs}
  cite relevant medical references}.
\newblock \emph{\bibinfo{journal}{Nature Communications}}
  \textbf{\bibinfo{volume}{16}}, \bibinfo{pages}{3615} (\bibinfo{year}{2025}).

\bibitem{physionet2023llm}
\bibinfo{author}{{PhysioNet}}.
\newblock \bibinfo{title}{Responsible use of {MIMIC} data with online services
  like {GPT}}.
\newblock \bibinfo{howpublished}{PhysioNet news} (\bibinfo{year}{2023}).
\newblock \urlprefix\url{https://physionet.org/news/post/gpt-responsible-use/}.

\bibitem{physionet2025llm}
\bibinfo{author}{{PhysioNet}}.
\newblock \bibinfo{title}{Use of {MIMIC} data with large language models and
  online services}.
\newblock \bibinfo{howpublished}{PhysioNet news} (\bibinfo{year}{2025}).
\newblock \urlprefix\url{https://physionet.org/news/post/llm-responsible-use/}.

\bibitem{typesafe2026jev}
\bibinfo{author}{Almeida, D.}
\newblock \bibinfo{title}{Introducing {System One} models \& {Jev}}.
\newblock \bibinfo{howpublished}{TypeSafe AI blog} (\bibinfo{year}{2026}).
\newblock
  \urlprefix\url{https://typesafe.ai/blog/introducing-system-one-models-and-jev}.

\bibitem{autotrust2026jev27b}
\bibinfo{author}{{AutoTrust AI}}.
\newblock \bibinfo{title}{{JEV-27B}: a second-generation integrated {System 1}
  + {System 2} open model built with the {Blocks of Experts} recipe
  ({Qwen3.8-27B}; {System 1} distilled from {TypeSafe Jev 1.13})}.
\newblock \bibinfo{howpublished}{Model card, revision 962701f3, 1 October}
  (\bibinfo{year}{2026}).
\newblock \urlprefix\url{https://huggingface.co/autotrust/JEV-27B}.

\bibitem{cloudflare2026clef}
\bibinfo{author}{{Cloudflare}}.
\newblock \bibinfo{title}{{Clef}: a typed-decision model with a joint schema
  head ({27B}, {Qwen3.8}-based)}.
\newblock \bibinfo{howpublished}{Model card, Hugging Face
  \texttt{Cloudflare/clef}, Apache-2.0} (\bibinfo{year}{2026}).
\newblock \urlprefix\url{https://huggingface.co/Cloudflare/clef}.

\bibitem{vllmsr2026decision20}
\bibinfo{author}{{vLLM Semantic Router Team}}.
\newblock \bibinfo{title}{Decision 2.0 ({Kai-0.6B}, {Eos-0.8B}, {Sol-2B},
  {Nox-4B}, {Lux-9B}, {Vega-27B})}.
\newblock \bibinfo{howpublished}{Model cards, Hugging Face collection
  \texttt{vllm-sr/decision-20}, Apache-2.0} (\bibinfo{year}{2026}).
\newblock
  \urlprefix\url{https://huggingface.co/collections/vllm-sr/decision-20}.

\bibitem{madridgarcia2026jevmedicine}
\bibinfo{author}{Madrid-Garc{\'\i}a, A.} \& \bibinfo{author}{Merino-Barbancho,
  B.}
\newblock \bibinfo{title}{{Jev} in medicine: A benchmark evaluation}.
\newblock \bibinfo{howpublished}{arXiv preprint arXiv:2609.34024 (version 3)}
  (\bibinfo{year}{2026}).
\newblock \urlprefix\url{https://arxiv.org/abs/2609.34024v3}.
\newblock \bibinfo{note}{Submitted 27 September 2026; revised 4 October 2026}.

\bibitem{huang2026jevradiology}
\bibinfo{author}{Huang, J.} \emph{et~al.}
\newblock \bibinfo{title}{Can {Jev} judge radiology reports? {Evaluating} a
  {System One} model for clinical factuality}.
\newblock \bibinfo{howpublished}{arXiv:2609.27607} (\bibinfo{year}{2026}).
\newblock \urlprefix\url{https://arxiv.org/abs/2609.27607}.

\bibitem{lighteternal2026biodecision}
\bibinfo{author}{{lighteternal}}.
\newblock \bibinfo{title}{{BioDecision-4B v2}: a system-1 decision model for
  biomedicine, pharma and clinical trials}.
\newblock \bibinfo{howpublished}{Hugging Face model card,
  \url{https://huggingface.co/lighteternal/biodecision-v2-4b}}
  (\bibinfo{year}{2026}).

\bibitem{lionai2026meddecider}
\bibinfo{author}{{thelion.ai}}.
\newblock \bibinfo{title}{{MedDecider}: calibrated medical decision models
  ({MedDecider-9B} and {MedDecider-27B})}.
\newblock \bibinfo{howpublished}{Hugging Face model cards,
  \url{https://huggingface.co/lion-ai/MedDecider-9B} and
  \url{https://huggingface.co/lion-ai/MedDecider-27B}} (\bibinfo{year}{2026}).

\bibitem{ma2026medjev}
\bibinfo{author}{Ma, J.}
\newblock \bibinfo{title}{{MedJev}: Ultra-fast clinical variable extraction
  from free-text notes}.
\newblock \bibinfo{howpublished}{GitHub repository, version 0.1.0}
  (\bibinfo{year}{2026}).
\newblock \urlprefix\url{https://github.com/JunMa11/MedJev}.

\bibitem{jiang2023health}
\bibinfo{author}{Jiang, L.~Y.} \emph{et~al.}
\newblock \bibinfo{title}{Health system-scale language models are all-purpose
  prediction engines}.
\newblock \emph{\bibinfo{journal}{Nature}} \textbf{\bibinfo{volume}{619}},
  \bibinfo{pages}{357--362} (\bibinfo{year}{2023}).

\bibitem{steinberg2021clmbr}
\bibinfo{author}{Steinberg, E.} \emph{et~al.}
\newblock \bibinfo{title}{Language models are an effective representation
  learning technique for electronic health record data}.
\newblock \emph{\bibinfo{journal}{Journal of Biomedical Informatics}}
  \textbf{\bibinfo{volume}{113}}, \bibinfo{pages}{103637}
  (\bibinfo{year}{2021}).

\bibitem{wong2021external}
\bibinfo{author}{Wong, A.} \emph{et~al.}
\newblock \bibinfo{title}{External validation of a widely implemented
  proprietary sepsis prediction model in hospitalized patients}.
\newblock \emph{\bibinfo{journal}{JAMA Internal Medicine}}
  \textbf{\bibinfo{volume}{181}}, \bibinfo{pages}{1065--1070}
  (\bibinfo{year}{2021}).

\bibitem{futoma2020myth}
\bibinfo{author}{Futoma, J.}, \bibinfo{author}{Simons, M.},
  \bibinfo{author}{Panch, T.}, \bibinfo{author}{Doshi-Velez, F.} \&
  \bibinfo{author}{Celi, L.~A.}
\newblock \bibinfo{title}{The myth of generalisability in clinical research and
  machine learning in health care}.
\newblock \emph{\bibinfo{journal}{The Lancet Digital Health}}
  \textbf{\bibinfo{volume}{2}}, \bibinfo{pages}{e489--e492}
  (\bibinfo{year}{2020}).

\bibitem{zech2018variable}
\bibinfo{author}{Zech, J.~R.} \emph{et~al.}
\newblock \bibinfo{title}{Variable generalization performance of a deep
  learning model to detect pneumonia in chest radiographs: a cross-sectional
  study}.
\newblock \emph{\bibinfo{journal}{PLoS Medicine}}
  \textbf{\bibinfo{volume}{15}}, \bibinfo{pages}{e1002683}
  (\bibinfo{year}{2018}).

\bibitem{wornow2023shaky}
\bibinfo{author}{Wornow, M.} \emph{et~al.}
\newblock \bibinfo{title}{The shaky foundations of large language models and
  foundation models for electronic health records}.
\newblock \emph{\bibinfo{journal}{npj Digital Medicine}}
  \textbf{\bibinfo{volume}{6}}, \bibinfo{pages}{135} (\bibinfo{year}{2023}).

\bibitem{zai2026glm53flash}
\bibinfo{author}{{Z.ai (Zhipu AI)}}.
\newblock \bibinfo{title}{{GLM-5.3-Flash}}.
\newblock \bibinfo{howpublished}{Model card} (\bibinfo{year}{2026}).
\newblock \urlprefix\url{https://huggingface.co/zai-org/GLM-5.3-Flash}.

\bibitem{perplexity2026pplxdecider}
\bibinfo{author}{{Perplexity AI}}.
\newblock \bibinfo{title}{pplx-decider-v1-27b}.
\newblock \bibinfo{howpublished}{Hugging Face model card,
  \url{https://huggingface.co/perplexity-ai/pplx-decider-v1-27b}}
  (\bibinfo{year}{2026}).

\bibitem{johnson2023mimiciv}
\bibinfo{author}{Johnson, A. E.~W.} \emph{et~al.}
\newblock \bibinfo{title}{{MIMIC-IV}, a freely accessible electronic health
  record dataset}.
\newblock \emph{\bibinfo{journal}{Scientific Data}}
  \textbf{\bibinfo{volume}{10}}, \bibinfo{pages}{1} (\bibinfo{year}{2023}).

\bibitem{johnson2023mimicivnote}
\bibinfo{author}{Johnson, A.}, \bibinfo{author}{Pollard, T.},
  \bibinfo{author}{Horng, S.}, \bibinfo{author}{Celi, L.~A.} \&
  \bibinfo{author}{Mark, R.}
\newblock \bibinfo{title}{{MIMIC-IV-Note}: Deidentified free-text clinical
  notes (version 2.2)}.
\newblock \bibinfo{howpublished}{PhysioNet} (\bibinfo{year}{2023}).
\newblock \urlprefix\url{https://physionet.org/content/mimic-iv-note/}.

\bibitem{pollard2018eicu}
\bibinfo{author}{Pollard, T.~J.} \emph{et~al.}
\newblock \bibinfo{title}{The {eICU Collaborative Research Database}, a freely
  available multi-center database for critical care research}.
\newblock \emph{\bibinfo{journal}{Scientific Data}}
  \textbf{\bibinfo{volume}{5}}, \bibinfo{pages}{180178} (\bibinfo{year}{2018}).

\bibitem{wornow2023ehrshot}
\bibinfo{author}{Wornow, M.}, \bibinfo{author}{Thapa, R.},
  \bibinfo{author}{Steinberg, E.}, \bibinfo{author}{Fries, J.~A.} \&
  \bibinfo{author}{Shah, N.~H.}
\newblock \bibinfo{title}{{EHRSHOT}: An {EHR} benchmark for few-shot evaluation
  of foundation models}.
\newblock In \emph{\bibinfo{booktitle}{Advances in Neural Information
  Processing Systems 36 (Datasets and Benchmarks Track)}},
  \bibinfo{pages}{67125--67137} (\bibinfo{year}{2023}).

\bibitem{sainz2023nlp}
\bibinfo{author}{Sainz, O.} \emph{et~al.}
\newblock \bibinfo{title}{{NLP} evaluation in trouble: On the need to measure
  {LLM} data contamination for each benchmark}.
\newblock In \emph{\bibinfo{booktitle}{Findings of the Association for
  Computational Linguistics: EMNLP 2023}}, \bibinfo{pages}{10776--10787}
  (\bibinfo{year}{2023}).

\bibitem{saerens2002adjusting}
\bibinfo{author}{Saerens, M.}, \bibinfo{author}{Latinne, P.} \&
  \bibinfo{author}{Decaestecker, C.}
\newblock \bibinfo{title}{Adjusting the outputs of a classifier to new a priori
  probabilities: A simple procedure}.
\newblock \emph{\bibinfo{journal}{Neural Computation}}
  \textbf{\bibinfo{volume}{14}}, \bibinfo{pages}{21--41}
  (\bibinfo{year}{2002}).

\bibitem{deyoung2020eraser}
\bibinfo{author}{DeYoung, J.} \emph{et~al.}
\newblock \bibinfo{title}{{ERASER}: A benchmark to evaluate rationalized {NLP}
  models}.
\newblock In \emph{\bibinfo{booktitle}{Proceedings of the 58th Annual Meeting
  of the Association for Computational Linguistics}},
  \bibinfo{pages}{4443--4458} (\bibinfo{year}{2020}).

\bibitem{cohenwang2024contextcite}
\bibinfo{author}{Cohen-Wang, B.}, \bibinfo{author}{Shah, H.},
  \bibinfo{author}{Georgiev, K.} \& \bibinfo{author}{M{\k{a}}dry, A.}
\newblock \bibinfo{title}{{ContextCite}: Attributing model generation to
  context}.
\newblock In \emph{\bibinfo{booktitle}{Advances in Neural Information
  Processing Systems 37}}, \bibinfo{pages}{95764--95807}
  (\bibinfo{year}{2024}).

\bibitem{fleming2024medalign}
\bibinfo{author}{Fleming, S.~L.} \emph{et~al.}
\newblock \bibinfo{title}{{MedAlign}: A clinician-generated dataset for
  instruction following with electronic medical records}.
\newblock In \emph{\bibinfo{booktitle}{Proceedings of the AAAI Conference on
  Artificial Intelligence}}, vol.~\bibinfo{volume}{38},
  \bibinfo{pages}{22021--22030} (\bibinfo{year}{2024}).

\bibitem{adams2025longhealth}
\bibinfo{author}{Adams, L.} \emph{et~al.}
\newblock \bibinfo{title}{{LongHealth}: A question answering benchmark with
  long clinical documents}.
\newblock \emph{\bibinfo{journal}{Journal of Healthcare Informatics Research}}
  \textbf{\bibinfo{volume}{9}}, \bibinfo{pages}{280--296}
  (\bibinfo{year}{2025}).

\bibitem{cahoon2026brie}
\bibinfo{author}{Cahoon, J.~L.} \emph{et~al.}
\newblock \bibinfo{title}{A living benchmark for information retrieval from
  electronic health records}.
\newblock \bibinfo{howpublished}{arXiv preprint arXiv:2609.30205}
  (\bibinfo{year}{2026}).

\bibitem{zai2026glm53}
\bibinfo{author}{{Z.ai (Zhipu AI)}}.
\newblock \bibinfo{title}{{GLM-5.3}}.
\newblock \bibinfo{howpublished}{Model card} (\bibinfo{year}{2026}).
\newblock \urlprefix\url{https://huggingface.co/zai-org/GLM-5.3}.

\bibitem{yao2023react}
\bibinfo{author}{Yao, S.} \emph{et~al.}
\newblock \bibinfo{title}{{ReAct}: Synergizing reasoning and acting in language
  models}.
\newblock In \emph{\bibinfo{booktitle}{International Conference on Learning
  Representations}} (\bibinfo{year}{2023}).

\bibitem{schick2023toolformer}
\bibinfo{author}{Schick, T.} \emph{et~al.}
\newblock \bibinfo{title}{Toolformer: Language models can teach themselves to
  use tools}.
\newblock In \emph{\bibinfo{booktitle}{Advances in Neural Information
  Processing Systems 36}}, \bibinfo{pages}{68539--68551}
  (\bibinfo{year}{2023}).

\bibitem{zheng2023judging}
\bibinfo{author}{Zheng, L.} \emph{et~al.}
\newblock \bibinfo{title}{Judging {LLM}-as-a-judge with {MT-Bench} and {Chatbot
  Arena}}.
\newblock In \emph{\bibinfo{booktitle}{Advances in Neural Information
  Processing Systems 36 (Datasets and Benchmarks Track)}},
  \bibinfo{pages}{46595--46623} (\bibinfo{year}{2023}).

\bibitem{wacker2022vitaminc}
\bibinfo{author}{Wacker, D.~A.} \emph{et~al.}
\newblock \bibinfo{title}{Evaluating vitamin {C} in septic shock: A randomized
  controlled trial of vitamin {C} monotherapy}.
\newblock \emph{\bibinfo{journal}{Critical Care Medicine}}
  \textbf{\bibinfo{volume}{50}}, \bibinfo{pages}{e458--e467}
  (\bibinfo{year}{2022}).

\bibitem{cohen1968weighted}
\bibinfo{author}{Cohen, J.}
\newblock \bibinfo{title}{Weighted kappa: Nominal scale agreement with
  provision for scaled disagreement or partial credit}.
\newblock \emph{\bibinfo{journal}{Psychological Bulletin}}
  \textbf{\bibinfo{volume}{70}}, \bibinfo{pages}{213--220}
  (\bibinfo{year}{1968}).

\bibitem{jarvelin2002cumulated}
\bibinfo{author}{J{\"a}rvelin, K.} \& \bibinfo{author}{Kek{\"a}l{\"a}inen, J.}
\newblock \bibinfo{title}{Cumulated gain-based evaluation of {IR} techniques}.
\newblock \emph{\bibinfo{journal}{ACM Transactions on Information Systems}}
  \textbf{\bibinfo{volume}{20}}, \bibinfo{pages}{422--446}
  (\bibinfo{year}{2002}).

\bibitem{unlu2025manual}
\bibinfo{author}{Unlu, O.} \emph{et~al.}
\newblock \bibinfo{title}{Manual vs {AI}-assisted prescreening for trial
  eligibility using large language models---a randomized clinical trial}.
\newblock \emph{\bibinfo{journal}{JAMA}} \textbf{\bibinfo{volume}{333}},
  \bibinfo{pages}{1084--1087} (\bibinfo{year}{2025}).

\bibitem{unlu2024rectifier}
\bibinfo{author}{Unlu, O.} \emph{et~al.}
\newblock \bibinfo{title}{Retrieval-augmented generation--enabled {GPT-4} for
  clinical trial screening}.
\newblock \emph{\bibinfo{journal}{NEJM AI}} \textbf{\bibinfo{volume}{1}}
  (\bibinfo{year}{2024}).

\bibitem{wornow2024zeroshot}
\bibinfo{author}{Wornow, M.} \emph{et~al.}
\newblock \bibinfo{title}{Zero-shot clinical trial patient matching with
  {LLMs}}.
\newblock \bibinfo{howpublished}{arXiv:2402.05125} (\bibinfo{year}{2024}).
\newblock \urlprefix\url{https://arxiv.org/abs/2402.05125}.

\bibitem{syed2026translating}
\bibinfo{author}{Syed, M.} \emph{et~al.}
\newblock \bibinfo{title}{Translating evidence into practice: adapting
  {TrialGPT} for real-world clinical trial eligibility screening}.
\newblock \emph{\bibinfo{journal}{Journal of the American Medical Informatics
  Association}} \textbf{\bibinfo{volume}{33}}, \bibinfo{pages}{909--913}
  (\bibinfo{year}{2026}).

\bibitem{yuan2019criteria2query}
\bibinfo{author}{Yuan, C.} \emph{et~al.}
\newblock \bibinfo{title}{{Criteria2Query}: a natural language interface to
  clinical databases for cohort definition}.
\newblock \emph{\bibinfo{journal}{Journal of the American Medical Informatics
  Association}} \textbf{\bibinfo{volume}{26}}, \bibinfo{pages}{294--305}
  (\bibinfo{year}{2019}).

\bibitem{zarin2011clinicaltrials}
\bibinfo{author}{Zarin, D.~A.}, \bibinfo{author}{Tse, T.},
  \bibinfo{author}{Williams, R.~J.}, \bibinfo{author}{Califf, R.~M.} \&
  \bibinfo{author}{Ide, N.~C.}
\newblock \bibinfo{title}{The {ClinicalTrials.gov} results database---update
  and key issues}.
\newblock \emph{\bibinfo{journal}{New England Journal of Medicine}}
  \textbf{\bibinfo{volume}{364}}, \bibinfo{pages}{852--860}
  (\bibinfo{year}{2011}).

\bibitem{panickssery2024llm}
\bibinfo{author}{Panickssery, A.}, \bibinfo{author}{Bowman, S.~R.} \&
  \bibinfo{author}{Feng, S.}
\newblock \bibinfo{title}{{LLM} evaluators recognize and favor their own
  generations}.
\newblock In \emph{\bibinfo{booktitle}{Advances in Neural Information
  Processing Systems 37}}, \bibinfo{pages}{68772--68802}
  (\bibinfo{year}{2024}).

\bibitem{kraljevic2024foresight}
\bibinfo{author}{Kraljevic, Z.} \emph{et~al.}
\newblock \bibinfo{title}{Foresight---a generative pretrained transformer for
  modelling of patient timelines using electronic health records: a
  retrospective modelling study}.
\newblock \emph{\bibinfo{journal}{The Lancet Digital Health}}
  \textbf{\bibinfo{volume}{6}}, \bibinfo{pages}{e281--e290}
  (\bibinfo{year}{2024}).

\bibitem{shmatko2025delphi}
\bibinfo{author}{Shmatko, A.} \emph{et~al.}
\newblock \bibinfo{title}{Learning the natural history of human disease with
  generative transformers}.
\newblock \emph{\bibinfo{journal}{Nature}} \textbf{\bibinfo{volume}{647}},
  \bibinfo{pages}{248--256} (\bibinfo{year}{2025}).

\bibitem{zhang2026apollo}
\bibinfo{author}{Zhang, A.} \emph{et~al.}
\newblock \bibinfo{title}{A multimodal and temporal foundation model for
  virtual patient representations at healthcare system scale}.
\newblock \emph{\bibinfo{journal}{arXiv preprint arXiv:2604.18570}}
  (\bibinfo{year}{2026}).

\bibitem{omalley2005measuring}
\bibinfo{author}{O'Malley, K.~J.} \emph{et~al.}
\newblock \bibinfo{title}{Measuring diagnoses: {ICD} code accuracy}.
\newblock \emph{\bibinfo{journal}{Health Services Research}}
  \textbf{\bibinfo{volume}{40}}, \bibinfo{pages}{1620--1639}
  (\bibinfo{year}{2005}).

\bibitem{lin2018outofsystem}
\bibinfo{author}{Lin, K.~J.} \emph{et~al.}
\newblock \bibinfo{title}{Out-of-system care and recording of patient
  characteristics critical for comparative effectiveness research}.
\newblock \emph{\bibinfo{journal}{Epidemiology}} \textbf{\bibinfo{volume}{29}},
  \bibinfo{pages}{356--363} (\bibinfo{year}{2018}).

\bibitem{yamga2026llmphenotyping}
\bibinfo{author}{Yamga, E.}, \bibinfo{author}{Murphy, S.} \&
  \bibinfo{author}{Despr{\'e}s, P.}
\newblock \bibinfo{title}{A systematic exploration of {LLM} behavior for {EHR}
  phenotyping}.
\newblock \bibinfo{howpublished}{medRxiv} (\bibinfo{year}{2026}).
\newblock \urlprefix\url{https://doi.org/10.64898/2026.04.16.26350890}.

\bibitem{johnson2024mimiciv31}
\bibinfo{author}{Johnson, A.} \emph{et~al.}
\newblock \bibinfo{title}{{MIMIC-IV} (version 3.1)}.
\newblock \bibinfo{howpublished}{PhysioNet} (\bibinfo{year}{2024}).
\newblock \urlprefix\url{https://doi.org/10.13026/kpb9-mt58}.

\bibitem{pollard2019eicu20}
\bibinfo{author}{Pollard, T.} \emph{et~al.}
\newblock \bibinfo{title}{{eICU Collaborative Research Database} (version
  2.0)}.
\newblock \bibinfo{howpublished}{PhysioNet} (\bibinfo{year}{2019}).
\newblock \urlprefix\url{https://doi.org/10.13026/C2WM1R}.

\bibitem{kirby2016phekb}
\bibinfo{author}{Kirby, J.~C.} \emph{et~al.}
\newblock \bibinfo{title}{{PheKB}: a catalog and workflow for creating
  electronic phenotype algorithms for transportability}.
\newblock \emph{\bibinfo{journal}{Journal of the American Medical Informatics
  Association}} \textbf{\bibinfo{volume}{23}}, \bibinfo{pages}{1046--1052}
  (\bibinfo{year}{2016}).

\bibitem{quan2005coding}
\bibinfo{author}{Quan, H.} \emph{et~al.}
\newblock \bibinfo{title}{Coding algorithms for defining comorbidities in
  {ICD-9-CM} and {ICD-10} administrative data}.
\newblock \emph{\bibinfo{journal}{Medical Care}} \textbf{\bibinfo{volume}{43}},
  \bibinfo{pages}{1130--1139} (\bibinfo{year}{2005}).

\bibitem{horan2008nhsn}
\bibinfo{author}{Horan, T.~C.}, \bibinfo{author}{Andrus, M.} \&
  \bibinfo{author}{Dudeck, M.~A.}
\newblock \bibinfo{title}{{CDC/NHSN} surveillance definition of health
  care-associated infection and criteria for specific types of infections in
  the acute care setting}.
\newblock \emph{\bibinfo{journal}{American Journal of Infection Control}}
  \textbf{\bibinfo{volume}{36}}, \bibinfo{pages}{309--332}
  (\bibinfo{year}{2008}).

\bibitem{hanley1982meaning}
\bibinfo{author}{Hanley, J.~A.} \& \bibinfo{author}{McNeil, B.~J.}
\newblock \bibinfo{title}{The meaning and use of the area under a receiver
  operating characteristic ({ROC}) curve}.
\newblock \emph{\bibinfo{journal}{Radiology}} \textbf{\bibinfo{volume}{143}},
  \bibinfo{pages}{29--36} (\bibinfo{year}{1982}).

\bibitem{jiang2025medagentbench}
\bibinfo{author}{Jiang, Y.} \emph{et~al.}
\newblock \bibinfo{title}{{MedAgentBench}: a virtual {EHR} environment to
  benchmark medical {LLM} agents}.
\newblock \emph{\bibinfo{journal}{NEJM AI}} \textbf{\bibinfo{volume}{2}},
  \bibinfo{pages}{AIdbp2500144} (\bibinfo{year}{2025}).

\bibitem{gemmateam2026gemma4}
\bibinfo{author}{{Gemma Team}}.
\newblock \bibinfo{title}{{Gemma 4} technical report}.
\newblock \bibinfo{howpublished}{arXiv:2607.02770} (\bibinfo{year}{2026}).
\newblock \urlprefix\url{https://arxiv.org/abs/2607.02770}.

\bibitem{google2026gemma4card}
\bibinfo{author}{{Google}}.
\newblock \bibinfo{title}{{Gemma-4-26B-A4B}}.
\newblock \bibinfo{howpublished}{Hugging Face model card, Apache-2.0}
  (\bibinfo{year}{2026}).
\newblock \urlprefix\url{https://huggingface.co/google/gemma-4-26B-A4B}.

\bibitem{woo2025apixaban}
\bibinfo{author}{Woo, E.}, \bibinfo{author}{Burkhart, M.~C.},
  \bibinfo{author}{Alsentzer, E.} \& \bibinfo{author}{Beaulieu-Jones, B.}
\newblock \bibinfo{title}{{MIMIC-IV-Ext-Apixaban-Trial-Criteria-Questions}
  (version 1.0.0)}.
\newblock \bibinfo{howpublished}{PhysioNet} (\bibinfo{year}{2025}).
\newblock
  \urlprefix\url{https://physionet.org/content/mimic-iv-ext-apixaban-trial/1.0.0/}.

\bibitem{kweon2024ehrnoteqa}
\bibinfo{author}{Kweon, S.} \emph{et~al.}
\newblock \bibinfo{title}{{EHRNoteQA}: An {LLM} benchmark for real-world
  clinical practice using discharge summaries}.
\newblock In \emph{\bibinfo{booktitle}{Advances in Neural Information
  Processing Systems 37 (Datasets and Benchmarks Track)}},
  \bibinfo{pages}{124575--124611} (\bibinfo{year}{2024}).

\bibitem{kweon2024ehrnoteqadata}
\bibinfo{author}{Kweon, S.} \emph{et~al.}
\newblock \bibinfo{title}{{EHRNoteQA}: An {LLM} benchmark for real-world
  clinical practice using discharge summaries (version 1.0.1)}.
\newblock \bibinfo{howpublished}{PhysioNet} (\bibinfo{year}{2024}).
\newblock
  \urlprefix\url{https://physionet.org/content/ehr-notes-qa-llms/1.0.1/}.

\bibitem{moseley2020phenotype}
\bibinfo{author}{Moseley, E.}, \bibinfo{author}{Celi, L.~A.},
  \bibinfo{author}{Wu, J.} \& \bibinfo{author}{Dernoncourt, F.}
\newblock \bibinfo{title}{Phenotype annotations for patient notes in the
  {MIMIC-III} database}.
\newblock \bibinfo{howpublished}{PhysioNet, version 1.20.3}
  (\bibinfo{year}{2020}).
\newblock \urlprefix\url{https://doi.org/10.13026/txmt-8m40}.

\bibitem{johnson2016mimiciii}
\bibinfo{author}{Johnson, A. E.~W.} \emph{et~al.}
\newblock \bibinfo{title}{{MIMIC-III}, a freely accessible critical care
  database}.
\newblock \emph{\bibinfo{journal}{Scientific Data}}
  \textbf{\bibinfo{volume}{3}}, \bibinfo{pages}{160035} (\bibinfo{year}{2016}).

\bibitem{johnson2016mimiciiiv14}
\bibinfo{author}{Johnson, A.}, \bibinfo{author}{Pollard, T.} \&
  \bibinfo{author}{Mark, R.}
\newblock \bibinfo{title}{{MIMIC-III Clinical Database} (version 1.4)}.
\newblock \bibinfo{howpublished}{PhysioNet} (\bibinfo{year}{2016}).
\newblock \urlprefix\url{https://physionet.org/content/mimiciii/1.4/}.

\bibitem{bae2023ehrxqa}
\bibinfo{author}{Bae, S.} \emph{et~al.}
\newblock \bibinfo{title}{{EHRXQA}: A multi-modal question answering dataset
  for electronic health records with chest {X}-ray images}.
\newblock In \emph{\bibinfo{booktitle}{Advances in Neural Information
  Processing Systems 36, Datasets and Benchmarks Track}},
  \bibinfo{pages}{3867--3880} (\bibinfo{year}{2023}).

\bibitem{wu2021chestimagenome}
\bibinfo{author}{Wu, J.~T.} \emph{et~al.}
\newblock \bibinfo{title}{Chest {ImaGenome} dataset for clinical reasoning}.
\newblock In \emph{\bibinfo{booktitle}{Proceedings of the Neural Information
  Processing Systems Track on Datasets and Benchmarks}},
  vol.~\bibinfo{volume}{1} (\bibinfo{year}{2021}).
\newblock
  \urlprefix\url{https://datasets-benchmarks-proceedings.neurips.cc/paper/2021/hash/17e62166fc8586dfa4d1bc0e1742c08b-Abstract-round2.html}.

\bibitem{bae2024mimiccxrvqa}
\bibinfo{author}{Bae, S.} \emph{et~al.}
\newblock \bibinfo{title}{{MIMIC-Ext-MIMIC-CXR-VQA}: A complex, diverse, and
  large-scale visual question answering dataset for chest {X}-ray images
  (version 1.0.0)}.
\newblock \bibinfo{howpublished}{PhysioNet} (\bibinfo{year}{2024}).
\newblock
  \urlprefix\url{https://physionet.org/content/mimic-ext-mimic-cxr-vqa/1.0.0/}.

\bibitem{johnson2019mimiccxr}
\bibinfo{author}{Johnson, A. E.~W.} \emph{et~al.}
\newblock \bibinfo{title}{{MIMIC-CXR}, a de-identified publicly available
  database of chest radiographs with free-text reports}.
\newblock \emph{\bibinfo{journal}{Scientific Data}}
  \textbf{\bibinfo{volume}{6}}, \bibinfo{pages}{317} (\bibinfo{year}{2019}).

\bibitem{pal2025rexvqa}
\bibinfo{author}{Pal, A.} \emph{et~al.}
\newblock \bibinfo{title}{{ReXVQA}: A large-scale visual question answering
  benchmark for generalist chest {X}-ray understanding}.
\newblock In \emph{\bibinfo{booktitle}{Biocomputing 2026: Proceedings of the
  Pacific Symposium}}, \bibinfo{pages}{251--264} (\bibinfo{publisher}{World
  Scientific}, \bibinfo{year}{2026}).

\bibitem{weinstein2013tcga}
\bibinfo{author}{{Cancer Genome Atlas Research Network}} \emph{et~al.}
\newblock \bibinfo{title}{The {Cancer Genome Atlas} {Pan-Cancer} analysis
  project}.
\newblock \emph{\bibinfo{journal}{Nature Genetics}}
  \textbf{\bibinfo{volume}{45}}, \bibinfo{pages}{1113--1120}
  (\bibinfo{year}{2013}).

\bibitem{kefeli2024tcga}
\bibinfo{author}{Kefeli, J.} \& \bibinfo{author}{Tatonetti, N.}
\newblock \bibinfo{title}{{TCGA-Reports}: A machine-readable pathology report
  resource for benchmarking text-based {AI} models}.
\newblock \emph{\bibinfo{journal}{Patterns}} \textbf{\bibinfo{volume}{5}},
  \bibinfo{pages}{100933} (\bibinfo{year}{2024}).

\bibitem{grossman2016gdc}
\bibinfo{author}{Grossman, R.~L.} \emph{et~al.}
\newblock \bibinfo{title}{Toward a shared vision for cancer genomic data}.
\newblock \emph{\bibinfo{journal}{New England Journal of Medicine}}
  \textbf{\bibinfo{volume}{375}}, \bibinfo{pages}{1109--1112}
  (\bibinfo{year}{2016}).

\bibitem{nechaev2025histai}
\bibinfo{author}{Nechaev, D.}, \bibinfo{author}{Pchelnikov, A.} \&
  \bibinfo{author}{Ivanova, E.}
\newblock \bibinfo{title}{{HISTAI}: an open-source, large-scale whole slide
  image dataset for computational pathology}.
\newblock \bibinfo{howpublished}{arXiv preprint arXiv:2505.12120}
  (\bibinfo{year}{2025}).

\bibitem{yamga2026mipa}
\bibinfo{author}{Yamga, E.}, \bibinfo{author}{Goudrar, R.} \&
  \bibinfo{author}{Despr{\'e}s, P.}
\newblock \bibinfo{title}{{MIMIC-IV} {Phenotype Atlas} ({MIPA}): A publicly
  available dataset for {EHR} phenotyping}.
\newblock \bibinfo{howpublished}{medRxiv} (\bibinfo{year}{2026}).
\newblock \urlprefix\url{https://doi.org/10.64898/2026.04.16.26350888}.

\bibitem{vanaken2021assertion}
\bibinfo{author}{van Aken, B.} \emph{et~al.}
\newblock \bibinfo{title}{Assertion detection in clinical notes: medical
  language models to the rescue?}
\newblock In \emph{\bibinfo{booktitle}{Proceedings of the Second Workshop on
  Natural Language Processing for Medical Conversations}},
  \bibinfo{pages}{35--40} (\bibinfo{year}{2021}).

\bibitem{glm5team2026}
\bibinfo{author}{{GLM-5 Team}}.
\newblock \bibinfo{title}{{GLM-5}: from vibe coding to agentic engineering}.
\newblock \bibinfo{howpublished}{arXiv:2602.15763} (\bibinfo{year}{2026}).
\newblock \urlprefix\url{https://arxiv.org/abs/2602.15763}.

\bibitem{zheng2024robust}
\bibinfo{author}{Zheng, C.}, \bibinfo{author}{Zhou, H.}, \bibinfo{author}{Meng,
  F.}, \bibinfo{author}{Zhou, J.} \& \bibinfo{author}{Huang, M.}
\newblock \bibinfo{title}{Large language models are not robust multiple choice
  selectors}.
\newblock In \emph{\bibinfo{booktitle}{International Conference on Learning
  Representations (ICLR)}} (\bibinfo{year}{2024}).

\bibitem{pezeshkpour2024sensitivity}
\bibinfo{author}{Pezeshkpour, P.} \& \bibinfo{author}{Hruschka, E.}
\newblock \bibinfo{title}{Large language models sensitivity to the order of
  options in multiple-choice questions}.
\newblock In \emph{\bibinfo{booktitle}{Findings of the Association for
  Computational Linguistics: NAACL 2024}}, \bibinfo{pages}{2006--2017}
  (\bibinfo{year}{2024}).

\bibitem{inker2021ckdepi}
\bibinfo{author}{Inker, L.~A.} \emph{et~al.}
\newblock \bibinfo{title}{New creatinine- and cystatin {C}-based equations to
  estimate {GFR} without race}.
\newblock \emph{\bibinfo{journal}{New England Journal of Medicine}}
  \textbf{\bibinfo{volume}{385}}, \bibinfo{pages}{1737--1749}
  (\bibinfo{year}{2021}).

\bibitem{sterling2006fib4}
\bibinfo{author}{Sterling, R.~K.} \emph{et~al.}
\newblock \bibinfo{title}{Development of a simple noninvasive index to predict
  significant fibrosis in patients with {HIV/HCV} coinfection}.
\newblock \emph{\bibinfo{journal}{Hepatology}} \textbf{\bibinfo{volume}{43}},
  \bibinfo{pages}{1317--1325} (\bibinfo{year}{2006}).

\bibitem{qwen2026qwen35}
\bibinfo{author}{{Qwen Team}}.
\newblock \bibinfo{title}{{Qwen3.5}: Towards native multimodal agents}.
\newblock \bibinfo{howpublished}{Blog post} (\bibinfo{year}{2026}).
\newblock \urlprefix\url{https://qwen.ai/blog?id=qwen3.5}.

\bibitem{yang2025gated}
\bibinfo{author}{Yang, S.}, \bibinfo{author}{Kautz, J.} \&
  \bibinfo{author}{Hatamizadeh, A.}
\newblock \bibinfo{title}{Gated delta networks: Improving {Mamba2} with delta
  rule}.
\newblock In \emph{\bibinfo{booktitle}{International Conference on Learning
  Representations}} (\bibinfo{year}{2025}).

\bibitem{qiu2025gated}
\bibinfo{author}{Qiu, Z.} \emph{et~al.}
\newblock \bibinfo{title}{Gated attention for large language models:
  Non-linearity, sparsity, and attention-sink-free}.
\newblock In \emph{\bibinfo{booktitle}{Advances in Neural Information
  Processing Systems 38 (NeurIPS 2025)}}, \bibinfo{pages}{110931--110957}
  (\bibinfo{year}{2025}).

\bibitem{dong2024flex}
\bibinfo{author}{Dong, J.}, \bibinfo{author}{Feng, B.},
  \bibinfo{author}{Guessous, D.}, \bibinfo{author}{Liang, Y.} \&
  \bibinfo{author}{He, H.}
\newblock \bibinfo{title}{{FlexAttention}: A programming model for generating
  fused attention variants}.
\newblock In \emph{\bibinfo{booktitle}{Proceedings of Machine Learning and
  Systems}}, vol.~\bibinfo{volume}{7} (\bibinfo{year}{2025}).

\bibitem{krell2021efficient}
\bibinfo{author}{Krell, M.~M.}, \bibinfo{author}{Kosec, M.},
  \bibinfo{author}{Perez, S.~P.} \& \bibinfo{author}{Fitzgibbon, A.}
\newblock \bibinfo{title}{Efficient sequence packing without
  cross-contamination: Accelerating large language models without impacting
  performance}.
\newblock \bibinfo{howpublished}{arXiv:2107.02027} (\bibinfo{year}{2021}).

\bibitem{guo2017calibration}
\bibinfo{author}{Guo, C.}, \bibinfo{author}{Pleiss, G.}, \bibinfo{author}{Sun,
  Y.} \& \bibinfo{author}{Weinberger, K.~Q.}
\newblock \bibinfo{title}{On calibration of modern neural networks}.
\newblock In \emph{\bibinfo{booktitle}{Proceedings of the 34th International
  Conference on Machine Learning (ICML)}}, vol.~\bibinfo{volume}{70} of
  \emph{\bibinfo{series}{Proceedings of Machine Learning Research}},
  \bibinfo{pages}{1321--1330} (\bibinfo{year}{2017}).

\bibitem{gneiting2007strictly}
\bibinfo{author}{Gneiting, T.} \& \bibinfo{author}{Raftery, A.~E.}
\newblock \bibinfo{title}{Strictly proper scoring rules, prediction, and
  estimation}.
\newblock \emph{\bibinfo{journal}{Journal of the American Statistical
  Association}} \textbf{\bibinfo{volume}{102}}, \bibinfo{pages}{359--378}
  (\bibinfo{year}{2007}).

\bibitem{hinton2015distilling}
\bibinfo{author}{Hinton, G.}, \bibinfo{author}{Vinyals, O.} \&
  \bibinfo{author}{Dean, J.}
\newblock \bibinfo{title}{Distilling the knowledge in a neural network}.
\newblock \bibinfo{howpublished}{arXiv:1503.02531} (\bibinfo{year}{2015}).

\bibitem{loshchilov2019adamw}
\bibinfo{author}{Loshchilov, I.} \& \bibinfo{author}{Hutter, F.}
\newblock \bibinfo{title}{Decoupled weight decay regularization}.
\newblock In \emph{\bibinfo{booktitle}{International Conference on Learning
  Representations (ICLR)}} (\bibinfo{year}{2019}).
\newblock \urlprefix\url{https://openreview.net/forum?id=Bkg6RiCqY7}.

\bibitem{qwen2026qwen38}
\bibinfo{author}{{Qwen Team}}.
\newblock \bibinfo{title}{{Qwen3.8-27B}}.
\newblock \bibinfo{howpublished}{Hugging Face model card}
  (\bibinfo{year}{2026}).
\newblock \urlprefix\url{https://huggingface.co/Qwen/Qwen3.8-27B}.

\bibitem{wang2024fair}
\bibinfo{author}{Wang, P.} \emph{et~al.}
\newblock \bibinfo{title}{Large language models are not fair evaluators}.
\newblock In \emph{\bibinfo{booktitle}{Proceedings of the 62nd Annual Meeting
  of the Association for Computational Linguistics (Volume 1: Long Papers)}},
  \bibinfo{pages}{9440--9450} (\bibinfo{year}{2024}).

\bibitem{efron1979bootstrap}
\bibinfo{author}{Efron, B.}
\newblock \bibinfo{title}{Bootstrap methods: Another look at the jackknife}.
\newblock \emph{\bibinfo{journal}{The Annals of Statistics}}
  \textbf{\bibinfo{volume}{7}}, \bibinfo{pages}{1--26} (\bibinfo{year}{1979}).

\bibitem{li2016erasure}
\bibinfo{author}{Li, J.}, \bibinfo{author}{Monroe, W.} \&
  \bibinfo{author}{Jurafsky, D.}
\newblock \bibinfo{title}{Understanding neural networks through representation
  erasure}.
\newblock \bibinfo{howpublished}{arXiv:1612.08220} (\bibinfo{year}{2016}).
\newblock \urlprefix\url{https://arxiv.org/abs/1612.08220}.

\bibitem{pan2025marconi}
\bibinfo{author}{Pan, R.} \emph{et~al.}
\newblock \bibinfo{title}{Marconi: Prefix caching for the era of hybrid
  {LLMs}}.
\newblock In \emph{\bibinfo{booktitle}{Proceedings of Machine Learning and
  Systems}}, vol.~\bibinfo{volume}{7} (\bibinfo{year}{2025}).

\bibitem{samek2017evaluating}
\bibinfo{author}{Samek, W.}, \bibinfo{author}{Binder, A.},
  \bibinfo{author}{Montavon, G.}, \bibinfo{author}{Lapuschkin, S.} \&
  \bibinfo{author}{M{\"u}ller, K.-R.}
\newblock \bibinfo{title}{Evaluating the visualization of what a deep neural
  network has learned}.
\newblock \emph{\bibinfo{journal}{IEEE Transactions on Neural Networks and
  Learning Systems}} \textbf{\bibinfo{volume}{28}}, \bibinfo{pages}{2660--2673}
  (\bibinfo{year}{2017}).

\bibitem{brodersen2010balanced}
\bibinfo{author}{Brodersen, K.~H.}, \bibinfo{author}{Ong, C.~S.},
  \bibinfo{author}{Stephan, K.~E.} \& \bibinfo{author}{Buhmann, J.~M.}
\newblock \bibinfo{title}{The balanced accuracy and its posterior
  distribution}.
\newblock In \emph{\bibinfo{booktitle}{Proceedings of the 20th International
  Conference on Pattern Recognition}}, \bibinfo{pages}{3121--3124}
  (\bibinfo{publisher}{IEEE}, \bibinfo{year}{2010}).

\bibitem{naeini2015obtaining}
\bibinfo{author}{Naeini, M.~P.}, \bibinfo{author}{Cooper, G.~F.} \&
  \bibinfo{author}{Hauskrecht, M.}
\newblock \bibinfo{title}{Obtaining well calibrated probabilities using
  {Bayesian} binning}.
\newblock In \emph{\bibinfo{booktitle}{Proceedings of the Twenty-Ninth AAAI
  Conference on Artificial Intelligence}}, \bibinfo{pages}{2901--2907}
  (\bibinfo{year}{2015}).

\bibitem{nixon2019measuring}
\bibinfo{author}{Nixon, J.}, \bibinfo{author}{Dusenberry, M.~W.},
  \bibinfo{author}{Zhang, L.}, \bibinfo{author}{Jerfel, G.} \&
  \bibinfo{author}{Tran, D.}
\newblock \bibinfo{title}{Measuring calibration in deep learning}.
\newblock In \emph{\bibinfo{booktitle}{Proceedings of the IEEE/CVF Conference
  on Computer Vision and Pattern Recognition (CVPR) Workshops}},
  \bibinfo{pages}{38--41} (\bibinfo{year}{2019}).

\bibitem{brier1950verification}
\bibinfo{author}{Brier, G.~W.}
\newblock \bibinfo{title}{Verification of forecasts expressed in terms of
  probability}.
\newblock \emph{\bibinfo{journal}{Monthly Weather Review}}
  \textbf{\bibinfo{volume}{78}}, \bibinfo{pages}{1--3} (\bibinfo{year}{1950}).

\bibitem{saito2015precision}
\bibinfo{author}{Saito, T.} \& \bibinfo{author}{Rehmsmeier, M.}
\newblock \bibinfo{title}{The precision-recall plot is more informative than
  the {ROC} plot when evaluating binary classifiers on imbalanced datasets}.
\newblock \emph{\bibinfo{journal}{PLoS ONE}} \textbf{\bibinfo{volume}{10}},
  \bibinfo{pages}{e0118432} (\bibinfo{year}{2015}).

\bibitem{steyerberg2010assessing}
\bibinfo{author}{Steyerberg, E.~W.} \emph{et~al.}
\newblock \bibinfo{title}{Assessing the performance of prediction models: a
  framework for traditional and novel measures}.
\newblock \emph{\bibinfo{journal}{Epidemiology}} \textbf{\bibinfo{volume}{21}},
  \bibinfo{pages}{128--138} (\bibinfo{year}{2010}).

\bibitem{vancalster2019calibration}
\bibinfo{author}{Van~Calster, B.}, \bibinfo{author}{McLernon, D.~J.},
  \bibinfo{author}{van Smeden, M.}, \bibinfo{author}{Wynants, L.} \&
  \bibinfo{author}{Steyerberg, E.~W.}
\newblock \bibinfo{title}{Calibration: the {Achilles} heel of predictive
  analytics}.
\newblock \emph{\bibinfo{journal}{BMC Medicine}} \textbf{\bibinfo{volume}{17}},
  \bibinfo{pages}{230} (\bibinfo{year}{2019}).

\bibitem{chow1970optimum}
\bibinfo{author}{Chow, C.~K.}
\newblock \bibinfo{title}{On optimum recognition error and reject tradeoff}.
\newblock \emph{\bibinfo{journal}{IEEE Transactions on Information Theory}}
  \textbf{\bibinfo{volume}{16}}, \bibinfo{pages}{41--46}
  (\bibinfo{year}{1970}).

\bibitem{geifman2017selective}
\bibinfo{author}{Geifman, Y.} \& \bibinfo{author}{El-Yaniv, R.}
\newblock \bibinfo{title}{Selective classification for deep neural networks}.
\newblock In \emph{\bibinfo{booktitle}{Advances in Neural Information
  Processing Systems 30}}, \bibinfo{pages}{4878--4887} (\bibinfo{year}{2017}).

\bibitem{field2007bootstrapping}
\bibinfo{author}{Field, C.~A.} \& \bibinfo{author}{Welsh, A.~H.}
\newblock \bibinfo{title}{Bootstrapping clustered data}.
\newblock \emph{\bibinfo{journal}{Journal of the Royal Statistical Society:
  Series B (Statistical Methodology)}} \textbf{\bibinfo{volume}{69}},
  \bibinfo{pages}{369--390} (\bibinfo{year}{2007}).

\bibitem{rubin1981bayesian}
\bibinfo{author}{Rubin, D.~B.}
\newblock \bibinfo{title}{The {B}ayesian bootstrap}.
\newblock \emph{\bibinfo{journal}{The Annals of Statistics}}
  \textbf{\bibinfo{volume}{9}}, \bibinfo{pages}{130--134}
  (\bibinfo{year}{1981}).

\bibitem{wolf2020transformers}
\bibinfo{author}{Wolf, T.} \emph{et~al.}
\newblock \bibinfo{title}{Transformers: State-of-the-art natural language
  processing}.
\newblock In \emph{\bibinfo{booktitle}{Proceedings of the 2020 Conference on
  Empirical Methods in Natural Language Processing: System Demonstrations}},
  \bibinfo{pages}{38--45} (\bibinfo{publisher}{Association for Computational
  Linguistics}, \bibinfo{year}{2020}).

\bibitem{pollard2026physionet}
\bibinfo{author}{Pollard, T.} \emph{et~al.}
\newblock \bibinfo{title}{{PhysioNet} as a global platform for biomedical
  research}.
\newblock \emph{\bibinfo{journal}{Nature Health}} \textbf{\bibinfo{volume}{1}},
  \bibinfo{pages}{792--795} (\bibinfo{year}{2026}).

\bibitem{paszke2019pytorch}
\bibinfo{author}{Paszke, A.} \emph{et~al.}
\newblock \bibinfo{title}{{PyTorch}: An imperative style, high-performance deep
  learning library}.
\newblock In \emph{\bibinfo{booktitle}{Advances in Neural Information
  Processing Systems 32 (NeurIPS 2019)}}, \bibinfo{pages}{8026--8037}
  (\bibinfo{year}{2019}).

\bibitem{ansel2024pytorch2}
\bibinfo{author}{Ansel, J.}, \bibinfo{author}{Yang, E.}, \bibinfo{author}{He,
  H.} \emph{et~al.}
\newblock \bibinfo{title}{{PyTorch 2}: Faster machine learning through dynamic
  {Python} bytecode transformation and graph compilation}.
\newblock In \emph{\bibinfo{booktitle}{Proceedings of the 29th ACM
  International Conference on Architectural Support for Programming Languages
  and Operating Systems, Volume 2 (ASPLOS '24)}}, \bibinfo{pages}{929--947}
  (\bibinfo{publisher}{ACM}, \bibinfo{year}{2024}).

\bibitem{yang2024fla}
\bibinfo{author}{Yang, S.} \& \bibinfo{author}{Zhang, Y.}
\newblock \bibinfo{title}{{FLA}: A {Triton}-based library for
  hardware-efficient implementations of linear attention mechanism}.
\newblock \bibinfo{howpublished}{GitHub repository} (\bibinfo{year}{2024}).
\newblock \urlprefix\url{https://github.com/fla-org/flash-linear-attention}.

\bibitem{mangrulkar2022peft}
\bibinfo{author}{Mangrulkar, S.} \emph{et~al.}
\newblock \bibinfo{title}{{PEFT}: State-of-the-art parameter-efficient
  fine-tuning methods}.
\newblock \bibinfo{howpublished}{GitHub repository} (\bibinfo{year}{2022}).
\newblock \urlprefix\url{https://github.com/huggingface/peft}.

\bibitem{harris2020numpy}
\bibinfo{author}{Harris, C.~R.} \emph{et~al.}
\newblock \bibinfo{title}{Array programming with {NumPy}}.
\newblock \emph{\bibinfo{journal}{Nature}} \textbf{\bibinfo{volume}{585}},
  \bibinfo{pages}{357--362} (\bibinfo{year}{2020}).

\bibitem{virtanen2020scipy}
\bibinfo{author}{Virtanen, P.} \emph{et~al.}
\newblock \bibinfo{title}{{SciPy} 1.0: fundamental algorithms for scientific
  computing in {Python}}.
\newblock \emph{\bibinfo{journal}{Nature Methods}}
  \textbf{\bibinfo{volume}{17}}, \bibinfo{pages}{261--272}
  (\bibinfo{year}{2020}).

\bibitem{pedregosa2011scikit}
\bibinfo{author}{Pedregosa, F.} \emph{et~al.}
\newblock \bibinfo{title}{Scikit-learn: Machine learning in {Python}}.
\newblock \emph{\bibinfo{journal}{Journal of Machine Learning Research}}
  \textbf{\bibinfo{volume}{12}}, \bibinfo{pages}{2825--2830}
  (\bibinfo{year}{2011}).
\newblock \urlprefix\url{http://jmlr.org/papers/v12/pedregosa11a.html}.

\bibitem{mckinney2010pandas}
\bibinfo{author}{McKinney, W.}
\newblock \bibinfo{title}{Data structures for statistical computing in
  {Python}}.
\newblock In \bibinfo{editor}{van~der Walt, S.} \& \bibinfo{editor}{Millman,
  J.} (eds.) \emph{\bibinfo{booktitle}{Proceedings of the 9th Python in Science
  Conference}}, \bibinfo{pages}{56--61} (\bibinfo{year}{2010}).

\bibitem{raasveldt2019duckdb}
\bibinfo{author}{Raasveldt, M.} \& \bibinfo{author}{M{\"u}hleisen, H.}
\newblock \bibinfo{title}{{DuckDB}: an embeddable analytical database}.
\newblock In \emph{\bibinfo{booktitle}{Proceedings of the 2019 International
  Conference on Management of Data (SIGMOD '19)}}, \bibinfo{pages}{1981--1984}
  (\bibinfo{publisher}{ACM}, \bibinfo{year}{2019}).

\bibitem{hunter2007matplotlib}
\bibinfo{author}{Hunter, J.~D.}
\newblock \bibinfo{title}{Matplotlib: A {2D} graphics environment}.
\newblock \emph{\bibinfo{journal}{Computing in Science \& Engineering}}
  \textbf{\bibinfo{volume}{9}}, \bibinfo{pages}{90--95} (\bibinfo{year}{2007}).

\bibitem{ye2025flashinfer}
\bibinfo{author}{Ye, Z.} \emph{et~al.}
\newblock \bibinfo{title}{{FlashInfer}: Efficient and customizable attention
  engine for {LLM} inference serving}.
\newblock In \emph{\bibinfo{booktitle}{Proceedings of Machine Learning and
  Systems}}, vol.~\bibinfo{volume}{7} (\bibinfo{year}{2025}).

\bibitem{goldberger2000physionet}
\bibinfo{author}{Goldberger, A.~L.} \emph{et~al.}
\newblock \bibinfo{title}{{PhysioBank}, {PhysioToolkit}, and {PhysioNet}:
  Components of a new research resource for complex physiologic signals}.
\newblock \emph{\bibinfo{journal}{Circulation}} \textbf{\bibinfo{volume}{101}},
  \bibinfo{pages}{e215--e220} (\bibinfo{year}{2000}).

\end{thebibliography}

\begin{addendum}\setlength{\itemsep}{5pt}%
\item[Data availability] The credentialed data sets used for evaluation (PhysioNet\cite{goldberger2000physionet,pollard2026physionet}, EHRSHOT and others) remain available from their providers under their data-use agreements. Following institution policies, all requests for data collected or curated in-house will be evaluated on a case-by-case basis. Data can only be shared for academic research purposes and will require a material transfer agreement.

\item[Code availability] The code (inference and training code, and code to run selected benchmarks) and the \mainmodel{} weights will be released upon institutional approval.

\item[Acknowledgements] This work was funded in part by the Brigham and Women's Hospital (BWH) and Massachusetts General Hospital (MGH) internal funds. R.K. is supported by a Doctoral Research Award from the Canadian Institutes of Health Research (FRN 209970).

\item[Author contributions] A.Z., F.M. conceived the study and designed the experiments. A.Z., T.T.L.V., S.J.W., D.B., T.D., Y.S., M.C., C.S.C., J.L., R.K., M.Y.L., C.T., T.L., J.M. curated the training data. A.Z., M.C., J.L., R.K., C.S.C. curated and evaluated the benchmarks. A.Z. developed the model architecture and training pipeline. A.Z., Y.S., C.T., J.L., T.D. designed the patient-QA and trial screening experiments. A.Z., S.J.W., M.C., T.T.L.V., D.B., C.S.C., F.M. prepared the manuscript. All authors contributed to the writing. L.P.L., F.M. supervised the research.
\end{addendum}

\EDsetup
\section*{Extended Data}

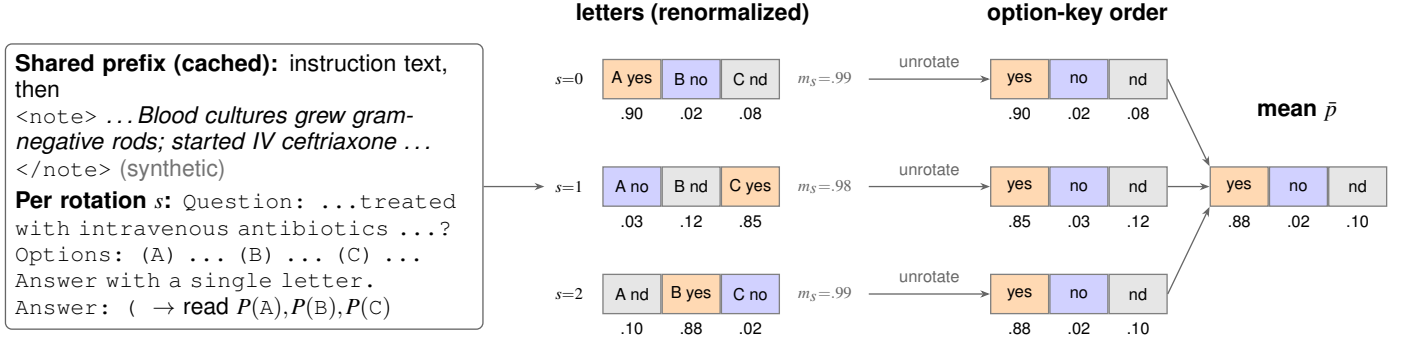
\begin{figure}[H]
\centering
\resizebox{\linewidth}{!}{%
\begin{tikzpicture}[
  font=\sffamily\scriptsize,
  box/.style={draw=black!60,rounded corners=2pt,align=left,inner sep=3pt},
  cell/.style={draw=black!50,minimum width=0.62cm,minimum height=0.42cm,inner sep=0pt,font=\sffamily\tiny},
  kyes/.style={fill=orange!30}, kno/.style={fill=blue!18}, knd/.style={fill=black!10},
  arr/.style={-{Stealth[length=3pt]},black!60},
]
\node[box,text width=4.9cm] (prompt) at (0,0) {%
  \textbf{Shared prefix (cached):} instruction text, then\\
  \texttt{<note>} \emph{\dots Blood cultures grew gram-negative rods; started IV ceftriaxone \dots} \texttt{</note>} {\color{black!55}(synthetic)}\\[2pt]
  \textbf{Per rotation $s$:} \texttt{Question: \dots treated with intravenous antibiotics \dots?}\\
  \texttt{Options: (A) \dots\ (B) \dots\ (C) \dots}\\
  \texttt{Answer with a single letter.}\\
  \texttt{Answer: (}\ \ $\rightarrow$ read $P(\texttt{A}),P(\texttt{B}),P(\texttt{C})$};

\def\rx{3.8}
\foreach \s/\y/\la/\lb/\lc/\sa/\sb/\sc/\pa/\pb/\pc/\m in {
    0/1.15/yes/no/nd/kyes/kno/knd/.90/.02/.08/.99,
    1/0/no/nd/yes/kno/knd/kyes/.03/.12/.85/.98,
    2/-1.15/nd/yes/no/knd/kyes/kno/.10/.88/.02/.99} {
  \node[anchor=east,font=\sffamily\tiny] at (\rx-0.05,\y) {$s{=}\s$};
  \node[cell,\sa] (a\s) at (\rx+0.35,\y) {A \la};
  \node[cell,\sb,right=0pt of a\s] (b\s) {B \lb};
  \node[cell,\sc,right=0pt of b\s] (c\s) {C \lc};
  \node[font=\sffamily\tiny,below=-1pt of a\s] {\pa};
  \node[font=\sffamily\tiny,below=-1pt of b\s] {\pb};
  \node[font=\sffamily\tiny,below=-1pt of c\s] {\pc};
  \node[anchor=west,font=\sffamily\tiny,black!60] at ([xshift=2pt]c\s.east) {$m_s{=}\m$};
}
\node[font=\sffamily\scriptsize\bfseries] at (\rx+1.0,1.85) {letters (renormalized)};
\draw[arr] (prompt.east) -- ([xshift=-0.62cm]a1.west);

\def\kx{8.3}
\foreach \s/\y/\py/\pn/\pd in {0/1.15/.90/.02/.08, 1/0/.85/.03/.12, 2/-1.15/.88/.02/.10} {
  \node[cell,kyes] (ky\s) at (\kx,\y) {yes};
  \node[cell,kno,right=0pt of ky\s] (kn\s) {no};
  \node[cell,knd,right=0pt of kn\s] (kd\s) {nd};
  \node[font=\sffamily\tiny,below=-1pt of ky\s] {\py};
  \node[font=\sffamily\tiny,below=-1pt of kn\s] {\pn};
  \node[font=\sffamily\tiny,below=-1pt of kd\s] {\pd};
  \draw[arr] ([xshift=0.95cm]c\s.east) -- node[above,font=\sffamily\tiny]{unrotate} (ky\s.west);
}
\node[font=\sffamily\scriptsize\bfseries] at (\kx+0.62,1.85) {option-key order};

\node[cell,kyes] (ay) at (\kx+2.35,0) {yes};
\node[cell,kno,right=0pt of ay] (an) {no};
\node[cell,knd,right=0pt of an] (ad) {nd};
\node[font=\sffamily\tiny,below=-1pt of ay] {.88};
\node[font=\sffamily\tiny,below=-1pt of an] {.02};
\node[font=\sffamily\tiny,below=-1pt of ad] {.10};
\node[font=\sffamily\scriptsize\bfseries,align=center] at (\kx+2.97,0.8) {mean $\bar p$};
\draw[arr] (kd0.east) -- (ay.north west);
\draw[arr] (kd1.east) -- (ay.west);
\draw[arr] (kd2.east) -- (ay.south west);
\end{tikzpicture}}%
\caption{\textbf{Logit-readout teacher.} For one document--question pair (synthetic excerpt; illustrative probabilities), the teacher (GLM-5.3-Flash) receives one raw completion prompt per cyclic rotation of the options and generates no text: the probabilities of the valid letters are read from one next-token distribution after the prefill \texttt{Answer: (}. Rotation $s$ lists the option keys starting from the $s$-th, each with its criterion. Letter probabilities are renormalized over the valid letters (their sum, the valid-letter mass, is logged as a quality check), mapped back to option-key order and averaged over the $n$ rotations, which cancels purely positional bias. A temperature can then sharpen the under-confident average; the \dataset{} training labels use none. For \institution{} data, the teacher runs on institutional servers.}
\label{edfig:readout}
\end{figure}

\begin{figure}[H]
\centering
\resizebox{\linewidth}{!}{%
\begin{tikzpicture}[font=\sffamily\scriptsize,
  tok/.style={draw=black!60,minimum height=0.6cm,inner sep=1pt,anchor=west,align=center},
  dc/.style={fill=blue!12}, q1/.style={fill=orange!22}, op/.style={fill=orange!40},
  q2/.style={fill=green!18}, q3/.style={fill=violet!14},
  ep/.style={fill=black!80,minimum width=0.2cm}, qy/.style={fill=red!65!black,minimum width=0.2cm},
  st/.style={circle,draw=black!70,fill=blue!30,inner sep=1pt,minimum size=0.55cm,font=\sffamily\tiny\bfseries},
  ar/.style={-{Stealth[length=1.6mm]},black!70}]
\node[tok,dc,minimum width=3.6cm] (d) at (0,0) {\texttt{Context:}\\document};
\node[tok,q1,minimum width=1.75cm,right=3pt of d] (h1) {\texttt{Task type:} \dots\\\texttt{Options:}};
\node[tok,op,minimum width=0.8cm,right=0pt of h1] (o1a) {option A};
\node[tok,ep,right=0pt of o1a] (e1a) {};
\node[tok,op,minimum width=0.8cm,right=0pt of e1a] (o1b) {option B};
\node[tok,ep,right=0pt of o1b] (e1b) {};
\node[tok,op,minimum width=0.8cm,right=0pt of e1b] (o1c) {option C};
\node[tok,ep,right=0pt of o1c] (e1c) {};
\node[tok,q1,minimum width=1.25cm,right=0pt of e1c] (s1) {\dots\\\texttt{Decision:}};
\node[tok,qy,right=0pt of s1] (k1) {};
\node[tok,q2,minimum width=1.3cm,right=3pt of k1] (s2) {Q2 \dots\\\texttt{Decision:}};
\node[tok,qy,right=0pt of s2] (k2) {};
\node[tok,q3,minimum width=1.3cm,right=3pt of k2] (s3) {Q3 \dots\\\texttt{Decision:}};
\node[tok,qy,right=0pt of s3] (k3) {};
\node[anchor=south west,font=\sffamily\scriptsize\bfseries] at ([yshift=0.75cm]d.north west) {(a) one packed sequence: the document prefix once, then one segment per question};
\draw[decorate,decoration={brace,amplitude=3pt},black!60] ([yshift=2pt]d.north west) -- node[above=2pt,font=\sffamily\tiny] {document prefix} ([yshift=2pt]d.north east);
\draw[decorate,decoration={brace,amplitude=3pt},black!60] ([yshift=2pt]h1.north west) -- node[above=2pt,font=\sffamily\tiny] {segment of question 1: header, option blocks, \texttt{Select the single option \dots\ Decision:}} ([yshift=2pt]k1.north east);
\node[below=1pt of d,font=\sffamily\tiny] {positions $0,\dots,L-1$};
\node[below=1pt of h1,font=\sffamily\tiny] {$L,\dots$};
\node[below=1pt of s2,font=\sffamily\tiny] {$L,\dots$};
\node[below=1pt of s3,font=\sffamily\tiny] {$L,\dots$};
\node[draw=black!70,fill=black!5,rounded corners=2pt,minimum height=0.6cm,text width=6.2cm,align=center,font=\sffamily\tiny]
  (head) at ($(o1a.west)!0.5!(k1.east)+(0,-1.75)$)
  {\textbf{CandidateHead} (FP32): each option endpoint's hidden state with the query state\\$\rightarrow$ one logit per option (A, B, C) $\rightarrow$ softmax over the question's options};
\foreach \e in {e1a,e1b,e1c} {\draw[ar] (\e.south) -- (\e.south |- head.north);}
\draw[ar,red!65!black] (k1.south) -- (k1.south |- head.north);
\node[tok,ep,minimum height=0.22cm] (lg1) at (0,-1.45) {};
\node[anchor=west,font=\sffamily\tiny] at (lg1.east) {option endpoint: last token of an option block};
\node[tok,qy,minimum height=0.22cm] (lg2) at (0,-1.85) {};
\node[anchor=west,font=\sffamily\tiny] at (lg2.east) {query: the segment's last token (after \texttt{Decision:})};
\node[anchor=north west,font=\sffamily\tiny,text width=4.6cm,align=left] at (0,-2.15)
  {Option block: \texttt{<option>} \{\texttt{"description":} criterion, \texttt{"key":"A"}\} \texttt{</option>}};
\begin{scope}[shift={(0.9,-7.4)},scale=0.5]
\fill[blue!30] (0,6) -- (3,3) -- (0,3) -- cycle;
\fill[orange!45] (0,2) rectangle (3,3); \fill[orange!45] (3,3) -- (4,2) -- (3,2) -- cycle;
\fill[green!40] (0,1.2) rectangle (3,2); \fill[green!40] (4,2) -- (4.8,1.2) -- (4,1.2) -- cycle;
\fill[violet!35] (0,0) rectangle (3,1.2); \fill[violet!35] (4.8,1.2) -- (6,0) -- (4.8,0) -- cycle;
\draw[black!70] (0,0) rectangle (6,6);
\foreach \x in {3,4,4.8} {\draw[black!40] (\x,0) -- (\x,6);}
\foreach \y in {3,2,1.2} {\draw[black!40] (0,\y) -- (6,\y);}
\node[above] at (1.5,6) {doc}; \node[above] at (3.5,6) {Q1}; \node[above] at (4.4,6) {Q2}; \node[above] at (5.4,6) {Q3};
\node[left] at (0,4.5) {doc}; \node[left] at (0,2.5) {Q1}; \node[left] at (0,1.6) {Q2}; \node[left] at (0,0.6) {Q3};
\node[anchor=west,font=\sffamily\scriptsize\bfseries] at (-1.6,8.1) {(b) full-attention layers (8 of 32)};
\node[anchor=west,font=\sffamily\tiny] at (-1.6,7.45) {block attention mask (white = masked)};
\end{scope}
\begin{scope}[shift={(5.6,0)}]
\node[anchor=west,font=\sffamily\scriptsize\bfseries] at (0,-3.35) {(c) Gated DeltaNet layers (24 of 32): state handoff};
\node[st,fill=white] (z) at (0.3,-5.0) {$0$};
\node[tok,dc,minimum width=3.0cm,right=4pt of z] (gd) {document (run once)};
\node[st] (sd) at ($(gd.east)+(0.75,0)$) {$S_D$};
\draw[ar] (z) -- (gd);
\draw[ar] (gd) -- (sd);
\node[st] (c1) at ($(sd)+(1.55,0.75)$) {$S_D$};
\node[st] (c2) at ($(sd)+(1.55,0)$) {$S_D$};
\node[st] (c3) at ($(sd)+(1.55,-0.75)$) {$S_D$};
\foreach \c in {c1,c2,c3} {\draw[ar,dashed] (sd) -- (\c);}
\node[tok,q1,minimum width=1.6cm,minimum height=0.45cm,right=4pt of c1] (g1) {Q1 segment};
\node[tok,q2,minimum width=1.6cm,minimum height=0.45cm,right=4pt of c2] (g2) {Q2 segment};
\node[tok,q3,minimum width=1.6cm,minimum height=0.45cm,right=4pt of c3] (g3) {Q3 segment};
\foreach \c/\g in {c1/g1,c2/g2,c3/g3} {\draw[ar] (\c) -- (\g);}
\node[font=\sffamily\tiny,align=left,anchor=north west] at ($(sd)+(-0.5,-0.55)$) {final state\\copied to\\every segment};
\node[anchor=north west,font=\sffamily\tiny,text width=8.6cm,align=left] at (0,-6.35)
  {The gated delta rule runs over the document once from a zero state, and its final recurrent state $S_D$ is kept. Every segment then runs from a copy of $S_D$ (variable-length kernel with initial states), never from another segment's state. The causal convolution (four-token window) at the first three tokens of each segment is recomputed with the document's last tokens as left context, so no segment reads another. When several documents share a training sequence, each starts from a zero state and its segments start from its own final state.};
\end{scope}
\end{tikzpicture}}%
\caption{\textbf{Shared-document packing on the hybrid backbone of \mainmodel{}.} \textbf{a}, One packed sequence holds the document prefix once, in the Decision 2.0 runtime's prompt format (\texttt{Context:} and the document), followed by one segment per question: the task type and question, one block per option, and a closing line ending in \texttt{Decision:}. Segment position indices restart at the end of the document, so the prefix and any one segment are token for token the runtime's input for that question. The candidate head takes the hidden states at the last token of every option block (option endpoints) and at the segment's last token (query) and returns one logit per option. \textbf{b}, The block attention mask of the full-attention layers: causal within the document; each segment attends to the whole document and causally to itself, never to another segment. \textbf{c}, The Gated DeltaNet layers run the document once from a zero state and copy its final recurrent state to the start of every segment; the causal convolution at each segment's first tokens takes its left context from the document's last tokens. Every question therefore sees the document exactly as in a single-question prompt, its answer does not depend on the other questions, and an extra question costs only its own tokens. The same structure supports deletion attribution: because the text before a deleted sentence is unchanged, the model's state up to it is computed once and reused, and only the rest of the document and the questions are re-encoded (Methods).}
\label{edfig:packing}
\end{figure}

\begin{figure}[H]
\centering
\figinclude[width=\linewidth]{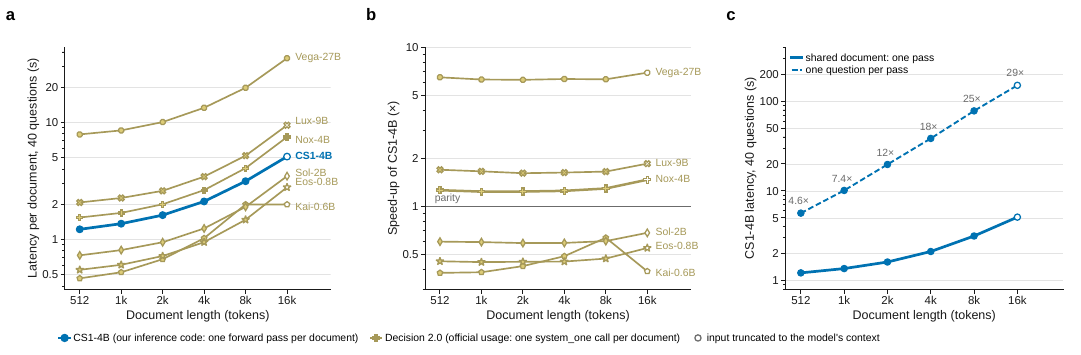}
\caption{\textbf{Speed against document length.} \textbf{a}, Latency per document at 40 questions on one idle NVIDIA GB10 for synthetic documents of 512 to 16,384 tokens: \mainmodel{} and the six Decision 2.0 models, each through its released inference code (Decision 2.0: one \texttt{system\_one} call per document, not further optimized; Methods); median of 10 timed repetitions (5 for runs longer than 5\,s). Open markers, input truncated to the model's context (at 16,384 tokens; Kai-0.6B at 8,192). \textbf{b}, \mainmodel{}'s speed-up over each Decision 2.0 model (ratio of latencies; 1, parity). \textbf{c}, \mainmodel{} reading the document once for all 40 questions (its default) against one forward pass per question, with the same weights; both modes gave the same top answers.}
\label{edfig:speed}
\end{figure}

\begin{figure}[H]
\centering
\figinclude[width=\linewidth]{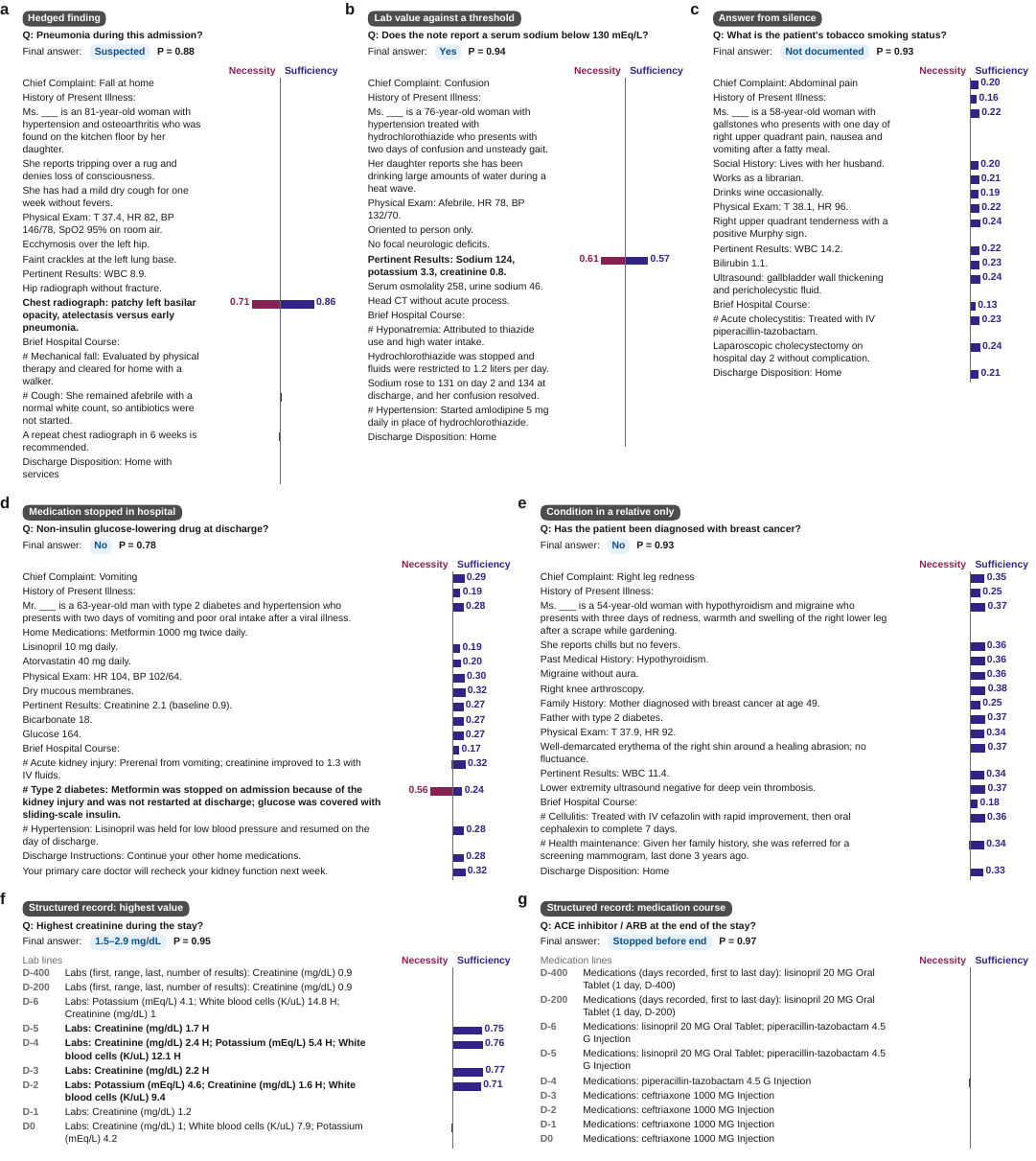}
\caption{\textbf{More evidence attributions by \mainmodel{} on synthetic documents.} Drawn as \cref{fig:attribution}d--f: necessity left and sufficiency right, as defined in \cref{fig:attribution}a; values below zero (a sentence that lowers the answer's probability) are drawn as zero; bold, the sentences that carry the answer. All text is synthetic, and the bars are model attributions, not ground truth. \textbf{a}, A hedged finding (pneumonia this admission; suspected, $P=0.88$): one radiograph sentence, ``atelectasis versus early pneumonia'', carries the answer (necessity 0.71, sufficiency 0.86). \textbf{b}, A laboratory value against a threshold (sodium below 130\,mEq/L; yes, $P=0.94$): the sentence with sodium 124 (necessity 0.61, sufficiency 0.57); the later values of 131 and 134 alone point the other way. \textbf{c}, An answer from silence (smoking status; not documented, $P=0.93$): no sentence is necessary, and every sentence alone raises the answer a little, because none mentions smoking. \textbf{d}, A medication stopped in hospital (non-insulin glucose-lowering drug at discharge; no, $P=0.78$): deleting the sentence that metformin was stopped and not restarted flips the answer (necessity 0.56). \textbf{e}, A condition in a relative only (breast cancer; no, $P=0.93$): no sentence is necessary; the family-history sentence does not move the answer towards yes. \textbf{f},\,\textbf{g}, The structured record of \cref{fig:attribution}f, other questions. \textbf{f}, Highest creatinine of the stay (1.5--2.9\,mg/dL, $P=0.95$): the measurements are redundant; no line is necessary, and each of the four elevated values alone gives the answer (sufficiency 0.71--0.77). \textbf{g}, ACE inhibitor or ARB at the end of the stay (stopped before the end, $P=0.97$): no single line moves the answer, which rests on the pattern across days, a limit of single-sentence attribution.}
\label{edfig:attribution}
\end{figure}

\begin{figure}[H]
\centering
\figinclude[width=\linewidth]{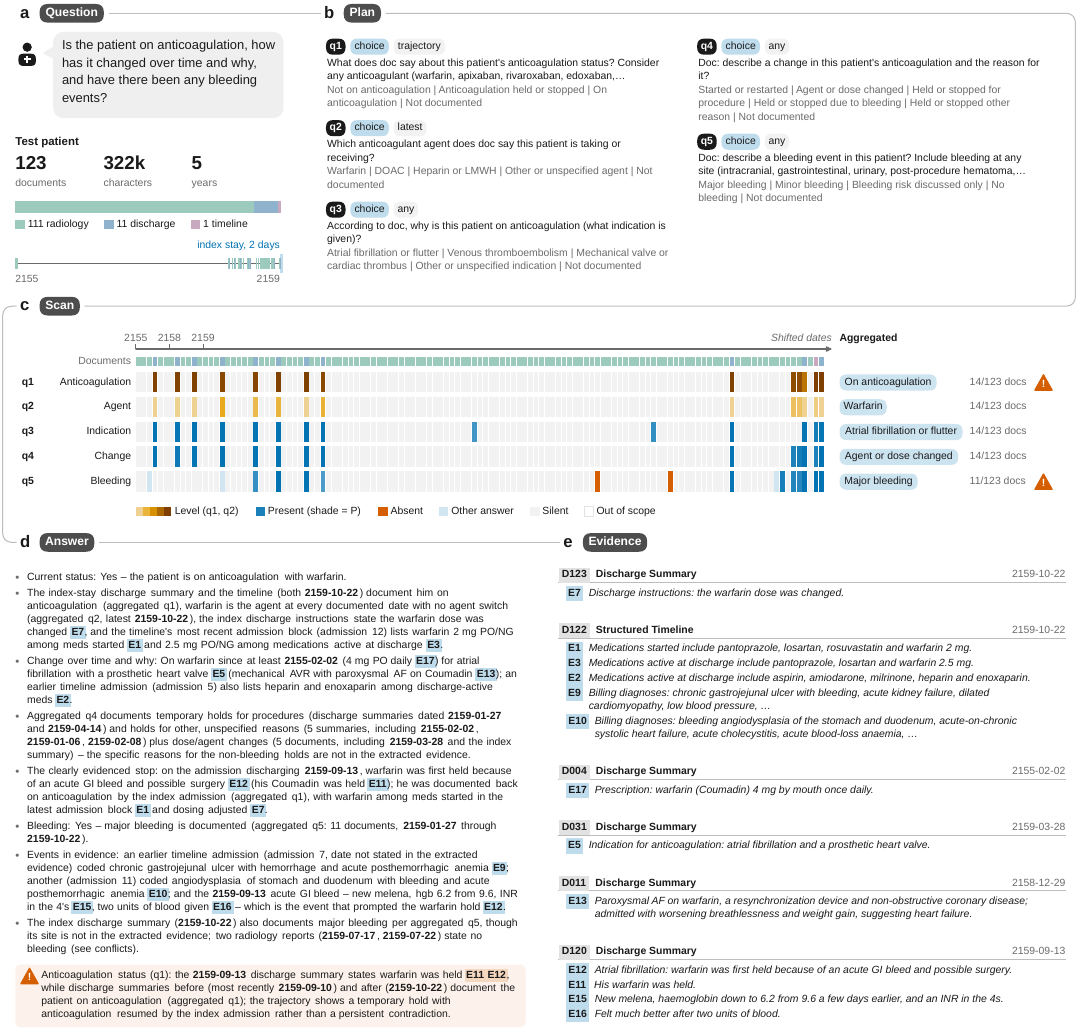}
\caption{\textbf{Patient example 2: anticoagulation and bleeding over a whole record.} As \cref{fig:patientqa}, for another MIMIC-IV test patient and question. \textbf{a}, A clinical question and the summarized record of a MIMIC-IV test patient. \textbf{b}, An LLM (GLM-5.3) plans 5 typed sub-questions, each with an aggregation rule and a scope. \textbf{c}, \mainmodel{} answers every sub-question for every document (620 typed decisions; long documents chunked, nothing truncated); rows are labelled by topic. Level rows (trajectory and latest rules) show each document's level, light to dark in option order; the other rows show whether a document reports the finding (blue, shaded by probability), reports it absent (orange), gives another answer or is silent. Right, code aggregates each sub-question across documents and time: the aggregated answer and the documents that support it; a warning triangle marks conflicting documents, explained in d. \textbf{d}, The LLM composes a cited answer from the aggregated answers and the evidence sentences (one bullet per sentence). \textbf{e}, \system{} attribution selects 18 evidence sentences; the 13 cited in d are shown, grouped by document (a sentence cited twice shares a line). The LLM plans from a manifest of the record (document types, counts and dates) and never reads the record; at the last step it sees only the aggregated answers and the evidence sentences. Aggregation rules: any, the highest probability over documents; latest, the most recent document that addresses the sub-question; trajectory, the per-document level over time; a conflict is confident support and confident contradiction in different documents. Sub-questions scoped to this admission are asked only of its documents (white cells, out of scope). Evidence sentences are selected by \system{} sufficiency attribution; in compliance with the MIMIC-IV data-use policy, the evidence sentences and any record text quoted in the answer are shown as paraphrases, not verbatim. Dates are shifted by MIMIC-IV de-identification and documents are relabelled D001\ldots in date order. Results shown are from a single run, with GLM-5.3 at its default reasoning effort and \mainmodel{} (raw outputs).}
\label{edfig:patient2}
\end{figure}

\begin{figure}[H]
\centering
\figinclude[width=\linewidth]{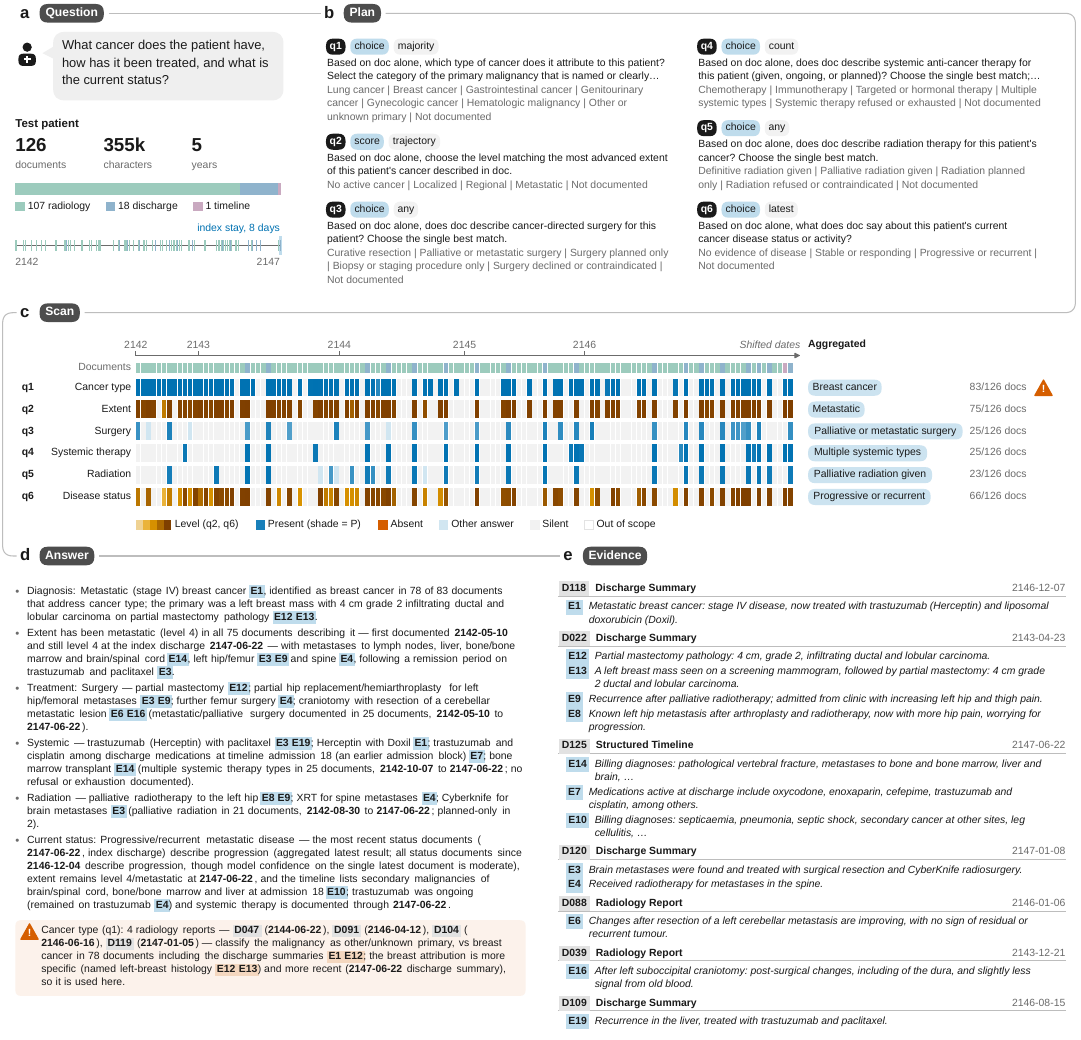}
\caption{\textbf{Patient example 3: cancer course over a whole record.} As \cref{fig:patientqa}, for another MIMIC-IV test patient and question. \textbf{a}, A clinical question and the summarized record of a MIMIC-IV test patient. \textbf{b}, An LLM (GLM-5.3) plans 6 typed sub-questions, each with an aggregation rule and a scope. \textbf{c}, \mainmodel{} answers every sub-question for every document (756 typed decisions; long documents chunked, nothing truncated); rows are labelled by topic. Level rows (trajectory and latest rules) show each document's level, light to dark in option order; the other rows show whether a document reports the finding (blue, shaded by probability), reports it absent (orange), gives another answer or is silent. Right, code aggregates each sub-question across documents and time: the aggregated answer and the documents that support it; a warning triangle marks conflicting documents, explained in d. \textbf{d}, The LLM composes a cited answer from the aggregated answers and the evidence sentences (one bullet per sentence). \textbf{e}, \system{} attribution selects 19 evidence sentences; the 13 cited in d are shown, grouped by document (a sentence cited twice shares a line). The LLM plans from a manifest of the record (document types, counts and dates) and never reads the record; at the last step it sees only the aggregated answers and the evidence sentences. Aggregation rules: any, the highest probability over documents; latest, the most recent document that addresses the sub-question; trajectory, the per-document level over time; a conflict is confident support and confident contradiction in different documents. Sub-questions scoped to this admission are asked only of its documents (white cells, out of scope). Evidence sentences are selected by \system{} sufficiency attribution; in compliance with the MIMIC-IV data-use policy, the evidence sentences and any record text quoted in the answer are shown as paraphrases, not verbatim. Dates are shifted by MIMIC-IV de-identification and documents are relabelled D001\ldots in date order. Results shown are from a single run, with GLM-5.3 at its default reasoning effort and \mainmodel{} (raw outputs).}
\label{edfig:patient3}
\end{figure}

\begin{table}[p]
\centering\footnotesize
\caption{\textbf{Fine-tuning data of \mainmodel{}.} \dataset{} only, in four groups; no MIMIC, other public or synthetic data. Splits are by patient; the test split, \testset{}, is used only for the internal evaluation (Extended Data Table~3). Pairs are document--question training pairs; a patient can contribute to several groups, so the total counts patients once.}
\label{edtab:data}
\setlength{\tabcolsep}{4pt}
\begin{tabularx}{\linewidth}{@{}>{\raggedright}X>{\raggedright}p{3.0cm}rrrr@{}}
\toprule
\dataset{} source & Labels & Documents & Unique questions & Patients & Pairs \\
\midrule
Clinical notes & teacher readout & 3,765,728 & 480,238 & 690,136 & 4,511,638 \\
Diagnostic reports & teacher readout & 687,809 & 23,063 & 458,666 & 2,460,078 \\
Structured records: outcome forecasting & observed outcomes & 431,890 & 6,013 & 431,862 & 433,430 \\
Structured records: structured criteria & exact (rules) & 2,581,620 & 497,312 & 2,425,282 & 5,011,302 \\
\midrule
Total & & 7,467,047 & 1,006,626 & 3,643,798 & 12,416,448 \\
\bottomrule
\end{tabularx}
\end{table}

\begin{table}[p]
\centering\footnotesize
\caption{\textbf{Speed, compute and calibration.} \textbf{a}, Latency per document on one NVIDIA GB10 (bf16, synthetic 2,900-token note, one document at a time) for 1, 10 and 40 questions; answers per second at 40 questions; and GB10-hours to answer 40 questions for each of $10^6$ documents, extrapolated. \mainmodel{} and the six Decision 2.0 models were timed back to back on one idle GB10 with no other GPU process, each through its released inference code (Decision 2.0: one \texttt{system\_one} call per document, as on its model cards, not further optimized; Methods); median of 10 timed repetitions after warm-up; interquartile ranges are within \CSoneTimingIQRMax\% of the median. Ratio, latency at 40 questions relative to \mainmodel{}. Latency against document length, Extended Data Fig.~3. \textbf{b}, Top-label expected calibration error (15 equal-count bins; raw model outputs, except the LLM teacher, temperature-scaled as labelled) on the public data sets; pooled, the five human-keyed sets (14,756 items, identical for every system). None of the open models ships a calibration file. \textbf{c}, Coverage at 99\% precision: share of judgements a system can decide, taken in order of confidence, while keeping 99\% of them correct (confidence, the top option's probability with the option probabilities renormalized to sum to 1, as in the calibration errors of Fig.~2h; ties, confidences equal to six decimals, counted together). Same items for every system within a column (Extended Data Table~4); the teacher was not run on the structured-record sets.}
\label{edtab:speed}
\setlength{\tabcolsep}{3pt}
\textbf{a}\\[2pt]
\begin{tabular}{@{}lrrrrrr@{}}
\toprule
System & ms/doc (1 q) & ms/doc (10 q) & ms/doc (40 q) & Answers/s (40 q) & GB10-h per $10^6$ docs & Ratio to \model{4B} \\
\midrule
\model{4B} & \CSoneMsOne & \CSoneMsTen & \CSoneMsForty & \CSoneAnswersPerSec & \CSoneGBHours & 1.0 \\
Decision 2.0 Eos-0.8B & 241 & 377 & 811 & 49.3 & 225 & 0.5 \\
Decision 2.0 Kai-0.6B & 186 & 408 & 818 & 48.9 & 227 & 0.5 \\
Decision 2.0 Sol-2B & 357 & 535 & 1,059 & 37.8 & 294 & 0.6 \\
Decision 2.0 Nox-4B & 797 & 1,136 & 2,283 & 17.5 & 634 & 1.3 \\
Decision 2.0 Lux-9B & 1,159 & 1,586 & 2,978 & 13.4 & 827 & 1.7 \\
Decision 2.0 Vega-27B & 4,559 & 6,042 & 11,540 & 3.5 & 3,206 & 6.4 \\
\bottomrule
\end{tabular}

\vspace{6pt}
\textbf{b}\quad ECE (raw)\\[2pt]
\resizebox{\linewidth}{!}{%
\begin{tabular}{@{}lrrrrrrrrrrrrr@{}}
\toprule
System & \hd{MIMIC-}{Timelines} & \hd{eICU-}{Timelines} & \hd{EHRSHOT-}{Timelines} & \hd{MIMIC-}{Composite} & \hd{eICU-}{Composite} & \hd{EHRSHOT-}{Composite} & \hd{Apixaban}{} & \hd{EHRNoteQA}{} & \hd{LongHealth}{} & \hd{Phenotypes}{} & \hd{CXR-VQA}{} & \hd{ReXVQA}{} & \hd{Pooled}{} \\
\midrule
LLM teacher (temperature-scaled) & -- & -- & -- & -- & -- & -- & .019 & .020 & .045 & .019 & .032 & .002 & .013 \\
JEV-27B & .022 & .038 & .023 & .012 & .008 & .052 & .015 & .014 & .028 & .008 & .067 & .005 & .010 \\
Clef & .043 & .047 & .047 & .036 & .054 & .029 & .036 & .024 & .057 & .030 & .066 & .026 & .024 \\
Decision 2.0 Kai-0.6B & .057 & .234 & .104 & .281 & .163 & .031 & .100 & .266 & .164 & .152 & .090 & .111 & .133 \\
Decision 2.0 Eos-0.8B & .131 & .057 & .097 & .078 & .087 & .066 & .036 & .071 & .063 & .024 & .104 & .021 & .019 \\
Decision 2.0 Sol-2B & .088 & .057 & .078 & .077 & .116 & .062 & .103 & .219 & .107 & .054 & .039 & .039 & .062 \\
Decision 2.0 Nox-4B & .046 & .032 & .029 & .091 & .080 & .057 & .076 & .117 & .118 & .055 & .038 & .026 & .058 \\
Decision 2.0 Lux-9B & .034 & .044 & .042 & .097 & .119 & .055 & .035 & .050 & .068 & .024 & .020 & .006 & .024 \\
Decision 2.0 Vega-27B & .068 & .124 & .077 & .101 & .150 & .104 & .076 & .166 & .119 & .067 & .053 & .075 & .073 \\
\model{4B} (raw) & .026 & .007 & .038 & .022 & .019 & .100 & .029 & .019 & .046 & .041 & .066 & .022 & .029 \\
\bottomrule
\end{tabular}}

\vspace{6pt}
\textbf{c}\quad Coverage at 99\% precision\\[2pt]
\resizebox{\linewidth}{!}{%
\begin{tabular}{@{}lrrrrrrrrrrrr@{}}
\toprule
System & \hd{MIMIC-}{Timelines} & \hd{eICU-}{Timelines} & \hd{EHRSHOT-}{Timelines} & \hd{MIMIC-}{Composite} & \hd{eICU-}{Composite} & \hd{EHRSHOT-}{Composite} & \hd{Apixaban}{} & \hd{EHRNoteQA}{} & \hd{LongHealth}{} & \hd{Phenotypes}{} & \hd{CXR-VQA}{} & \hd{ReXVQA}{} \\
\midrule
LLM teacher & -- & -- & -- & -- & -- & -- & 0.01 & 0.86 & 0.00 & 0.01 & 0.25 & 0.97 \\
JEV-27B & 0.76 & 0.45 & 0.46 & 0.84 & 0.76 & 0.00 & 0.09 & 0.84 & 0.00 & 0.60 & 0.38 & 0.97 \\
Clef & 0.65 & 0.73 & 0.26 & 0.80 & 0.79 & 0.02 & 0.34 & 0.85 & 0.09 & 0.78 & 0.00 & 0.97 \\
Decision 2.0 Kai-0.6B & 0.02 & 0.00 & 0.02 & 0.04 & 0.04 & 0.00 & 0.40 & 0.19 & 0.07 & 0.57 & 0.19 & 0.87 \\
Decision 2.0 Eos-0.8B & 0.05 & 0.10 & 0.03 & 0.14 & 0.00 & 0.00 & 0.29 & 0.24 & 0.17 & 0.10 & 0.02 & 0.87 \\
Decision 2.0 Sol-2B & 0.01 & 0.35 & 0.03 & 0.29 & 0.14 & 0.03 & 0.60 & 0.53 & 0.25 & 0.55 & 0.39 & 0.95 \\
Decision 2.0 Nox-4B & 0.31 & 0.26 & 0.23 & 0.67 & 0.55 & 0.03 & 0.76 & 0.68 & 0.07 & 0.52 & 0.55 & 0.97 \\
Decision 2.0 Lux-9B & 0.59 & 0.33 & 0.43 & 0.70 & 0.50 & 0.02 & 0.76 & 0.77 & 0.07 & 0.47 & 0.52 & 0.97 \\
Decision 2.0 Vega-27B & 0.76 & 0.03 & 0.44 & 0.87 & 0.68 & 0.00 & 0.59 & 0.79 & 0.00 & 0.08 & 0.63 & 0.98 \\
\model{4B} (raw) & 0.66 & 0.84 & 0.27 & 0.73 & 0.77 & 0.09 & 0.00 & 0.82 & 0.10 & 0.76 & 0.46 & 0.96 \\
\bottomrule
\end{tabular}}

\vspace{2pt}
{\scriptsize --, not run (the teacher: not run on the structured-record sets).}
\end{table}

\begin{table}[p]
\centering\scriptsize
\gdef\IHtasks{28}
\gdef\IHpairs{963,408}
\caption[Internal test on held-out \institution{} patients, per task]{\textbf{Internal test on held-out \institution{} patients, per task.} Balanced accuracy (\%; mean recall over the key's answer classes per question, averaged over questions) of every system on the \IHtasks{} tasks of the conformed \testset{} test set (\IHpairs{} held-out patients, one test question each), grouped by task type (Fig.~2a) and kind of key; the means of the rule-, outcome- and LLM-keyed rows are plotted in Fig.~2b. Key: rule, computed by a rule from the record or structured tables; outcome, observed after the decision point; LLM, written by an open LLM (Gemma-4-26B-A4B\cite{gemmateam2026gemma4,google2026gemma4card}); teacher (task in italics), the GLM-5.3-Flash teacher's answer, so these tasks measure agreement with the teacher and are secondary. Raw outputs; per-type temperature scaling leaves the top answer, and so balanced accuracy, unchanged on every task. Most common, the most common training answer per question (scored on pairs whose question occurs in training). Raw accuracy is dominated by the most common answer on several note tasks (for example 98.6\% on addenda and 86.3\% on coding review), so balanced accuracy is the primary metric. Bold, best system per task.}
\label{edtab:internal}
\setlength{\tabcolsep}{2.6pt}
\resizebox{\linewidth}{!}{%
\begin{tabular}{@{}llrrrrrrrrrrrrr@{}}
\toprule
& & & \multicolumn{1}{c}{Internal} & \multicolumn{10}{c}{Open decision models} & \\
\cmidrule(lr){4-4}\cmidrule(lr){5-14}
Task & Key & Pairs & \hd{CS1-}{4B} & \hd{JEV-}{27B} & \hd{Clef-}{27B} & \hd{pplx-}{27B} & \hd{Vega-}{27B} & \hd{Lux-}{9B} & \hd{Nox-}{4B} & \hd{Sol-}{2B} & \hd{Eos-}{0.8B} & \hd{Kai-}{0.6B} & \hd{BioDec-}{4B} & \hd{Most}{common} \\
\midrule
\multicolumn{15}{@{}l}{\textbf{1 Patient timelines (rule keys)}} \\
\quad Care process & rule & 47,091 & \textbf{95.4} & 72.1 & 66.2 & 74.5 & 64.2 & 58.7 & 49.2 & 47.8 & 43.9 & 38.6 & 31.2 & 46.6 \\
\quad Conditions & rule & 97,491 & \textbf{98.4} & 93.1 & 90.9 & 92.2 & 94.3 & 89.0 & 85.3 & 74.2 & 69.1 & 60.1 & 38.7 & 64.1 \\
\quad Demographics & rule & 31,782 & \textbf{95.8} & 90.0 & 89.7 & 90.3 & 88.7 & 85.6 & 83.9 & 76.7 & 81.4 & 75.7 & 48.3 & 46.4 \\
\quad Measurements & rule & 102,182 & \textbf{98.7} & 90.1 & 89.2 & 90.9 & 90.5 & 81.9 & 79.5 & 60.8 & 59.1 & 51.0 & 29.7 & 49.9 \\
\quad Medications & rule & 106,638 & \textbf{92.9} & 76.8 & 72.5 & 72.9 & 80.1 & 65.3 & 58.8 & 41.1 & 43.4 & 41.5 & 43.6 & 38.5 \\
\quad Miscellaneous & rule & 67,137 & \textbf{94.7} & 91.4 & 88.6 & 91.1 & 92.5 & 83.3 & 80.7 & 69.0 & 64.4 & 49.4 & 35.2 & 57.5 \\
\quad Timing & rule & 44,906 & \textbf{98.8} & 87.8 & 85.9 & 85.8 & 88.6 & 78.4 & 76.9 & 55.7 & 58.0 & 47.2 & 30.5 & 45.0 \\
\quad Utilization & rule & 92,362 & \textbf{97.7} & 93.0 & 91.2 & 94.2 & 93.7 & 91.7 & 92.4 & 89.4 & 89.6 & 86.6 & 75.1 & 66.3 \\
\midrule
\multicolumn{15}{@{}l}{\textbf{2 Clinical definitions (rule keys)}} \\
\quad BP, BMI, FIB-4, eGFR definitions & rule & 62,045 & \textbf{92.1} & 82.2 & 79.7 & 78.9 & 85.0 & 73.3 & 63.9 & 39.8 & 43.2 & 27.4 & 35.6 & 18.3 \\
\midrule
\multicolumn{15}{@{}l}{\textbf{3 Outcome forecasting (observed outcomes)}} \\
\quad Survival and onset forecasts & outcome & 51,426 & \textbf{75.6} & 69.4 & 68.7 & 63.5 & 67.9 & 64.7 & 64.6 & 54.2 & 54.1 & 51.8 & 51.4 & 50.8 \\
\midrule
\multicolumn{15}{@{}l}{\textbf{4 Phenotypes and assertions: medication status (rule keys)}} \\
\quad Medication status & rule & 8,865 & \textbf{95.0} & 79.7 & 79.8 & 71.8 & 79.4 & 76.8 & 74.0 & 69.2 & 68.9 & 74.5 & 82.6 & 41.6 \\
\midrule
\multicolumn{15}{@{}l}{\textbf{5 Questions over notes, rule- or LLM-keyed}} \\
\quad Coding review & rule & 14,813 & \textbf{89.3} & 67.9 & 73.5 & 66.3 & 76.7 & 64.5 & 61.9 & 27.5 & 37.7 & 34.6 & 8.9 & 73.1 \\
\quad BRIE-style chart review & LLM & 499 & \textbf{94.7} & 93.6 & 92.9 & 82.2 & 93.6 & 94.5 & 91.7 & 86.4 & 74.2 & 67.5 & 78.1 & 65.7 \\
\midrule
\multicolumn{15}{@{}l}{\textbf{5 Questions over notes, teacher-keyed (secondary)}} \\
\quad \textit{Addenda} & teacher & 14,403 & \textbf{86.0} & 84.6 & 78.9 & 84.3 & 61.1 & 77.6 & 82.9 & 81.8 & 81.6 & 61.9 & 17.2 & 82.8 \\
\quad \textit{Care transitions} & teacher & 9,121 & \textbf{84.1} & 78.3 & 74.5 & 79.8 & 75.9 & 74.8 & 74.8 & 61.0 & 53.8 & 54.0 & 54.7 & 41.1 \\
\quad \textit{Cross-cutting clinical} & teacher & 21,826 & \textbf{91.3} & 89.5 & 88.6 & 90.8 & 90.4 & 88.7 & 87.6 & 80.0 & 72.5 & 67.7 & 61.8 & 53.5 \\
\quad \textit{Emergency department} & teacher & 21,316 & \textbf{87.9} & 85.9 & 84.6 & 86.5 & 85.0 & 83.5 & 82.9 & 74.3 & 68.1 & 64.1 & 63.6 & 49.4 \\
\quad \textit{H\&P and consults} & teacher & 11,065 & \textbf{85.4} & 83.9 & 82.3 & 84.2 & 82.5 & 80.1 & 79.2 & 68.6 & 61.7 & 57.9 & 61.7 & 36.5 \\
\quad \textit{Perioperative} & teacher & 18,991 & \textbf{86.8} & 83.4 & 80.7 & 83.8 & 82.4 & 82.1 & 81.4 & 73.5 & 68.6 & 63.2 & 63.0 & 54.1 \\
\quad \textit{Plan of care} & teacher & 12,455 & \textbf{83.5} & 81.8 & 78.2 & 81.5 & 75.6 & 78.6 & 76.8 & 66.6 & 59.5 & 57.3 & 50.4 & 51.1 \\
\quad \textit{Progress notes} & teacher & 19,084 & \textbf{83.3} & 80.1 & 77.2 & 80.7 & 78.6 & 77.7 & 77.5 & 66.8 & 59.6 & 57.6 & 53.8 & 46.2 \\
\quad \textit{Specialty notes} & teacher & 7,274 & \textbf{83.8} & 73.7 & 70.8 & 73.6 & 75.2 & 76.0 & 72.2 & 65.0 & 53.7 & 56.9 & 55.0 & 49.8 \\
\midrule
\multicolumn{15}{@{}l}{\textbf{6 Diagnostic reports, rule-keyed}} \\
\quad Cancer staging & rule & 3,242 & 86.6 & 85.7 & 86.2 & \textbf{86.9} & 86.2 & 85.7 & 84.7 & 69.3 & 65.8 & 54.8 & 71.3 & 49.3 \\
\midrule
\multicolumn{15}{@{}l}{\textbf{6 Diagnostic reports, teacher-keyed (secondary)}} \\
\quad \textit{Imaging} & teacher & 41,116 & \textbf{91.6} & 90.1 & 88.5 & 90.4 & 88.6 & 87.2 & 85.3 & 76.6 & 68.3 & 63.3 & 54.6 & 30.1 \\
\quad \textit{Molecular, cytogenetics} & teacher & 3,891 & \textbf{91.2} & 89.1 & 86.7 & 90.3 & 87.1 & 86.9 & 83.6 & 77.2 & 61.8 & 58.8 & 57.7 & 30.1 \\
\quad \textit{Ophthalmology} & teacher & 12,166 & \textbf{91.5} & 88.7 & 89.3 & 89.1 & 87.4 & 87.3 & 84.2 & 70.4 & 69.1 & 65.2 & 57.0 & 40.6 \\
\quad \textit{Pathology, cytology} & teacher & 38,967 & \textbf{91.7} & 89.1 & 88.9 & 88.8 & 89.0 & 86.6 & 85.0 & 73.1 & 69.2 & 62.9 & 63.6 & 32.5 \\
\quad \textit{Specialty reports} & teacher & 1,254 & 82.4 & 82.5 & 85.1 & 85.0 & 84.6 & \textbf{87.1} & 77.4 & 67.1 & 67.5 & 66.0 & 56.9 & 50.3 \\
\bottomrule
\end{tabular}}
\end{table}

\begin{table}[p]
\centering\scriptsize
\gdef\EXTsets{21}
\gdef\EXTpairs{86,830}
\gdef\EXTbrieItems{5,815}
\gdef\EXTbrieLongest{72.9}
\gdef\EXTscoringNote{eICU-Timelines and the three Composite sets: balanced accuracy per concept, averaged over concepts; most common answer, the out-of-sample majority (not defined for EHRSHOT-Composite). Prognosis sets: mean AUROC over the outcomes with at least 10 positives and 10 negatives (MIMIC 7, eICU 5, EHRSHOT 15); most common answer, a constant forecast (0.500).}
\gdef\EXTmeddeciderNote{MedDecider-9B returned non-finite probabilities on eICU-Timelines, MIMIC-Composite, eICU-Composite, EHRSHOT-Composite and EHRSHOT-Prognosis (--).}
\gdef\EXTteacherTCGA{1,516}
\gdef\EXTtcgaPairs{2,139}
\gdef\EXTprognosisCaption{\textbf{Outcome forecasting on the three prognosis sets.} AUROC per outcome (raw probabilities; risk = P(yes), or the probability of the event option for EHRSHOT question kinds with a clear event option) for the outcomes with at least 10 positives and 10 negatives (MIMIC-Prognosis, 7 of 12 outcomes, eICU-Prognosis, 5 of 6 outcomes and EHRSHOT-Prognosis, 15 of 20 outcomes (19 with any positive); $n$ / positives per outcome in parentheses), and their mean, the value of Fig.~2c and Extended Data Table~4. Mean ECE, over all outcomes with positives; Brier $>$ prev., outcomes with positives whose Brier score is worse than always predicting the prevalence. One question per patient. The LLM teacher was not run on these sets.}
\captionsetup{singlelinecheck=false,font=scriptsize}%
\caption{\textbf{Public data sets: balanced accuracy.} Balanced accuracy (\%) of every system on the \EXTsets{} public data sets (\EXTpairs{} scored pairs; the structured-record sets of MIMIC, EHRSHOT and eICU are in Fig.~2c), grouped by task type, with the source of the key (rule, outcome, human, LLM, registry or teacher), documents and scored pairs (MIMIC- and EHRSHOT-Timelines, a seeded subset; Methods). Most common answer: the key's per-question majority or prior, which marks chance for this metric. \EXTscoringNote{} $^\S$, built for this study from public records (13 sets); unmarked, published benchmarks converted without changing their keys (8 sets; Methods). BRIE$^\ddagger$: raw accuracy (model-written key); the longest option is correct for \EXTbrieLongest\% of its \EXTbrieItems{} items. TCGA: $^*$, the teacher on the \EXTteacherTCGA{} of \EXTtcgaPairs{} pairs it had labelled. HISTAI$^\dagger$: keyed by the teacher's answer, so the teacher is not scored (\textit{key}). The hosted TypeSafe API was run only on openly licensed sets. \mainmodel{}: fine-tuned on \dataset{} data only; raw probabilities. Bold, best measured system per set; --, not run or not applicable. \EXTmeddeciderNote{} Parameters, loaded parameters. 95\% intervals are in the source data.}
\label{edtab:external}
Balanced accuracy (\%; Prognosis sets, mean AUROC)\\
\setlength{\tabcolsep}{2.0pt}\renewcommand{\arraystretch}{0.78}
\resizebox{\linewidth}{!}{%
\begin{tabular}{@{}llrrrrrrrrrrrrrrrrrr@{}}
\toprule
 &  &  &  &  & \multicolumn{1}{c}{Teacher} & \multicolumn{4}{c}{Open 27B models, hosted API} & \multicolumn{6}{c}{Decision 2.0 (open)} & \multicolumn{3}{c}{Biomedical (open)} & \multicolumn{1}{c}{CS1} \\
\cmidrule(lr){6-6}\cmidrule(lr){7-10}\cmidrule(lr){11-16}\cmidrule(lr){17-19}\cmidrule(lr){20-20}
Set & Key & Docs & Pairs & \hd{Most}{common} & \hd{LLM}{teacher} & \hd{JEV-}{27B} & \hd{Clef-}{27B} & \hd{pplx-}{27B} & \hd{TypeSafe}{API} & \hd{Kai-}{0.6B} & \hd{Eos-}{0.8B} & \hd{Sol-}{2B} & \hd{Nox-}{4B} & \hd{Lux-}{9B} & \hd{Vega-}{27B} & \hd{BioDec-}{4B} & \hd{MedDec-}{9B} & \hd{MedDec-}{27B} & \hd{CS1-}{4B} \\
\midrule
\multicolumn{20}{@{}l}{\textbf{1 Patient timelines}} \\
\quad MIMIC-Timelines$^\S$ & rule & 45 & 1,800 & 38.5 & -- & 92.4 & \textbf{93.4} & 91.3 & -- & 58.4 & 61.3 & 66.9 & 85.9 & 88.0 & 93.0 & 86.0 & 86.2 & 91.0 & 92.7 \\
\quad eICU-Timelines$^\S$ & rule & 2,400 & 2,400 & 62.1 & -- & 91.1 & 89.6 & 91.0 & -- & 60.3 & 67.1 & 67.1 & 74.8 & 80.3 & 89.8 & 81.5 & -- & 90.6 & \textbf{92.8} \\
\quad EHRSHOT-Timelines$^\S$ & rule & 35 & 1,400 & 39.2 & -- & 90.5 & 88.9 & 88.8 & -- & 49.1 & 58.2 & 64.7 & 76.9 & 80.7 & 89.6 & 76.5 & 78.1 & 86.4 & \textbf{94.3} \\
\quad MedAgentBench records$^\S$ & rule & 402 & 5,698 & 45.2 & -- & 99.3 & 97.1 & 98.1 & 99.4 & 59.8 & 70.1 & 71.3 & 87.8 & 91.4 & \textbf{99.6} & 89.2 & 88.8 & 98.5 & 95.9 \\
\midrule
\multicolumn{20}{@{}l}{\textbf{2 Clinical definitions}} \\
\quad MIMIC-Composite$^\S$ & rule & 2,388 & 2,388 & 31.6 & -- & 93.1 & 86.4 & 89.9 & -- & 45.8 & 50.3 & 53.7 & 81.9 & 87.8 & \textbf{95.1} & 79.8 & -- & 93.0 & 85.4 \\
\quad eICU-Composite$^\S$ & rule & 2,400 & 2,400 & 36.0 & -- & \textbf{89.6} & 82.4 & 84.4 & -- & 41.8 & 43.9 & 56.4 & 71.0 & 78.1 & 86.9 & 71.6 & -- & 86.4 & 82.2 \\
\quad EHRSHOT-Composite$^\S$ & rule & 2,400 & 2,400 & -- & -- & 87.2 & 82.9 & 83.7 & -- & 31.1 & 43.7 & 46.5 & 69.6 & 75.5 & \textbf{89.3} & 67.0 & -- & 82.9 & 76.5 \\
\midrule
\multicolumn{20}{@{}l}{\textbf{3 Outcome forecasting}} \\
\quad MIMIC-Prognosis (AUROC)$^\S$ & outcome & 1,507 & 1,507 & 0.500 & -- & 0.766 & 0.753 & 0.764 & -- & 0.591 & 0.518 & 0.583 & 0.702 & 0.748 & 0.769 & 0.743 & 0.680 & \textbf{0.771} & 0.721 \\
\quad eICU-Prognosis (AUROC)$^\S$ & outcome & 2,253 & 2,253 & 0.500 & -- & 0.741 & 0.729 & 0.743 & -- & 0.509 & 0.540 & 0.630 & 0.702 & 0.713 & 0.745 & 0.715 & 0.701 & \textbf{0.748} & 0.725 \\
\quad EHRSHOT-Prognosis (AUROC)$^\S$ & outcome & 1,637 & 1,637 & 0.500 & -- & 0.652 & 0.681 & 0.664 & -- & 0.559 & 0.565 & 0.623 & 0.595 & 0.602 & 0.672 & 0.613 & -- & 0.643 & \textbf{0.766} \\
\midrule
\multicolumn{20}{@{}l}{\textbf{4 Phenotypes and assertions}} \\
\quad Apixaban trial criteria & human & 100 & 1,900 & 46.1 & 85.5 & 86.7 & 85.3 & 87.8 & -- & 61.2 & 69.0 & 80.8 & 86.8 & 86.4 & \textbf{88.1} & 81.8 & 83.4 & 84.7 & 84.3 \\
\quad MIMIC-III phenotypes & human & 844 & 10,147 & 50.0 & 90.5 & 91.1 & \textbf{91.9} & 90.9 & -- & 72.6 & 79.0 & 86.7 & 90.2 & 90.4 & 91.5 & 84.7 & 81.7 & 88.0 & 87.6 \\
\quad MIPA phenotypes & human & 1,388 & 22,207 & 50.0 & -- & 95.4 & 95.7 & 95.7 & -- & 76.2 & 85.7 & 90.3 & 95.1 & 95.5 & 96.4 & 95.6 & 94.6 & 95.5 & \textbf{96.9} \\
\quad Assertions (van Aken) & human & 237 & 3,052 & 33.3 & -- & 83.3 & \textbf{85.8} & 82.5 & -- & 52.1 & 71.8 & 79.4 & 80.7 & 82.5 & 84.4 & 80.5 & 81.1 & 82.3 & 80.6 \\
\midrule
\multicolumn{20}{@{}l}{\textbf{5 Questions over notes}} \\
\quad EHRNoteQA & human & 962 & 962 & 20 & 95.0 & 95.7 & 95.1 & \textbf{96.4} & -- & 75.2 & 81.0 & 87.0 & 92.8 & 94.4 & 95.2 & 92.5 & 95.1 & 95.8 & 95.2 \\
\quad LongHealth & human & 20 & 400 & 20 & 89.5 & 89.0 & 90.2 & 81.3 & \textbf{91.9} & 46.0 & 57.0 & 66.4 & 83.1 & 85.4 & 86.6 & 76.9 & 88.6 & 91.3 & 88.1 \\
\quad BRIE (MIMIC)$^\ddagger$$^\S$ & LLM & 599 & 5,815 & 28.8 & -- & 94.1 & 94.6 & 88.7 & -- & 74.3 & 77.5 & 79.6 & 91.5 & 92.7 & 94.1 & 91.3 & 94.2 & 94.7 & \textbf{94.8} \\
\midrule
\multicolumn{20}{@{}l}{\textbf{6 Diagnostic reports}} \\
\quad MIMIC-CXR-VQA (Reports) & human & 485 & 1,347 & 50.0 & 87.8 & 86.9 & 86.8 & 88.8 & -- & 73.1 & 81.9 & 85.2 & 87.8 & 90.1 & \textbf{91.2} & 87.4 & 87.2 & 86.2 & 85.2 \\
\quad ReXVQA (Reports) & LLM & 4,058 & 5,000 & 25.0 & 97.4 & 97.8 & 97.3 & 97.7 & -- & 93.8 & 93.4 & 96.6 & 97.5 & 97.7 & 97.8 & 97.3 & \textbf{97.8} & 97.7 & 96.7 \\
\quad TCGA pathology$^\S$ & registry & 859 & 2,139 & 42.0 & 98.0$^*$ & 96.6 & \textbf{97.8} & 97.5 & 97.6 & 83.5 & 84.1 & 86.8 & 93.2 & 94.5 & 96.8 & 94.3 & 95.7 & 96.5 & 95.7 \\
\quad HISTAI pathology$^\dagger$$^\S$ & teacher & 1,000 & 9,978 & 36.6 & \textit{key} & \textbf{88.3} & 85.7 & 87.6 & -- & 66.4 & 70.2 & 77.6 & 83.7 & 86.2 & 87.4 & 81.2 & 85.5 & 87.6 & 86.8 \\
\midrule
Parameters & & & & -- & -- & 27B & 27B & 27B & n/a & 0.6B & 0.75B & 1.9B & 4.2B & 8B & 29B & 4.2B & 9B & 27B & 4.2B \\
\bottomrule
\end{tabular}}
\end{table}

\begin{table}[p]
\centering\scriptsize
\caption{\textbf{Serializer comparison on MIMIC-IV.} Balanced accuracy (\%) or mean AUROC of every system on the same documents rendered in the timeline layout (periods counted back from the index day; the layout of the \dataset{} training timelines and of the eICU-CRD and EHRSHOT sets) and in the encounter layout (admissions; the layout of the MIMIC-IV sets), scored on the identical pairs valid under both; encounter $-$ timeline, the paired difference with its 95\% interval. Timelines: 2,377 MIMIC-IV timeline questions built as the public timeline sets (one per document; 352 concepts), not one of the public data sets. Composite definitions and prognosis: the documents of MIMIC-Composite and MIMIC-Prognosis, whose encounter-layout values are those of Extended Data Table~4. On timelines every model with valid predictions, \mainmodel{} included, scored higher in the encounter layout.}
\label{edtab:serializer}
\setlength{\tabcolsep}{2.6pt}\renewcommand{\arraystretch}{0.85}
\resizebox{\linewidth}{!}{%
\begin{tabular}{@{}lrrrrrrrrr@{}}
\toprule
 & \multicolumn{3}{c}{\textbf{Timelines} (bal.\ acc., \%; 352 concepts; 2,377 pairs)} & \multicolumn{3}{c}{\textbf{Composite definitions} (bal.\ acc., \%; 22 concepts; 2,388 pairs)} & \multicolumn{3}{c}{\textbf{Prognosis} (mean AUROC, 7 outcomes; 2,398 pairs)} \\
\cmidrule(lr){2-4}\cmidrule(lr){5-7}\cmidrule(lr){8-10}
System & \hd{Timeline}{(days)} & \hd{Encounter}{(admissions)} & \hd{Encounter $-$ timeline}{{}[95\% CI]} & \hd{Timeline}{(days)} & \hd{Encounter}{(admissions)} & \hd{Encounter $-$ timeline}{{}[95\% CI]} & \hd{Timeline}{(days)} & \hd{Encounter}{(admissions)} & \hd{Encounter $-$ timeline}{{}[95\% CI]} \\
\midrule
Out-of-sample majority$^a$ / constant forecast & 76.3 & 76.3 & -- & 31.6 & 31.6 & -- & 0.500 & 0.500 & -- \\
\midrule
JEV-27B & 92.7 & 97.1 & +4.4 {\scriptsize[+2.8, +6.0]} & 90.1 & 93.1 & +3.0 {\scriptsize[$-$0.9, +6.9]} & 0.739 & 0.766 & +0.027 {\scriptsize[+0.002, +0.057]} \\
Clef-27B & 93.6 & 97.1 & +3.5 {\scriptsize[+2.0, +5.0]} & 83.0 & 86.4 & +3.5 {\scriptsize[$-$0.6, +7.6]} & 0.748 & 0.753 & +0.005 {\scriptsize[$-$0.021, +0.031]} \\
pplx-decider-27B & 93.6 & 97.8 & +4.1 {\scriptsize[+2.5, +5.7]} & 84.6 & 89.9 & +5.3 {\scriptsize[$-$1.8, +12.5]} & 0.747 & 0.764 & +0.017 {\scriptsize[$-$0.008, +0.044]} \\
Kai-0.6B & 76.1 & 77.3 & +1.1 {\scriptsize[$-$0.8, +3.1]} & 42.6 & 45.8 & +3.2 {\scriptsize[$-$0.2, +6.5]} & 0.559 & 0.591 & +0.032 {\scriptsize[$-$0.024, +0.086]} \\
Eos-0.8B & 76.9 & 77.1 & +0.2 {\scriptsize[$-$2.0, +2.4]} & 57.2 & 50.3 & $-$6.9 {\scriptsize[$-$14.1, +0.3]} & 0.566 & 0.518 & $-$0.048 {\scriptsize[$-$0.107, +0.012]} \\
Sol-2B & 80.7 & 80.8 & +0.1 {\scriptsize[$-$1.7, +2.0]} & 55.3 & 53.7 & $-$1.7 {\scriptsize[$-$8.6, +5.3]} & 0.586 & 0.583 & $-$0.003 {\scriptsize[$-$0.057, +0.049]} \\
Nox-4B & 85.3 & 92.3 & +6.9 {\scriptsize[+4.9, +8.9]} & 80.9 & 81.9 & +0.9 {\scriptsize[$-$5.7, +7.5]} & 0.715 & 0.702 & $-$0.014 {\scriptsize[$-$0.041, +0.017]} \\
Lux-9B & 87.9 & 94.1 & +6.2 {\scriptsize[+4.1, +8.2]} & 81.4 & 87.8 & +6.4 {\scriptsize[$-$0.6, +13.5]} & 0.740 & 0.748 & +0.008 {\scriptsize[$-$0.019, +0.036]} \\
Vega-27B & 92.6 & 97.5 & +4.8 {\scriptsize[+3.2, +6.5]} & 91.0 & 95.1 & +4.1 {\scriptsize[$-$0.4, +8.7]} & 0.746 & 0.769 & +0.023 {\scriptsize[$-$0.004, +0.054]} \\
BioDecision-4B & 87.2 & 90.9 & +3.7 {\scriptsize[+1.7, +5.7]} & 78.0 & 79.8 & +1.8 {\scriptsize[$-$1.4, +5.0]} & 0.722 & 0.743 & +0.020 {\scriptsize[$-$0.014, +0.053]} \\
MedDecider-9B & -- & -- & -- & -- & -- & -- & 0.706 & 0.680 & $-$0.026 {\scriptsize[$-$0.064, +0.013]} \\
MedDecider-27B & 93.0 & 98.1 & +5.0 {\scriptsize[+3.3, +6.7]} & 89.8 & 93.0 & +3.2 {\scriptsize[$-$0.2, +6.7]} & 0.749 & 0.771 & +0.022 {\scriptsize[$-$0.003, +0.049]} \\
CS1-4B & 87.5 & 93.3 & +5.8 {\scriptsize[+4.4, +7.3]} & 83.1 & 85.4 & +2.3 {\scriptsize[$-$0.4, +5.0]} & 0.697 & 0.721 & +0.024 {\scriptsize[$-$0.002, +0.047]} \\
\bottomrule
\end{tabular}}

\vspace{2pt}{\scriptsize CIs, 1,000 resamples. Balanced accuracy: Bayesian bootstrap over concepts (Dirichlet(1) concept weights; the difference on the same weights), small-sample $t$ correction. Mean AUROC (outcomes with $\geq$10 positives and $\geq$10 negatives): 95\% percentile of Dirichlet(1) patient weights. $^a$ Per concept the most common key among unused test pairs of other patients (the same for both layouts); prognosis, a constant forecast (0.500). --, MedDecider-9B: invalid (non-finite) probabilities.}
\end{table}

\begin{table}[p]
\centering\footnotesize
\caption{\EXTprognosisCaption}
\label{edtab:prognosis}
\setlength{\tabcolsep}{2.4pt}
\resizebox{\linewidth}{!}{%
\begin{tabular}{@{}lrrrrrrrrrrrrr@{}}
\toprule
 &  &  &  & \multicolumn{6}{c}{Decision 2.0} &  &  &  &  \\
\cmidrule(lr){5-10}
Outcome ($n$ / positives) & \hd{JEV-}{27B} & \hd{Clef-}{27B} & \hd{pplx-}{27B} & \hd{Kai-}{0.6B} & \hd{Eos-}{0.8B} & \hd{Sol-}{2B} & \hd{Nox-}{4B} & \hd{Lux-}{9B} & \hd{Vega-}{27B} & \hd{BioDec-}{4B} & \hd{MedDec-}{9B} & \hd{MedDec-}{27B} & \hd{CS1-}{4B} \\
\midrule
\multicolumn{14}{@{}l}{\textbf{MIMIC-Prognosis} (MIMIC-IV, TD2 encounter layout; 2,398 patients, one question each)} \\
\quad AKI within 48\,h (224/16) & 0.670 & 0.633 & 0.701 & 0.664 & 0.325 & 0.582 & 0.697 & 0.669 & 0.750 & 0.709 & 0.639 & 0.674 & 0.690 \\
\quad 1-year death (165/17) & 0.913 & 0.909 & 0.908 & 0.631 & 0.604 & 0.742 & 0.860 & 0.879 & 0.907 & 0.896 & 0.885 & 0.909 & 0.890 \\
\quad 30-day emergency revisit (178/30) & 0.613 & 0.592 & 0.636 & 0.552 & 0.579 & 0.616 & 0.646 & 0.668 & 0.627 & 0.557 & 0.553 & 0.585 & 0.624 \\
\quad Intensive care after 24\,h (174/20) & 0.797 & 0.811 & 0.755 & 0.526 & 0.375 & 0.440 & 0.571 & 0.808 & 0.771 & 0.762 & 0.611 & 0.868 & 0.643 \\
\quad In-hospital death (296/16) & 0.928 & 0.916 & 0.922 & 0.747 & 0.711 & 0.760 & 0.895 & 0.911 & 0.934 & 0.894 & 0.869 & 0.914 & 0.883 \\
\quad Length of stay $>$7 days (299/86) & 0.696 & 0.681 & 0.685 & 0.533 & 0.488 & 0.406 & 0.565 & 0.638 & 0.686 & 0.681 & 0.529 & 0.711 & 0.603 \\
\quad 30-day readmission (171/28) & 0.748 & 0.726 & 0.738 & 0.486 & 0.541 & 0.536 & 0.679 & 0.664 & 0.706 & 0.699 & 0.671 & 0.735 & 0.716 \\
\quad Mean AUROC & 0.766 & 0.753 & 0.764 & 0.591 & 0.518 & 0.583 & 0.702 & 0.748 & 0.769 & 0.743 & 0.680 & 0.771 & 0.721 \\
\quad Mean ECE (all outcomes) & 0.264 & 0.190 & 0.196 & 0.244 & 0.201 & 0.216 & 0.246 & 0.210 & 0.261 & 0.160 & 0.201 & 0.173 & 0.285 \\
\quad Brier $>$ prev.\ (of 12) & 12 & 11 & 12 & 12 & 12 & 12 & 11 & 10 & 11 & 10 & 12 & 10 & 12 \\
\midrule
\multicolumn{14}{@{}l}{\textbf{eICU-Prognosis} (eICU-CRD, TD1 timeline layout; 2,400 patients, one question each)} \\
\quad AKI within 48\,h (254/26) & 0.671 & 0.629 & 0.659 & 0.480 & 0.567 & 0.641 & 0.648 & 0.622 & 0.654 & 0.699 & 0.638 & 0.666 & 0.684 \\
\quad ICU death (304/23) & 0.903 & 0.908 & 0.910 & 0.519 & 0.646 & 0.714 & 0.853 & 0.908 & 0.920 & 0.864 & 0.903 & 0.908 & 0.819 \\
\quad ICU readmission (1,113/70) & 0.567 & 0.590 & 0.593 & 0.460 & 0.448 & 0.440 & 0.574 & 0.538 & 0.578 & 0.575 & 0.529 & 0.591 & 0.600 \\
\quad In-hospital death (295/25) & 0.868 & 0.850 & 0.857 & 0.483 & 0.578 & 0.725 & 0.808 & 0.863 & 0.875 & 0.849 & 0.864 & 0.869 & 0.843 \\
\quad Length of stay $>$7 days (287/101) & 0.694 & 0.667 & 0.696 & 0.604 & 0.458 & 0.630 & 0.627 & 0.636 & 0.698 & 0.590 & 0.574 & 0.706 & 0.679 \\
\quad Mean AUROC & 0.741 & 0.729 & 0.743 & 0.509 & 0.540 & 0.630 & 0.702 & 0.713 & 0.745 & 0.715 & 0.701 & 0.748 & 0.725 \\
\quad Mean ECE (all outcomes) & 0.156 & 0.206 & 0.206 & 0.206 & 0.125 & 0.133 & 0.170 & 0.158 & 0.185 & 0.163 & 0.124 & 0.123 & 0.123 \\
\quad Brier $>$ prev.\ (of 6) & 6 & 6 & 6 & 6 & 5 & 5 & 4 & 5 & 6 & 6 & 6 & 6 & 5 \\
\midrule
\multicolumn{14}{@{}l}{\textbf{EHRSHOT-Prognosis} (EHRSHOT, TD1 timeline layout; 2,400 patients, one question each)} \\
\quad Admission within 30 days (110/13) & 0.544 & 0.665 & 0.665 & 0.726 & 0.723 & 0.647 & 0.670 & 0.560 & 0.694 & 0.623 & -- & 0.479 & 0.795 \\
\quad Admission within 1 year (110/34) & 0.505 & 0.571 & 0.569 & 0.581 & 0.435 & 0.610 & 0.641 & 0.654 & 0.559 & 0.556 & -- & 0.658 & 0.610 \\
\quad Admission within 90 days (109/12) & 0.803 & 0.834 & 0.855 & 0.733 & 0.527 & 0.631 & 0.731 & 0.818 & 0.822 & 0.758 & -- & 0.689 & 0.824 \\
\quad New diagnosis within 1 year (109/74) & 0.564 & 0.589 & 0.550 & 0.420 & 0.393 & 0.472 & 0.522 & 0.562 & 0.492 & 0.561 & -- & 0.612 & 0.663 \\
\quad Diagnosis recorded within 1 year (109/40) & 0.651 & 0.727 & 0.635 & 0.513 & 0.540 & 0.545 & 0.625 & 0.586 & 0.667 & 0.508 & -- & 0.640 & 0.755 \\
\quad Emergency visit within 30 days (109/13) & 0.687 & 0.649 & 0.651 & 0.638 & 0.720 & 0.565 & 0.501 & 0.355 & 0.738 & 0.395 & -- & 0.644 & 0.756 \\
\quad Emergency visit within 1 year (109/20) & 0.594 & 0.696 & 0.653 & 0.663 & 0.704 & 0.696 & 0.534 & 0.487 & 0.724 & 0.479 & -- & 0.670 & 0.815 \\
\quad Emergency visit within 90 days (109/11) & 0.710 & 0.658 & 0.663 & 0.474 & 0.535 & 0.528 & 0.543 & 0.586 & 0.688 & 0.521 & -- & 0.582 & 0.758 \\
\quad First admission within 30 days (109/42) & 0.698 & 0.758 & 0.666 & 0.371 & 0.596 & 0.682 & 0.480 & 0.668 & 0.698 & 0.668 & -- & 0.661 & 0.794 \\
\quad First admission within 1 year (109/86) & 0.612 & 0.568 & 0.692 & 0.531 & 0.524 & 0.584 & 0.655 & 0.622 & 0.585 & 0.675 & -- & 0.586 & 0.725 \\
\quad First emergency visit within 30 days (109/23) & 0.669 & 0.763 & 0.650 & 0.483 & 0.525 & 0.708 & 0.370 & 0.532 & 0.732 & 0.739 & -- & 0.782 & 0.840 \\
\quad First emergency visit within 1 year (109/70) & 0.655 & 0.600 & 0.660 & 0.389 & 0.518 & 0.632 & 0.567 & 0.480 & 0.535 & 0.618 & -- & 0.606 & 0.720 \\
\quad HbA1c $\geq$7\% (109/18) & 0.962 & 0.947 & 0.960 & 0.685 & 0.744 & 0.806 & 0.964 & 0.964 & 0.963 & 0.919 & -- & 0.953 & 0.954 \\
\quad New medication class within 1 year (109/68) & 0.604 & 0.620 & 0.581 & 0.578 & 0.614 & 0.594 & 0.589 & 0.666 & 0.648 & 0.667 & -- & 0.528 & 0.747 \\
\quad Medication class recorded within 1 year (109/36) & 0.524 & 0.563 & 0.509 & 0.601 & 0.381 & 0.651 & 0.534 & 0.486 & 0.531 & 0.512 & -- & 0.557 & 0.740 \\
\quad Mean AUROC & 0.652 & 0.681 & 0.664 & 0.559 & 0.565 & 0.623 & 0.595 & 0.602 & 0.672 & 0.613 & -- & 0.643 & 0.766 \\
\quad Mean ECE (all outcomes) & 0.136 & 0.125 & 0.151 & 0.242 & 0.171 & 0.206 & 0.197 & 0.207 & 0.172 & 0.149 & -- & 0.143 & 0.145 \\
\quad Brier $>$ prev.\ (of 19) & 14 & 13 & 12 & 18 & 18 & 16 & 14 & 13 & 14 & 15 & -- & 16 & 12 \\
\bottomrule
\end{tabular}}

\vspace{2pt}
{\scriptsize --, MedDecider-9B returned invalid (non-finite) probabilities on these sets.}
\end{table}

\end{document}